\documentclass[lettersize,journal]{IEEEtran}
\usepackage{paralist}

\usepackage{tabu}                      
\usepackage{booktabs}                  
\usepackage{lipsum}                    
\usepackage{mwe}                       

\usepackage{mathptmx}                  

\usepackage{microtype}                 
\PassOptionsToPackage{warn}{textcomp}  
\usepackage{textcomp}                  
\usepackage{times}                     
\usepackage{cite}                      
\usepackage{amsmath}                   
\usepackage{amssymb}                   

\usepackage{tabularx}

\usepackage[table]{xcolor}
\usepackage{todonotes}
\usepackage{amsfonts}
\usepackage[pagebackref,bookmarks]{hyperref}

\newcommand{\zadu}{forced scale KL}

\begin{document}

\title{{
How Scale Breaks ``Normalized Stress" and KL Divergence: Rethinking Quality Metrics }
 \\
 {\small A preliminary version of this article appeared in the BELIV 2024 workshop~\cite{smelser2024normalized}.}
}

\author{%
    Kiran Smelser\IEEEauthorrefmark{1}, 
    Kaviru Gunaratne\IEEEauthorrefmark{2},
    Jacob Miller\IEEEauthorrefmark{2},
    Stephen Kobourov\IEEEauthorrefmark{2}
  \thanks{
  	Kiran Smelser is with the University of Arizona.
  	E-mail: ksmelser@arizona.edu\\
  	Kaviru Gunaratne, Jacob Miller, and Stephen Kobourov are with the Technical University of Munich.
  	E-mail: \{kaviru.gunaratne,jacob.miller,stephen.kobourov\}@tum.de
}
}

\maketitle


\begin{abstract}%
Complex, high-dimensional data is ubiquitous across many scientific disciplines, including machine learning, biology, and the social sciences. One of the primary methods of visualizing these datasets is with two-dimensional scatter plots that visually capture some properties of the data. 
{Because visually determining the accuracy of these plots is challenging, researchers often use quality metrics to measure the projection’s accuracy and faithfulness to the original data.}
One of the most commonly employed metrics, normalized stress, is sensitive to uniform scaling (stretching, shrinking) of the projection, despite this act not
meaningfully changing anything about the projection.
{ Another quality metric, the Kullback–Leibler (KL) divergence used in the popular t-Distributed Stochastic Neighbor Embedding (t-SNE) technique, is also susceptible to this scale sensitivity.
}
We investigate the effect of scaling on stress and { KL divergence} analytically and empirically by showing just how much the values change and how this affects dimension reduction technique evaluations. We introduce a simple technique to make 
{ both metrics} scale-invariant and show that it accurately captures expected behavior on a small benchmark. 
\end{abstract}


\begin{figure*}
 \centering
 \includegraphics[width=.85\linewidth]{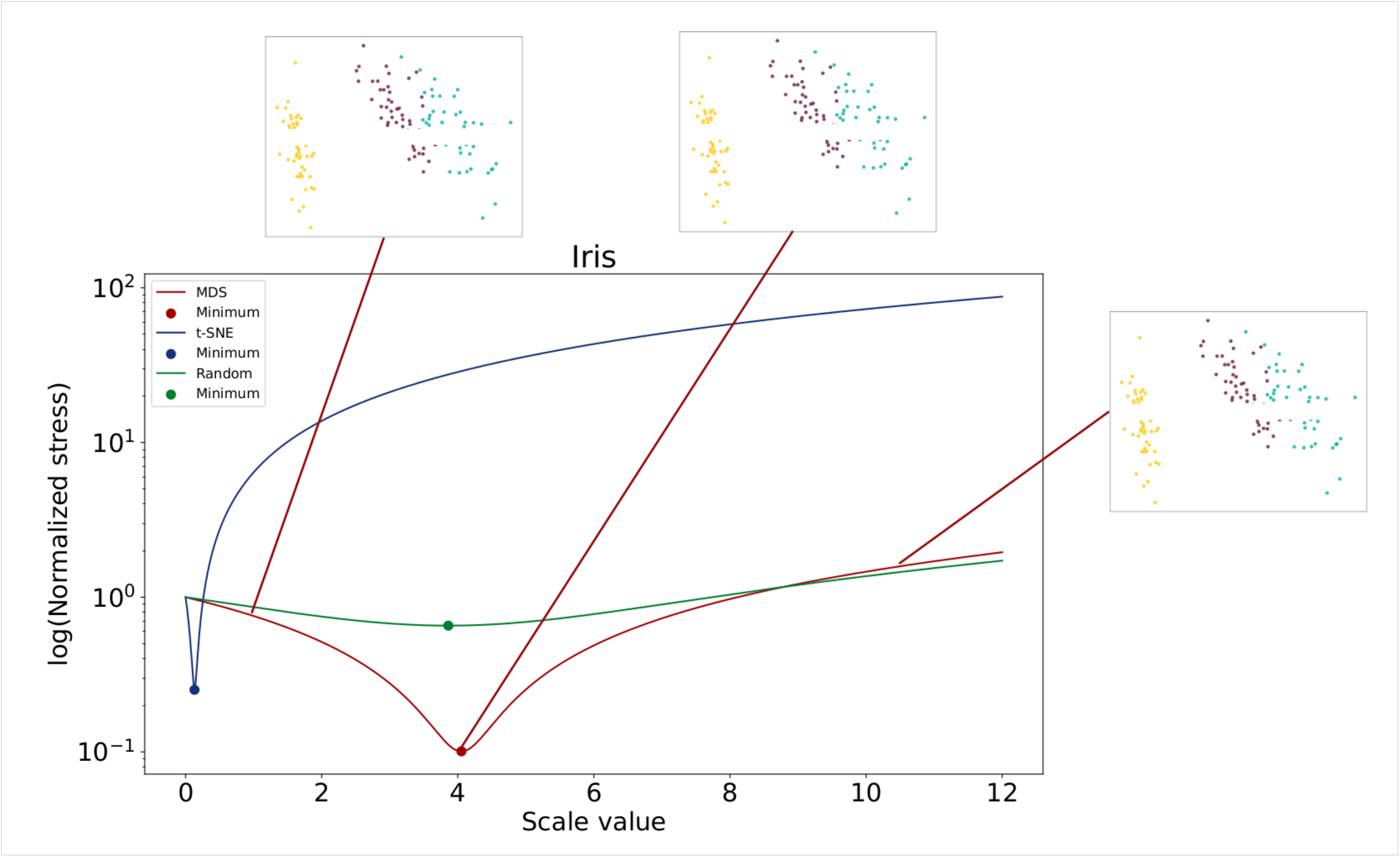}

 \includegraphics[width=0.32\linewidth]{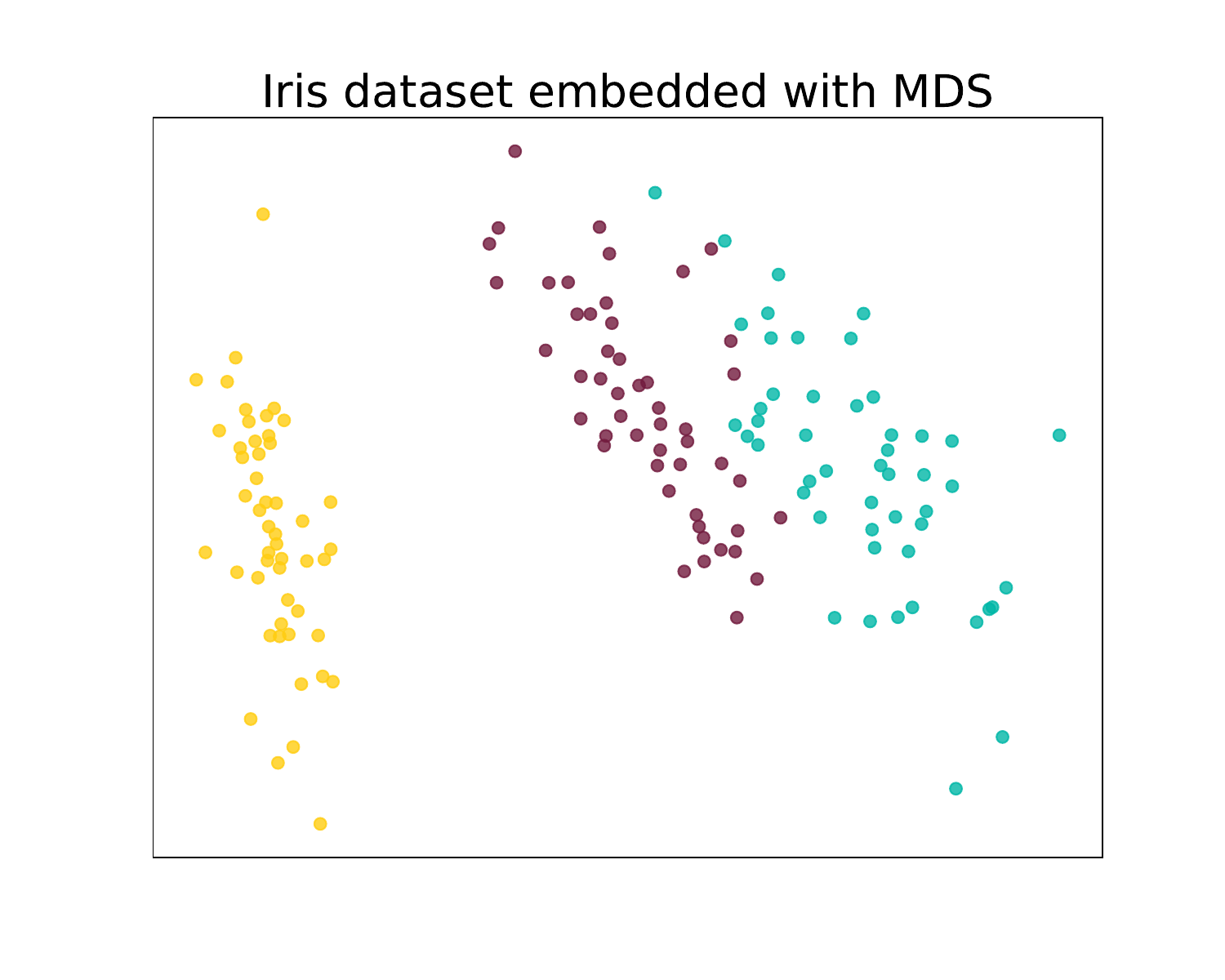}
 \includegraphics[width=0.32\linewidth]{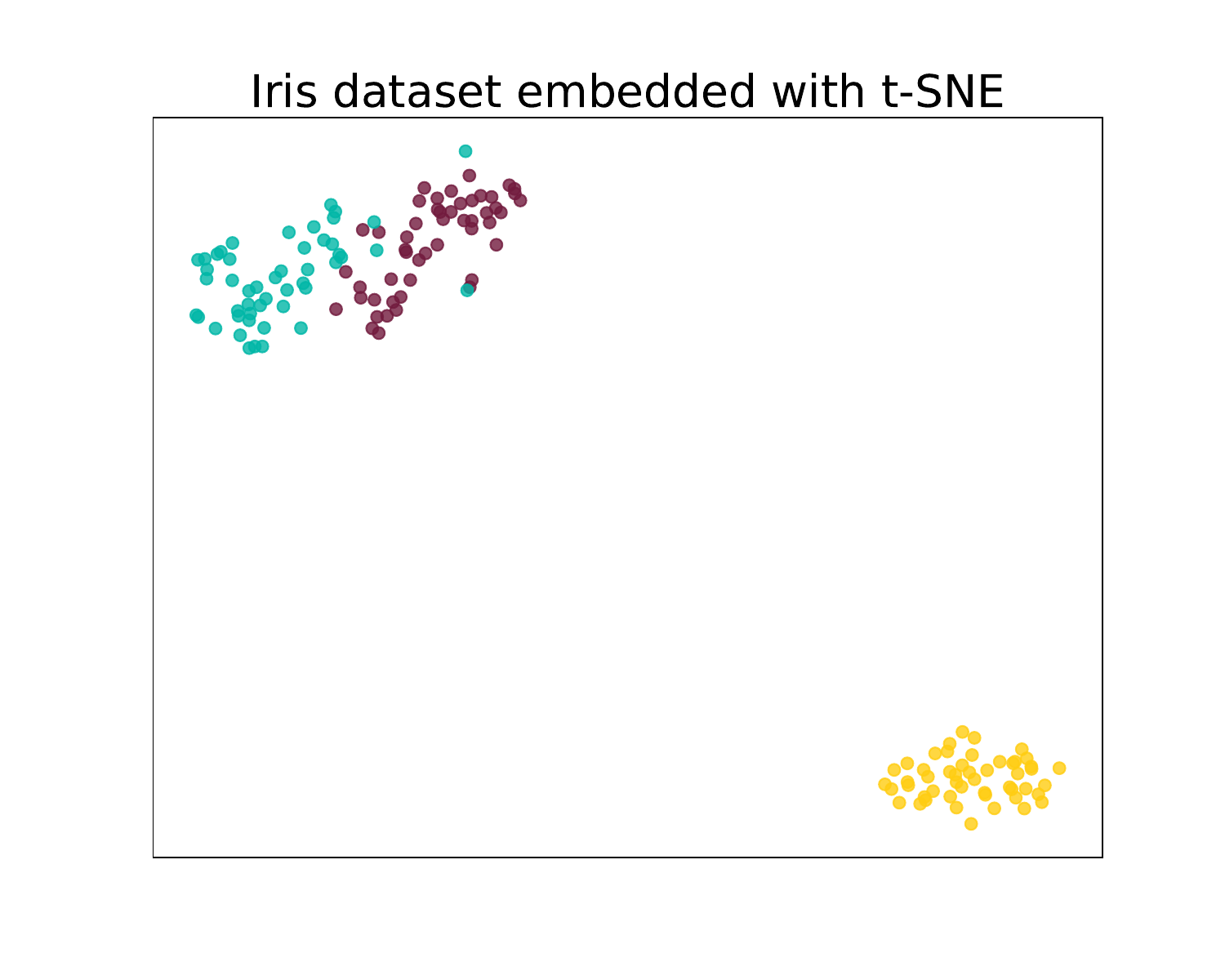}
 \includegraphics[width=0.32\linewidth]{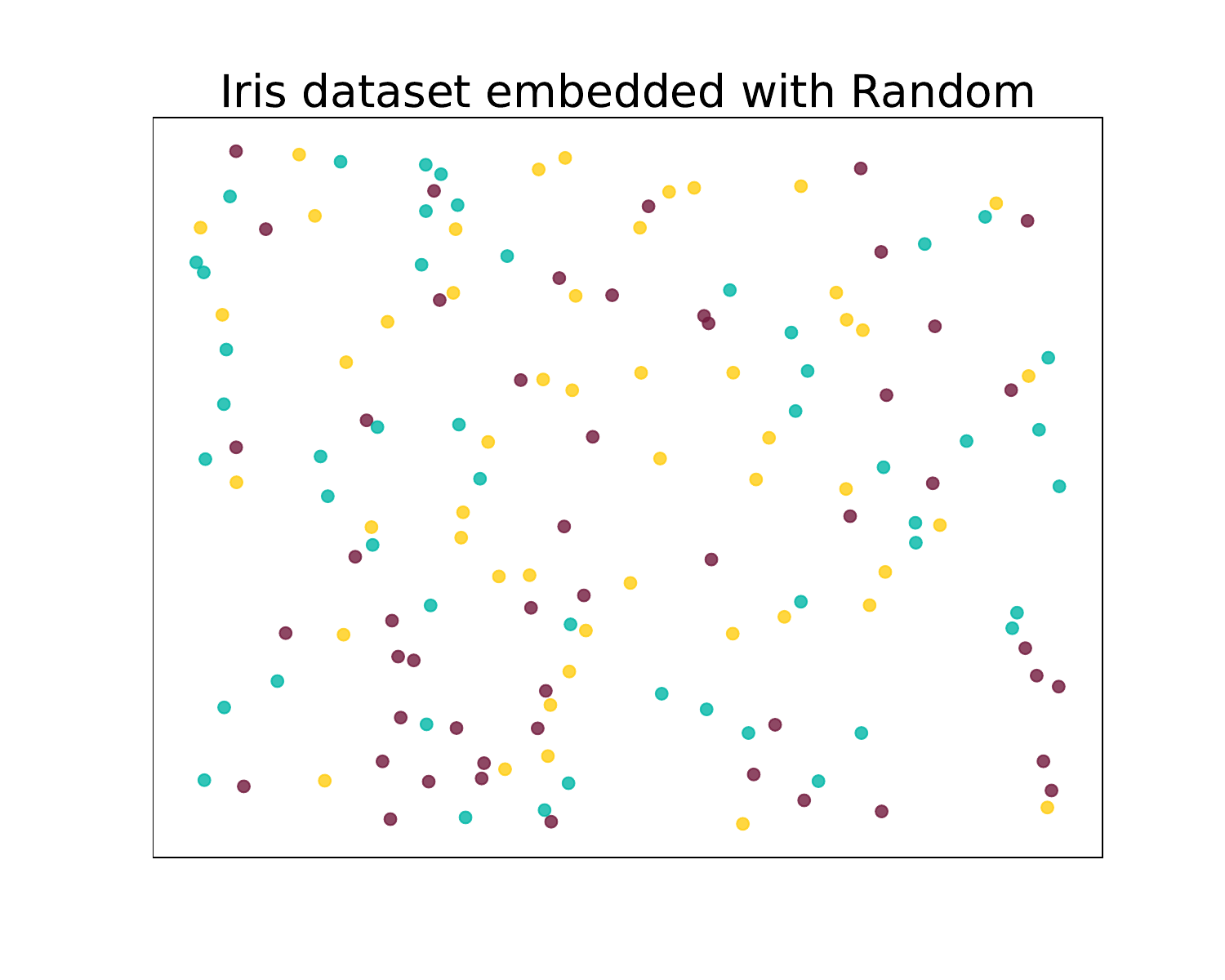} 
\caption{
{MDS~\cite{shepard1962analysis}, t-SNE~\cite{van2008visualizing}, and RND (random) embeddings of the well-known $Iris$ dataset from left to right (bottom).
The plot (top) shows the values of the {\em normalized stress} metric for these three embeddings and clearly illustrates the sensitivity to scale. As one uniformly scales the embeddings to be larger or smaller, the value of normalized stress changes. Notably, at different scales, different embeddings have lower stress, including the absurd situation where the random embedding has the lowest stress (beyond scale 9). 
Moreover, 
all six different algorithm orders can be found by selecting different scales.}}
 \label{fig:teaser}
\end{figure*}

\section{Introduction}

\IEEEPARstart{D}{imensionality} reduction (DR) techniques play an integral
role in the visualization of complex, high-dimensional datasets~\cite{10.5555/383784.383790,DBLP:conf/vissym/LiuMWBP15,DBLP:journals/tvcg/KehrerH13,DBLP:conf/www/TangLZM16}. DR techniques, which map high-dimensional data to the 2D computer screen, are pivotal in various fields, from bioinformatics~\cite{huang2022towards,DBLP:journals/bib/MaD11,DBLP:journals/bib/HilarioK08} to machine learning (ML)~\cite{DBLP:journals/corr/MakhzaniSJG15,DBLP:journals/corr/abs-2103-11251,osti_15002155}. 

Researchers typically evaluate the effectiveness of DR techniques such as Multidimensional Scaling (MDS)~\cite{torgerson1952multidimensional,shepard1962analysis,kruskal1964multidimensional} and t-Distributed Stochastic Neighbor Embedding (t-SNE)~\cite{van2008visualizing} using quantitative quality metrics.
Many of 
these metrics measure how well an embedding technique performs in preserving the structure and relationships inherent in the original high-dimensional data when projected onto a lower-dimensional space~\cite{DBLP:journals/tvcg/EspadotoMKHT21,DBLP:journals/tvcg/NonatoA19,DBLP:journals/ijon/LeeV09,DBLP:journals/ijon/MokbelLGH13,georg2004survey}.

One aspect that 
{receives}
little attention is 
the effect that scaling the projected data has on these quality metrics, which, to our knowledge, 
{is not}
well studied in the visualization literature. 
{
~\autoref{fig:teaser} shows a simple example of how much scale can affect evaluation results of stress -- by merely resizing the outputs of different techniques, one can dramatically alter the evaluation outcome, often leading to absurd conclusions (such as a random embedding having the lowest stress).
It should be clear that normalized stress
(perhaps often confused with the scale-invariant non-metric stress of Kruskal~\cite{kruskal1964multidimensional})
is sensitive to scale.

}

While stress is directly optimized by the MDS family of algorithms, it is a widely used evaluation metric for other DR methods, regardless of whether they optimize it explicitly, implicitly, or not at all. Nearly all successful DR techniques optimize some function that expresses the quality of the projection, e.g., the Rayleigh quotient for Principal Component Analysis (PCA)~\cite{DBLP:books/sp/Jolliffe86} or the cross-entropy for Uniform Manifold Approximation and Projection (UMAP)~\cite{DBLP:journals/corr/abs-1802-03426}. In addition to stress, we investigate the Kullback–Leibler (KL) divergence of the widely used t-SNE~\cite{van2008visualizing} algorithm and find that it also suffers from scale sensitivity. 

Only the SNE family of techniques optimize KL divergence directly, but this metric is valuable beyond its role as an optimization function. Evaluating with KL divergence provides a measure of how well the pairwise similarity distribution in high-dimensional space is preserved in the low-dimensional projection, making it a useful tool for benchmarking and diagnosing performance, even for algorithms that do not directly optimize it. This highlights the importance of considering metrics such as stress and KL divergence to understand the broader impacts of scale sensitivity when using optimization functions as quality measures. Additionally, evaluating such measures
is timely, addressing the critical research challenge of creating more thorough DR evaluations, recently recommended by Jeon et al.~\cite{jeon2025unveiling}.

Comparing embeddings generated by various DR techniques using a scale-sensitive metric {\em without taking scale into account} can be seen as picking an arbitrary point on the scale curve for \textit{each} embedding.  
This is a natural consequence of the fact that different DR techniques map the same dataset onto different low-dimensional spaces with different sizes. 
As a result, it is possible to achieve any desired ordering of the DR techniques, as demonstrated in~\autoref{fig:teaser} and further details in~\autoref{sec:metrics}.
We study the impact of scale on various measures~\cite{kruskal1964multidimensional} and propose safe and reliable variants for stress and KL divergence. 


Some passing mentions of scale sensitivity~\cite{DBLP:journals/tvcg/GansnerHN13,DBLP:journals/cgf/KruigerRMKKT17,DBLP:journals/cgf/WelchK17} and other explicit mentions~\cite{wang2023smartgd,van2024experimental} from papers in the graph visualization community show there is some awareness among experts in the community. Buja et al.~\cite{andreas2008data} were also aware of the scaling problem as early as 2008, and it is noted as a shortcoming of the stress metric by Nonato and Aupetit~\cite{DBLP:journals/tvcg/NonatoA19}. However, the scale-sensitivity property for DR methods does not seem to be widely known, as there are recent implementations that do not take scale into account.
{
Although scale-invariant versions of stress exist, such as the non-metric stress~\cite{kruskal1964multidimensional}, or the Shepard goodness score~\cite{shepard1962analysis}, they are less intuitive, and are not widely used in practice. We also analyze these variants as potential alternatives, though our proposed \textit{scale-normalized stress} (SNS) is conceptually simple in addition to being scale-invariant.
}
In particular, our contributions are: 
\begin{compactitem}
    \item {
    Showing analytically and empirically that normalized stress and KL divergence are severely affected by scale and can lead to incorrect conclusions
    }
    \item Demonstrating that using a scale-invariant stress measure impacts the results of previous work
    \item { Proposing scale-invariant options for stress and KL divergence}

\end{compactitem}
We also provide an interactive webpage that demonstrates the problems discussed here with 6 datasets.\hyperref[sup-1]{\footnotemark[1]}, and all experiments (including datasets, metrics, and embeddings) can be recreated from our open-source implementation\hyperref[sup-1]{\footnotemark[2]}.


\section{Background and Related Work}

{
We begin by defining terms used throughout the paper. We then give a brief background on DR techniques and their classification. Finally, we discuss the quality metrics used to evaluate these techniques and also address a less explored aspect: the impact of scaling 
on these metrics.
}

\subsection{Definitions}
We use the following definitions and notations, also found in
Espadoto et al.~\cite{DBLP:journals/tvcg/EspadotoMKHT21},
with an emphasis on visualization. DR is the act of performing a transformation on a collection of objects (observations) found in an $n$-dimensional space (typically, $n \geq 3$) and representing them in a lower $t$-dimensional space (typically, $t = 2$) while preserving some properties of the original data. 

A matrix $X \in \mathbb{R}^{N \times n}$ represents this collection of $N$ high-dimensional objects, with its rows corresponding to the positions of objects in this space.
The DR transformation results in a matrix $Y \in \mathbb{R}^{N \times t}$ called the projection. 
{For the purpose of visualization, $t$ is typically either $2$ or $3$.}

The $i^{\text{ }th}$ row of matrix $X$, denoted as $x_i$, represents the position of object $i$ in the $n$-dimensional space. Its corresponding position in low-dimensional space is $y_i$. 
{ We refer to the pairwise distance between data points $x_i$ and $x_j$ in the high-dimensional space as $d(x_i,x_j)$ (which may come from any metric), and we indicate the pairwise distance between the corresponding embedded points $y_i$ and $y_j$ in the low-dimensional space by $||y_i - y_j ||$, the Euclidean distance.}

\begin{table*}[ht]
    \centering
    \renewcommand{\arraystretch}{1.5}
    \caption{Overview of the stress metrics under consideration.}
    \begin{tabular}{| c c c c |}
         \hline 
         Name & Formula & Scale invariant? & References\\ [0.5ex] \hline
         Raw stress & $\sum_{i,j} [d(x_i,x_j) - ||y_i - y_j ||]^2$ & No & \cite{torgerson1952multidimensional}\\ [1ex] \hline 
         Normalized stress & $\frac{\sum_{i,j}[d(x_i,x_j) - ||y_i - y_j ||]^2}{\sum_{i,j}d(x_i,x_j)^2}$ & No & 
         \parbox[c]{2cm}{\centering \cite{DBLP:journals/corr/abs-2403-05882,DBLP:journals/bspc/DasanP21,DBLP:journals/corr/abs-2003-09017,DBLP:conf/sibgrapi/MarcilioEG17,DBLP:conf/apvis/AmorimBNSS14,DBLP:conf/ieeevast/AmorimBDJNS12,DBLP:journals/tvcg/PaulovichNML08,DBLP:journals/ivs/NevesFMFP18,DBLP:journals/cgf/PaulovichEPBMN11,DBLP:conf/eurova-ws/MachadoT023,DBLP:conf/igarss/ChenCG06,DBLP:conf/sigmod/FaloutsosL95,DBLP:journals/tvcg/NonatoA19, DBLP:conf/pacificvis/MillerHNHK24,hossain2020multi}}\\ [2ex] \hline 
         Shepard goodness & Spearman correlation $d(x_i,x_j), ||y_i - y_j ||$ & Yes & \cite{DBLP:journals/tvcg/JoiaCCPN11,DBLP:conf/eurova-ws/MachadoT023,DBLP:journals/tkdd/ConnorV24,DBLP:journals/ivs/NevesFMFP18,DBLP:journals/cgf/PaulovichEPBMN11}\\ [0.5ex]      \hline   
         Non-metric stress & $\frac{\sum_{i,j}[||\hat{d}(x_i,x_j) - ||y_i - y_j ||]^2}{\sum_{i,j}||y_i - y_j ||^2}$ & Yes & \cite{DBLP:journals/tkdd/ConnorV24,DBLP:journals/jcc/Agrafiotis03}\\ [2ex] \hline 
         Scale-normalized stress & $\min\limits_{\alpha > 0}\frac{\sum_{i,j}[d(x_i,x_j) - \alpha ||y_i - y_j ||]^2}{\sum_{i,j}d(x_i,x_j)^2}$ & Yes & [{proposed}]\\ [1.5ex]
         \hline 
         
    \end{tabular}
    \newline
    \label{tab:metrics}
\end{table*}




\subsection{DR Techniques}\label{sec:techniques}

{
Embedding a sufficiently complex high-dimensional dataset into a lower-dimensional space unavoidably causes distortion, which results in the loss of some information in the dataset~\cite{DBLP:conf/beliv/DasguptaK12}. The  information lost in the low-dimensional embedding may be distances, neighborhoods, clusters, or other aspects present in the true high-dimensional data. Distortion types include stretching, compression, gluing, and/or tearing~\cite{DBLP:journals/ijon/Aupetit07}. It is known that such distortions can cause misinterpretations; for instance, t-SNE can create the illusion of non-existent clusters~\cite{wattenberg2016how}.}

{
DR techniques can be broadly categorized by the type of structure they aim to preserve. \textit{Global} methods focus on preserving the overall structure and pairwise distances between all data points. PCA~\cite{DBLP:books/sp/Jolliffe86} and MDS~\cite{torgerson1952multidimensional,shepard1962analysis,kruskal1964multidimensional} are primary examples. In contrast, \textit{local} methods prioritize preserving local neighborhood relationships around each point, even at the expense of global accuracy. This category includes popular manifold learning techniques such as Locally Linear Embedding (LLE)~\cite{doi:10.1126/science.290.5500.2323}, Isomap (ISO)~\cite{doi:10.1126/science.290.5500.2319}, t-SNE~\cite{van2008visualizing}, and UMAP~\cite{DBLP:journals/corr/abs-1802-03426}. For a complete summary of relevant techniques, see the recent surveys by Espadoto et al.~\cite{DBLP:journals/tvcg/EspadotoMKHT21} and Nonato and Aupetit~\cite{DBLP:journals/tvcg/NonatoA19}.
The techniques of MDS and t-SNE are particularly notable landmark techniques in the DR literature. Both of these techniques 
{are} widely studied from a theoretical and mathematical viewpoint~\cite{DBLP:journals/corr/SorzanoVP14,DBLP:journals/ijautcomp/Yin07,DBLP:journals/jmlr/CunninghamG15,DBLP:conf/vluds/EngelHH11}, as well as for their effectiveness in visualizing high-dimensional datasets~\cite{laurens2008dimensionality}. Additionally, they represent opposite ends of the local-global spectrum for non-linear DR; t-SNE prioritizes local neighbors while MDS optimizes all pairwise distances.

While much of the DR literature focuses on proposing new techniques, we aim to evaluate the DR quality metrics of stress and KL divergence. Prior work has collected and analyzed quality metrics~\cite{DBLP:journals/tvcg/EspadotoMKHT21,DBLP:conf/visualization/JeonCJLHKJS23},and 
{note} limitations of relying purely on quality metrics~\cite{jeon2025unveiling,machado2025necessary}. However, to our knowledge, there is no work on the scale sensitivity of quality metrics.
}

\subsection{Quality Metrics and Scale}\label{sec:metrics}

{
To quantitatively assess the accuracy of embeddings and the distortion they introduce, DR researchers rely on quality metrics. There are many such metrics, each differing in what it computes. Formally, a (scalar) quality measure is a function $M: (\mathbb{R}^{N \times n}, \mathbb{R}^{N \times  t}) \rightarrow \mathbb{R}$ which describes the information faithfulness of an embedding. 
}

{We can categorize these metrics as either local measures, global measures, or cluster-level measures~\cite{DBLP:conf/visualization/JeonCJLHKJS23}.}

Local measures gauge the preservation of the neighborhood structure of the high-dimensional dataset in the low-dimensional embedding.
This includes measures such as Trustworthiness \& Continuity~\cite{DBLP:journals/nn/VennaK06}, Mean Relative Rank Errors~\cite{DBLP:journals/ijon/LeeV09}, and Neighborhood Hit~\cite{DBLP:journals/tvcg/PaulovichNML08}. 

Cluster-level measures assess the maintenance of the cluster structures inherent in the high-dimensional data in the low-dimensional embedding.
This includes  Steadiness \& Cohesiveness~\cite{DBLP:journals/tvcg/JeonKJKS22} and Distance Consistency~\cite{DBLP:journals/cgf/SipsNLH09}. 

Global measures evaluate the degree to which pairwise distances between objects remain consistent between the high- and low-dimensional data. 
All stress-based measures in~\autoref{tab:metrics} fall under this classification and are described in~\autoref{sec:stress}.  

The low-dimensional embeddings of a dataset by different DR techniques often lead to the original high-dimensional data being projected into distinct low-dimensional spaces. The size or bounding box of these spaces may vary greatly depending on the technique employed. For instance, a technique designed for visualization might prefer to output the units as pixel coordinates{~\cite{keim1996pixel}}, which will be a large region compared to another technique designed for ML, which outputs coordinates between 0 and 1{~\cite{DBLP:conf/cvpr/RedmonDGF16}}. 

This becomes a problem when an embedding evaluation metric is not scale-invariant, meaning that the value of the metric changes when the size of the output changes. Just as translation and rotation, uniformly scaling up or down, does not meaningfully change the relationships present in an embedding.

For this reason, when computing the score of a quality metric that is susceptible to scale, one must be careful when comparing the quality of two different embeddings produced by different DR techniques.
If the scale of one low-dimensional embedding is significantly smaller or larger than the scale of another low-dimensional embedding, this difference may lead to unreasonably high values of normalized stress, indicating poor preservation of the original high-dimensional data structure. 

{The three stress-scale curves plotted in~\autoref{fig:teaser} demonstrate this.}
{Computing a scale-sensitive metric for embeddings (such as normalized stress) from different DR techniques is equivalent to picking an arbitrary point on each technique's respective curve in~\autoref{fig:teaser}. Since different DR techniques  produce outputs at different scales, this procedure can lead to any desired ranking among them, making comparisons unreliable and potentially misleading.}

Several quality metrics first began life as optimization criteria for a successful DR technique. Stress is the most notable and widely used of this type, but KL divergence (optimized by t-SNE) is occasionally also employed as a quality metric. In fact, the more widely used trustworthiness and continuity measures~\cite{venna2001neighborhood,kaski2003trustworthiness} can be obtained from KL divergence, as shown by Venna et al.~\cite{venna2010information} (trustworthiness and continuity are analogous to precision and recall, a measure of the number of false and missing neighbors in the embedding, respectively).

In principle, any optimization function employed by a DR technique can be used to evaluate a dimension-reduced embedding, even if the embedding was obtained by a  different technique, and there are good reasons to do so. First, the quality metric can be used as a benchmark against proven standards; if a new technique achieves similar values of a quality metric to a well-established method, this indicates the new technique is likely trustworthy. Additionally, such quality metrics (derived from optimization functions) can help provide a holistic picture of the performance of a new DR technique by indicating how similar the new technique is to well-understood techniques.

\section{Stress}\label{sec:stress}
Stress is one of
the most commonly utilized metrics 
to gauge the quality of low-dimensional projections using DR techniques~\cite{DBLP:journals/tvcg/EspadotoMKHT21,DBLP:journals/tvcg/NonatoA19,DBLP:conf/sigmod/FaloutsosL95,DBLP:journals/tvcg/PaulovichNML08}.
{Shepard first introduced stress in the field of psychometrics in the 1960s, and Kruskal later expanded on it~\cite{kruskal1964multidimensional,kruskal1964nonmetric,shepard1962analysis}. While Kruskal's definition is in fact scale invariant, the normalized stress metric that 
{is} widely used is not.}
The goal of these papers 
{is} to develop a DR techniques that 
{maintain} the pairwise distances between data points as well as possible. 

{Researchers have proposed different stress variants as objective functions to achieve this type of embedding. In general, a stress measure, denoted as a real-valued function $M: (\mathbb{R}^{N \times n}, \mathbb{R}^{N \times t}) \rightarrow \mathbb{R}$, 
maps the pair of a high-dimensional matrix $X$ and its embedding $Y$ to a real number.}

\subsection{Scale-sensitive stress measures}
As discussed previously, the most popular stress measures are scale-sensitive. A scale-sensitive measure is a measure in which the value changes as the size of the projection changes. 
In other words, for a metric $M(X,\alpha Y)$, the result is a non-constant function of $\alpha$. 


\begin{figure*}[ht]
    \centering
    \includegraphics[width=0.49\linewidth]{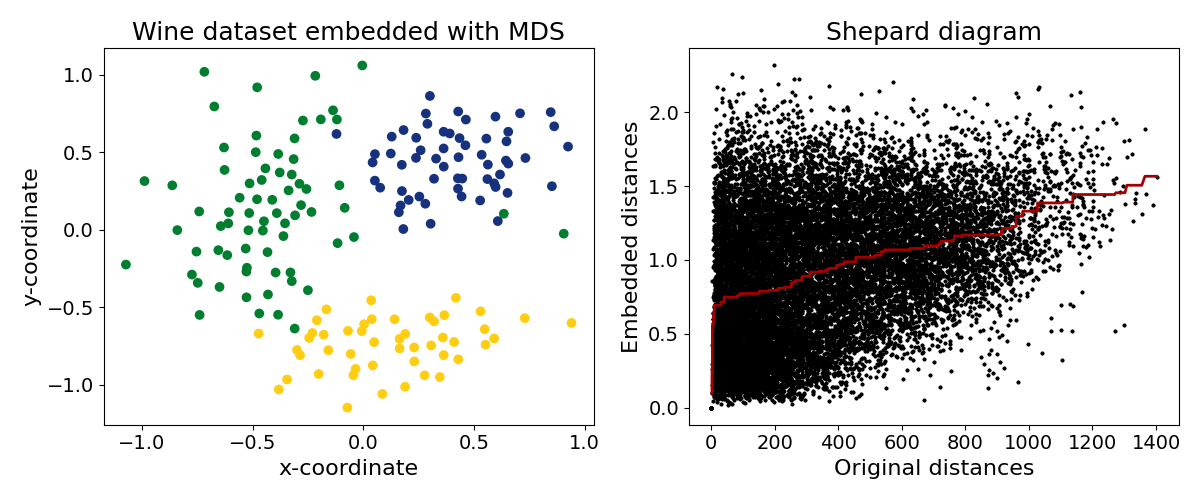}        
    \includegraphics[width=0.49\linewidth]{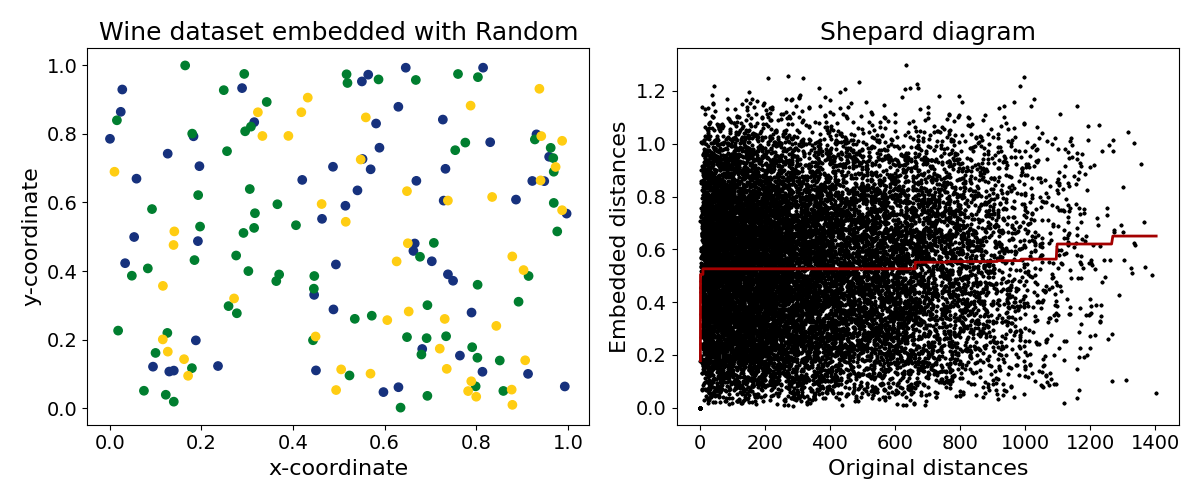}
    \caption{{
    Embeddings and Shepard diagrams with monotonic fitted line for the $Wine$ dataset captured by MDS (left, $\rho=0.36$) and RANDOM (right,$ \rho=0.001$). Note the positive correlation for MDS, which indicates good distance preservation and the low correlation for RANDOM, which  indicates bad distance preservation. Note that scaling the embedding corresponds to stretching or shrinking the $y$-axes in these plots, which does not effect the (rank) correlation.
    }}
    \label{fig:random-shepard}
\end{figure*}

\noindent \textbf{Raw Stress (RS)}
The first measure that arose is the measure of goodness of fit, referred to as raw stress (RS), which is the difference between distances in the high-dimensional space and distances in the low-dimensional space. This first appeared in Torgerson's {classical} MDS paper~\cite{torgerson1952multidimensional}. Raw stress is formally defined as follows:
\begin{equation}
    \label{eq:raw-stress}
    \text{RS(X,Y)}=\sum_{i,j}[d(x_i,x_j)-||y_i - y_j ||]^2.
\end{equation}
However, although raw stress is invariant under distortions such as rotations, translations, and reflections, it is variant under stretching and compression distortions~\cite{kruskal1964multidimensional}. We can show this analytically: 
\[\sum_{i,j}[d(x_i,x_j)- ||\alpha y_i - \alpha y_j||]^2 = \sum_{i,j}[d(x_i,x_j)- \alpha||y_i - y_j ||]^2\]
\[\sum_{i,j}d(x_i,x_j)^2 + \alpha\sum_{i,j}d(x_i,x_j) \cdot ||y_i - y_j || + \alpha^2\sum_{i,j}||y_i - y_j ||^2\text{.}\]
We see that we can rewrite the raw stress as a quadratic function of $\alpha$.

\noindent \textbf{Normalized Stress (NS)}
{
Raw stress can produce very large stress values even for reasonably good embeddings, making it not ideal to use as an evaluation metric. For this reason, the normalized stress (NS) measure has become more popular to use as the range of possible values tends to be much smaller. Likely the first instance of normalized stress is its use in Sammon mapping~\cite{sammon1969nonlinear}, a type of metric MDS. In its original context, normalized stress was not intended to be used as an evaluation metric but as an optimization function.
We refer to the square root of the sum of squared differences between the original and reduced distances, divided by the sum of squared original distances, as normalized stress~\cite{DBLP:journals/tvcg/EspadotoMKHT21}.
}
\begin{equation}
    \label{eq:norm-stress}
    \text{NS(X,Y)}=\sqrt{\frac{\sum_{i,j}[d(x_i,x_j)-||y_i - y_j ||]^2}{\sum_{i,j}d(x_i,x_j)^2}}
\end{equation}
Taking the square root reduces the impact of large residuals and emphasizes smaller ones (but does not change any ordering). This measure ensures that the stress value is more sensitive to smaller discrepancies between the original and reduced distances, which can be crucial when measuring the preservation of the structure of the data during DR. 

Just like raw stress, normalized stress is scale-sensitive.
Note that although one might assume that normalized stress has values in the range $[0,1]$ (e.g., as Espadoto et al.~\cite{DBLP:journals/tvcg/EspadotoMKHT21}), this is indeed not true. As $||y_i - y_j ||$ is unbounded, the range of normalized stress is also unbounded. 

Scaling the term $||y_i - y_j ||$ by $0$ results in a normalized stress value of $1$, 
because the low dimensional distances go to $0$.
Thus, the normalized stress curve of any DR technique (with respect to scale) initially starts at value $1$, decreases until reaching a minimum value, and then increases in an unbounded fashion. 
The quadratic nature of the normalized stress function, defined in~\autoref{eq:norm-stress}, causes this behavior.
Hence, the normalized stress curves of pairs of DR techniques have a common intersection at $x=0$, as illustrated in~\autoref{fig:teaser}, and possibly one more positive intersection further to the right.


\subsection{Scale-invariant Stress Measures}
While the most popular stress measures are scale-sensitive, there are indeed some scale-invariant ones, i.e., they are constant with respect to $\alpha$ for a measure $M(X,\alpha Y)$.

\noindent \textbf{Shepard Goodness Score (SGS)}
One variant of stress introduced in 1962 is now known as the Shepard goodness score (SGS)~\cite{shepard1962analysis}. It has two  components: the Shepard diagram and the Spearman rank correlation.

The Shepard diagram is a scatterplot of all $\binom{N}{2}$ pairwise distances between points in the high-dimensional space versus the corresponding pairwise distances in the low-dimensional space.
In other words, the Shepard diagram is a collection of two-dimensional points. 
\[c_{i,j} = (d(x_i,x_j), ||y_i - y_j ||) \; \forall 1 \leq i < j \leq N\]

This can be determined by visual assessment: as the scatter plot approaches a positive diagonal line, the preservation of distances between the two spaces improves~\cite{DBLP:journals/tvcg/EspadotoMKHT21}.~\autoref{fig:random-shepard} shows an example of a projection with a good Shepard diagram, and a bad Shepard diagram.

The Shepard goodness score is then the Spearman rank correlation of the $x$ and $y$ components of the diagram, i.e., the smallest distance in high-dimensional space should still be the smallest in the projection, and the $k$ smallest distance should still be the $k$ smallest. 
{The rank correlation, being a unitless measure based on the variance of the ranks, remains invariant to scale.}
Stretching or shrinking the projection uniformly will not change the ranking of distances. 

\noindent \textbf{Non-metric stress (NMS)}
\label{sec:kruskal}
Another variant of stress proposed in 1964 is non-metric stress (NMS), sometimes referred to as Kruskal stress~\cite{kruskal1964multidimensional}. Non-metric stress does not measure the preservation of the pairwise distances between points in high-dimensional and low-dimensional spaces. Instead, it measures how well the ranking (or ordering) of the distances is maintained between the spaces { trading specificity for generality}. Formally, non-metric stress is defined as follows:
{
\begin{equation}
    \label{eq:non-metric-stress}
    \text{NMS(X,Y)}=\frac{\sum_{i,j}[\hat{d}(x_i, x_j) - ||y_i - y_j ||]^2}{\sum_{i,j}||y_i - y_j ||^2},
\end{equation}
where $\hat{d}(x_i,x_j)$ is a monotone transformation of $d(x_i,x_j)$ obtained by isotonic regression, i.e. the fitted values of $d(x_i,x_j)$ against $||y_i - y_j||$, subject to the constraint that $\hat{d}$ is non-decreasing. 
~\autoref{fig:random-shepard} demonstrates this monotonic curve.

It is not immediately obvious why this metric is scale-invariant. 
The key lies in the fitted distances $\hat{d}$, which are obtained by isotonic regression of the low-dimensional distances. If the embedding $Y$ is scaled by a factor $\alpha$, then all low dimensional distances $||y_i - y_j||$ are scaled by the same factor. Since the regression fit is monotonic, the fitted values $\hat{d}(x_i,x_j)$ are also scaled by $\alpha$.
Thus, when scaled, the $\alpha$ factors cancel, and the metric is scale invariant:
\[NMS(X,\alpha Y) = \frac{\sum_{i,j}[\alpha \hat{d}(x_i,x_j) - \alpha ||y_i - y_j ||]^2}{\sum_{i,j}(\alpha ||y_i - y_j ||)^2}\]
\[ = \frac{\alpha^2 \sum_{i,j}[\hat{d}(x_i,x_j) - ||y_i - y_j ||]^2}{\alpha^2 \sum_{i,j}||y_i - y_j ||^2}\]
}






\noindent \textbf{Scale-normalized Stress (SNS)}\label{sec:SNS}
As we have already noted theoretically and empirically, both the raw stress and normalized stress metrics are scale-sensitive. We also 
{show} analytically that they are, in fact, quadratic functions of the scaling factor $\alpha$. This lends itself to the nice property that there is a unique minimum (since the leading term must be positive).  
As a result, these functions are parabolic in the scaling factor $\alpha$; see~\autoref{fig:teaser}.
%
With this in mind, we propose the {\em scale-normalized stress (SNS)} as a `quick fix' for the normalized scale metric:
\begin{equation}
    \label{eq:scale-norm-stress}
    SNS(X,Y) = \min\limits_{\alpha > 0} 
    {\frac{\sum_{i,j}[d(x_i,x_j) - \alpha ||y_i - y_j ||]^2}{\sum_{i,j}d(x_i,x_j)^2}}
\end{equation}



This produces the ``true” normalized stress score, as it gives us scale invariance and is straightforward to explain. 
%
We note that the scale sensitivity of stress-based metrics has been observed in the graph visualization literature, and scale-normalized stress values 
{have been} computed~\cite{DBLP:journals/tvcg/GansnerHN13,DBLP:journals/cgf/KruigerRMKKT17}. 
We  next show how to compute the minimum directly and efficiently. 

To find the value of $\alpha$ that minimizes the normalized stress function (\autoref{eq:norm-stress}), we take the derivative of the numerator with respect to $\alpha$ and set it equal to zero. This is given by:
\begin{gather*}
    \frac{d}{d\alpha} \sum_{i,j}[d(x_i,x_j)-\alpha||y_i - y_j ||]^2 = 0.
\end{gather*}
Solving this equation gives us the optimal value of $\alpha$:
\begin{gather*}
    \alpha = \frac{\sum_{i,j}[d(x_i,x_j)\cdot||y_i - y_j ||]}{\sum_{i,j}d(x_i,x_j)^2}.
\end{gather*}

The denominator in the $\alpha$ equation normalizes the distances so that the scale of the distances in the high-dimensional space does not affect the value of $\alpha$. The numerator is essentially a covariance between the high-dimensional distances and the low-dimensional distances, ensuring that $\alpha$ captures the relationship between the two sets of distances. 
{We can interpret the equation for $\alpha$ as a correlation coefficient between the high-dimensional and low-dimensional distances.}
An implementation that calculates the aforementioned scale-normalized stress metric is available in our code \hyperlink{https://github.com/KiranSmelser/dim-reduction-metrics-and-scale}{repository}.

{
\textbf{Forced-Scale Normalized Stress (FSNS)}
In addition to established metrics, we also evaluate a variant suggested during peer review, which we denote as Forced-Scale Normalized Stress (FSNS): 

\[
FSNS(X,Y) = \frac{\sum_{i<j} \left(d'(x_i,x_j) - \frac{1}{\delta'} \lVert Y_i - Y_j \rVert \right)^2}{\sum_{i<j} d'(x_i,x_j)} ,
\]

where \(d'(x_i,x_j) = \tfrac{d_(x_i,x_j)}{\max_{p,q} d_(x_p,x_q)}\) and \(\delta' = \tfrac{1}{\max_{p,q} \lVert Y_p - Y_q \rVert}\).  
In other words, both the input distances and the embedding distances are independently rescaled so that their respective maximum values equal one.  

While this definition is scale-invariant (any scalar applied to the embedding distances is canceled out by the normalization with $\delta'$), it implicitly assumes that the input and output distance distributions have similar overall shape. 

}

\section{Kullback-Liebler Divergence}
\label{sec:KL}

\begin{figure*}[ht]
\centering
    \centering
    \includegraphics[width=0.49\linewidth]{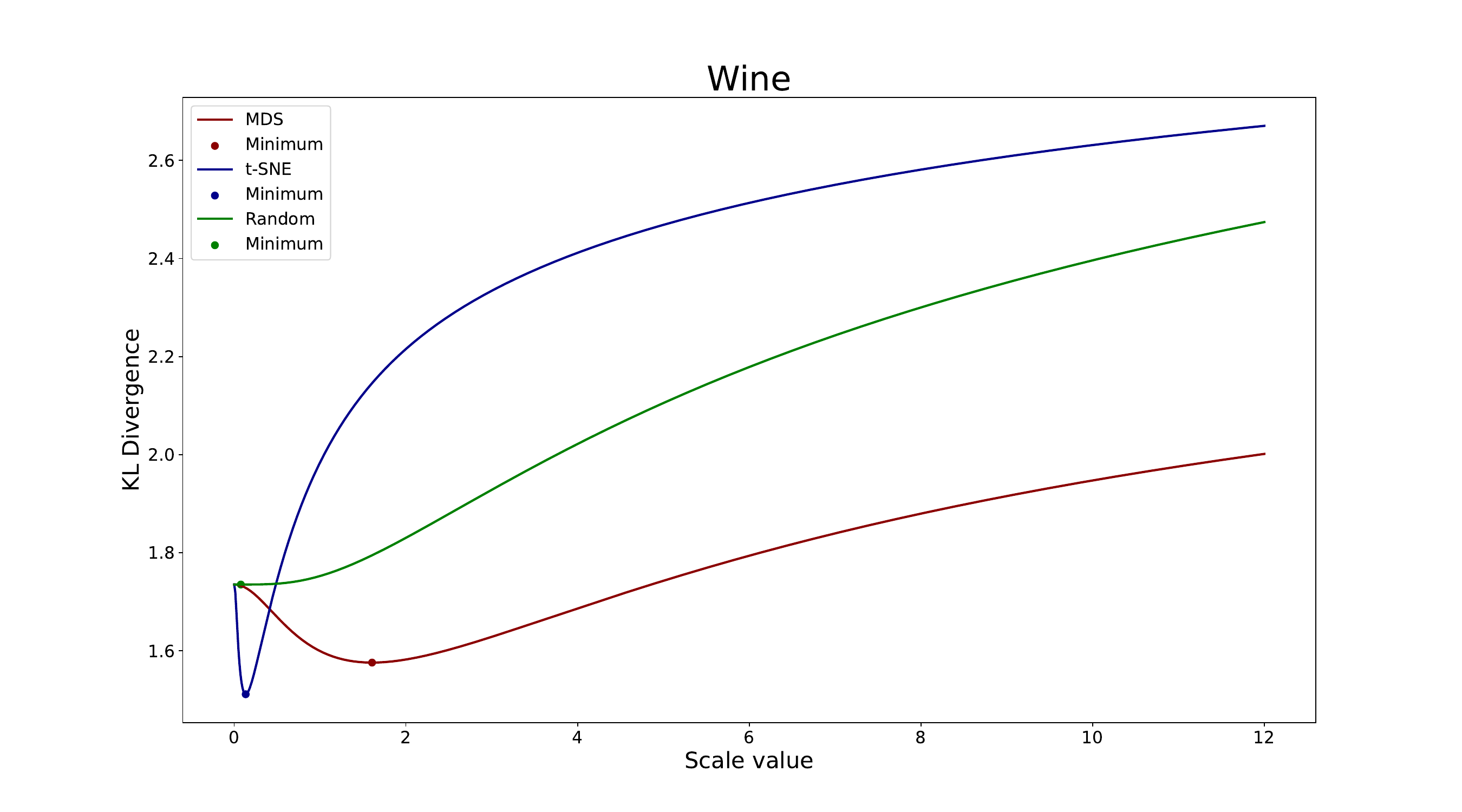}
    \includegraphics[width=0.49\linewidth,height=5cm]{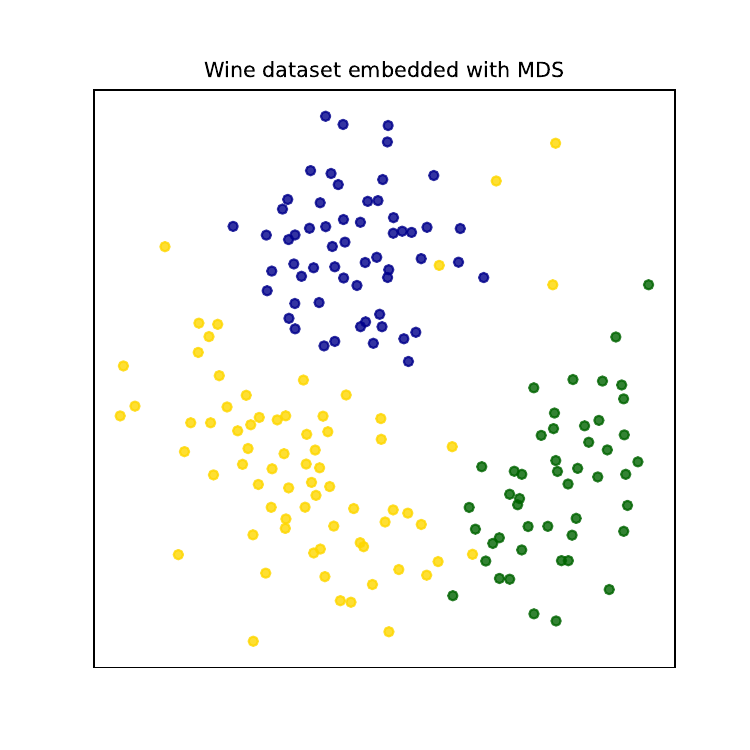}
    
    \includegraphics[width=0.49\linewidth,height=5cm]{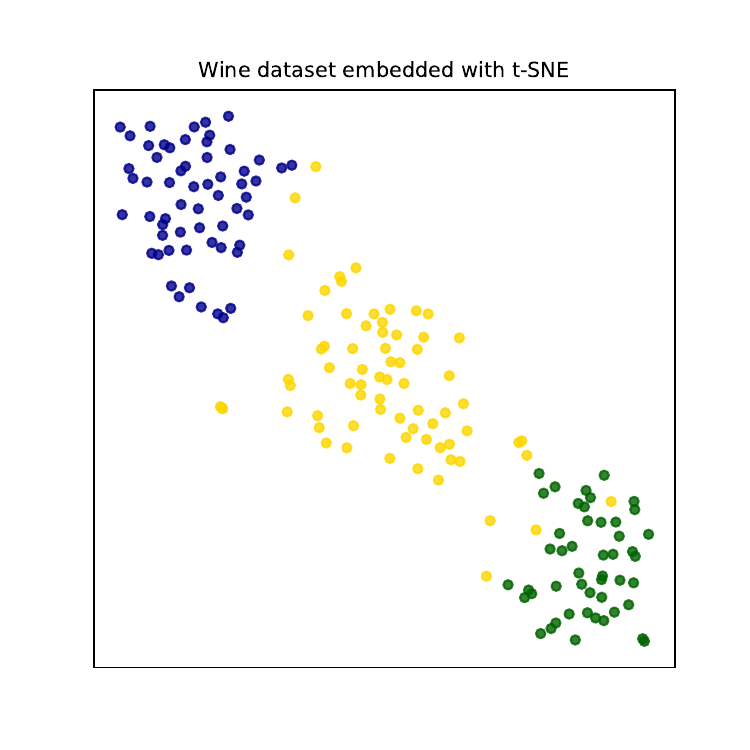}
    \includegraphics[width=0.49\linewidth,height=5cm]{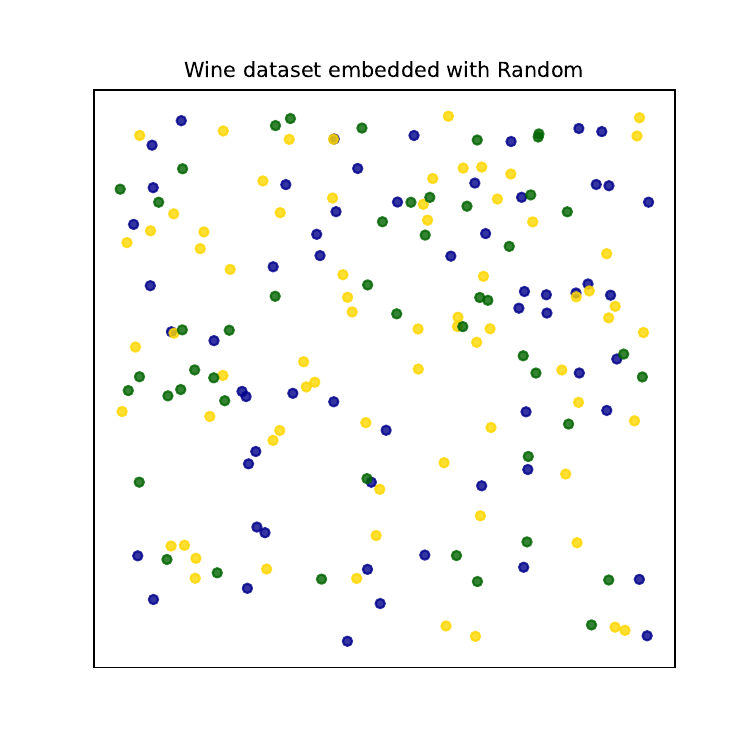}
\caption{
RND, MDS, and t-SNE embeddings of the Wine dataset (right, bottom left) compared against the variation of KL Divergence with scale (top left). 
KL divergence varies substantially with scale, and the expected order of t-SNE $<$ MDS $<$ Random is not always observed. As with stress, all six possible algorithm orders can occur depending on scale.
}
\label{fig:kl-wine}
\end{figure*}

While stress has become successful as a quality metric for evaluating dimension reduced embeddings, it was originally proposed as a loss function.
The success and popularity of MDS (which optimizes stress) indicate that low stress is a desirable quality in a projection. 
Loss functions for other techniques can be used in much the same way. 

Here, we consider the loss function of t-SNE \cite{van2008visualizing} as a quality metric,
defined as the KL divergence between 
two carefully selected probability distributions, $P$ and $Q$.
Since t-SNE is known to preserve local relationships well, KL divergence can be used as a local quality measure.
We show that KL divergence in the original implementation is scale-sensitive while proposing a few scale-invariant variants.
Just like with stress, a KL divergence variant is said to be scale-sensitive if the value of $KL(P\| Q(\alpha))$ varies with $\alpha$ and is scale-invariant otherwise.

\subsection{Scale-sensitive Divergences}
\label{sec:scale-sensitive-kls}
We discuss the divergences defined by variants of the t-SNE algorithm that are scale-sensitive when used as quality measures. 

\noindent
\textbf{t-SNE KL Divergence (Student's $t$)}
\label{sec:t-SNE-KL}
{
In t-SNE, the joint probability distributions $P = \{p_{ij}\}$ and $Q = \{q_{ij}\}$ act as similarity measures between points in the dataset and embedding respectively. When $p_{ij}$ is close to 1, this indicates that points $x_i$ and $x_j$ are similar in the high-dimensional dataset.
The similarity matrix $P$ is derived from the high dimensional input data $X$ as follows:
\begin{equation*}
    p_{ij} = \dfrac{p_{i|j} + p_{j|i}}{2N}  \\
\end{equation*}
\begin{equation*}
    p_{j|i} = \dfrac{\exp{[-d(x_i,x_j)^2 / 2\sigma_i^2]}}{\sum_{k\neq l} \exp{[-d(x_i,x_j)^2 / 2\sigma_i^2]}}
\end{equation*}

This describes a Gaussian kernel, whose variance $\sigma_i$ is obtained through the selection of the perplexity parameter. 

$Q$, on the other hand, is derived from the embedding, and is thus potentially affected by scale. We denote this as $Q(a)$, which is the probability distribution derived from $\alpha Y$. Unlike with $P$, $Q$ is modelled using a Student's $t$ kernel as follows:

\begin{equation}
    \label{eq:q-tsne}
    q^{ST}_{ij}(\alpha) = \frac{(1+\alpha^2||y_i - y_j||^2)^{-1}}{\sum\limits_{k\neq l} (1+\alpha^2||y_k - y_l||^2)^{-1}}
\end{equation}

The KL divergence of $P, Q(\alpha)$ is:
\begin{equation}
    \label{eq:kl-div-short}
    \text{KL}(P \| Q(\alpha)) = \sum_{i \neq j} p_{ij} \times \log \frac{p_{ij}}{q_{ij}(\alpha)}
\end{equation}

}

\autoref{fig:kl-wine} illustrates that this results in a metric that is sensitive to scale. 
The value of the function can and does change when varying the size of outputs from the embedding algorithms. More importantly, the order of the three algorithms with respect to this value changes as well.

The Student's $t$ KL divergence as a function of $\alpha$ begins at $\alpha =0$, which is a constant point that only depends on the high-dimensional dataset.
This point can either be a local maximum or a local minimum (although empirically it is only a local minimum in worse-than-random embeddings). 
In all cases, we observe at most one local minimum, much like stress. 
Unlike stress, however, 
it is unclear how to analytically derive where this point occurs.
In all cases, Student's $t$ KL divergence has an asymptotic behavior at infinity, while stress increases arbitrarily.
We further analyze this behavior in supplemental material.




\noindent \textbf{Gaussian $Q$ KL Divergence }
\label{sec:Gaussian-KL}
{
t-SNE models $Q$ with the Student's $t$ kernel in order to allow moderately far away points in the dataset to be projected much further away, preventing clusters from clumping together too closely. This change from a Gaussian kernel to a Student's $t$ kernel is considered to be the key difference that made t-SNE more effective than its predecessor, SNE~\cite{DBLP:conf/nips/HintonR02}.

However, one may still be interested in making a more symmetric KL divergence metric using a Gaussian kernel instead. This is similar to the method proposed by Moor et al.~\cite{DBLP:conf/icml/MoorHRB20} but without using density distributions. 
This measure, $KL_G$, uses a Gaussian kernel in both $P$ and $Q$ by leaving $P$ as before and replacing $Q(\alpha)$ with $Q_G(\alpha)$ as follows:

}
\begin{equation}
    \label{eq:q-sne}
    q_{ij}^{G}(\alpha) = \frac{\exp (-\alpha^2 ||y_i - y_j||^2)}{\sum_{k\neq l} \exp (-\alpha^2 ||y_k - y_l||^2)}
\end{equation}
\[Q_G(\alpha) = \{q_{ij}^G(\alpha)\}\]
\begin{equation*}
    KL_G = KL(P\| Q_G)
\end{equation*}

{

$KL_G$ shares some characteristics with t-SNE's KL divergence as a function of scale, which are explored in supplemental material.
However, as seen in \autoref{fig:kl-coil20-gaussian} unlike t-SNE's KL divergence, we may observe multiple local minima as the Gaussian KL divergence varies with scale. Since $KL_G$ is not unimodal, finding its global minimum with respect to scale is a much more difficult task.  

This shows that the exact choice of kernel used in $Q$ greatly affects the behavior of KL divergence with respect to scale. In general, it is not recommended to use $KL_G$ as a quality measure unless it is important to have a symmetric divergence between the high and low distances.

\begin{figure}[ht]
    \centering
    \includegraphics[width=\linewidth]{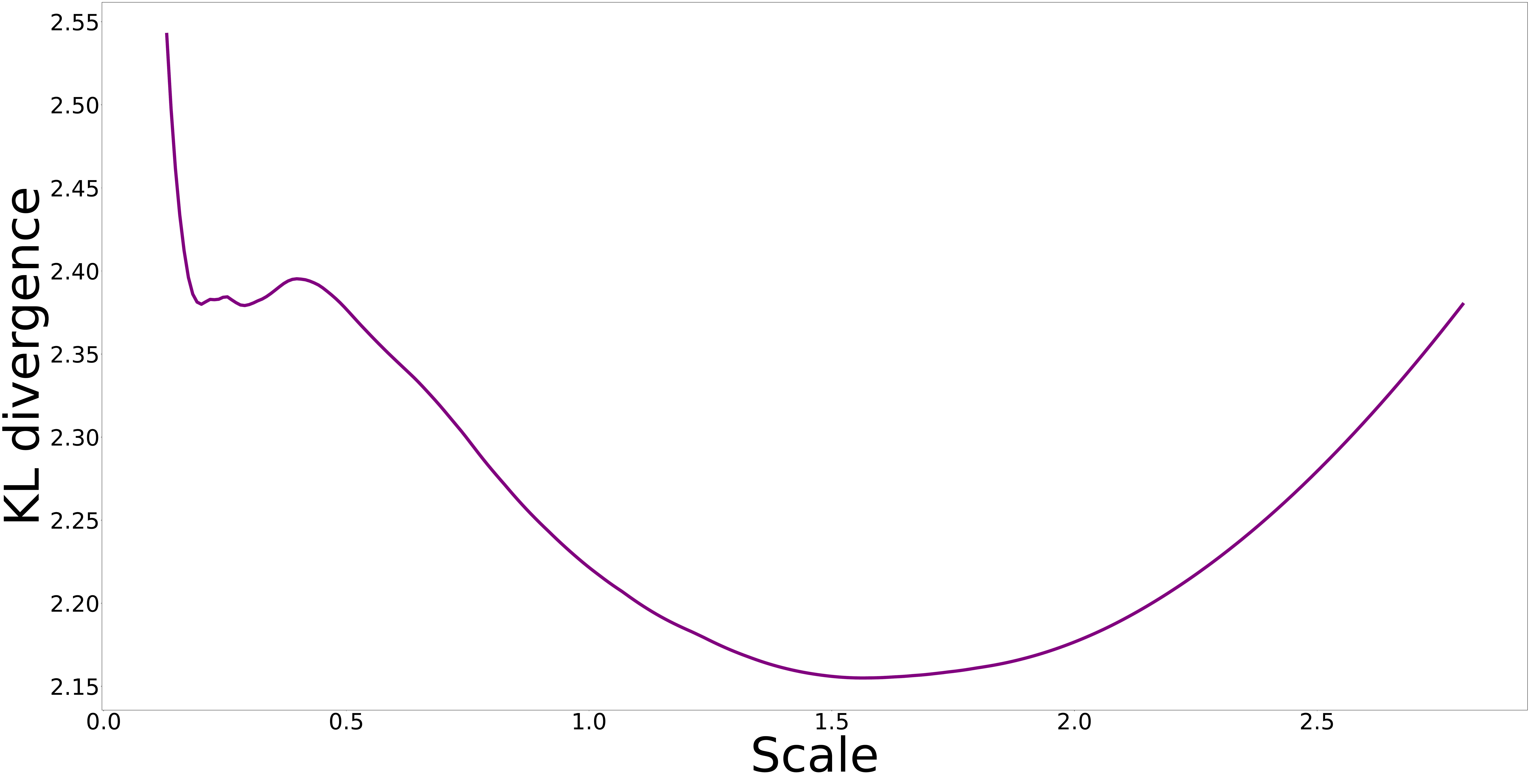}
    \caption{The variation of KL divergence with the Gaussian kernel implementation described in \autoref{sec:Gaussian-KL} with respect to scale. The COIL-20 dataset is used here and an embedding is generated from UMAP with the default parameters. The \texttt{random\_state} and \texttt{n\_jobs} parameters in UMAP 
    {are} set to 42 and 1 respectively for reproducibility.}
    \label{fig:kl-coil20-gaussian}
\end{figure}

}

{
\subsection{Scale-invariant Divergences}

The following divergences are scale-invariant, i.e., they do not depend on the size of the input coordinates.
{
Due to the erratic behavior of $KL_G$, all the following metrics are derived from the original t-SNE KL divergence implementation described in \autoref{sec:t-SNE-KL}.}

\noindent \textbf{Scale-normalized KL Divergence}
Our experiments indicate that KL divergence as defined in \autoref{sec:t-SNE-KL} is a unimodal function with respect to scale. Considering this, we utilize the core idea from \textit{scale-normalized stress} and propose \textit{scale-normalized KL divergence}:

\begin{equation*}
    SNKL(X,Y) = \min\limits_{\alpha > 0} \; KL(P \| Q_{ST}(\alpha))
\end{equation*}

SNKL has many of the same advantages as \textit{scale-normalized stress}, including that it is relatively straightforward to understand and scale-invariant by definition. One notable downside is 
that 
we do not have a closed-form equation for the minimum.

{
Since KL divergence is continuous and empirically unimodal with respect to scale, the minimum value for $\alpha$ can be found efficiently. 
In our experiments we use
the golden-section search as implemented in Scipy~\cite{2020SciPy-NMeth}.
}
It is possible that KL divergence strictly decreases as $\alpha\rightarrow \infty$ (see example in supplemental material
), although not for any of the 16 datasets 
{we test.}

\noindent \textbf{Forced Scale KL Divergence (FSKL)}
\label{sec:fskl}
Moor et al.~\cite{DBLP:conf/icml/MoorHRB20} outline a metric that compares density distributions of the dataset and the projection using KL divergence. These density distributions are generated from the pairwise distance matrix according to the distance-to-a-measure density estimator~\cite{DBLP:journals/focm/ChazalCM11}. In order to make these measures scale invariant, they normalize the pairwise distances between 0 and 1 before calculating the density distributions, an approach also used by Jeon et al.~\cite{ jeon2023classes,DBLP:conf/visualization/JeonCJLHKJS23}. 

This idea can be used to make KL divergence scale-invariant by normalizing the distance matrices before calculating $P$ and $Q$. We call this \zadu ~(FSKL).
While FSKL has the benefit of scale invariance, it is simply choosing a  fixed scale in which to evaluate. This chosen scale is only determined by the greatest distance in the embedding and is not informed at all by the structure. So, while conceptually simple, we do not expect this measure to perform well in practice.


\section{Stress Experiments}
\label{sec:stress-experiments}

In this section, we demonstrate, beyond the few examples shown up to this point, the frequency at which stress metrics behave in an expected manner and agree with one another when comparing the performance of DR techniques. The goal is to identify the stress metrics that most often agree as well as consistently rank embeddings from high-quality DR techniques better than embeddings from poor-quality DR techniques. 
In particular, we 
{address three questions} through empirical experiments: 
\begin{compactenum}
    \item[\textbf{Q1}]: { Which stress metrics behave consistently as the embedding quality worsens?} 
    \item[\textbf{Q2}]: { Is it possible that incorrect conclusions could be drawn by not accounting for scale? }
    \item[\textbf{Q3}]: Have scale-sensitive metrics affected the results of previous DR evaluations? 
\end{compactenum}

{We 
{design} three experiments to address each research question. Below, we describe the datasets, DR techniques, and quality metrics that we use in the experiments.}

\subsection{Experiment description}
\label{sec:stress-exp-desc}

{We implement the experiments in Python, using} the SKLearn, UMAP, Scipy, and ZADU libraries; the code is 
available.\hyperref[sup-2]{\footnotemark[2]}
An accompanying webpage to help  explore the behavior of normalized stress is also available.\hyperref[sup-1]{\footnotemark[1]}

\noindent \textbf{Datasets}

\begin{table}[ht]
\centering

\caption{\label{tab:datasets}
Size and dimensionality of the dataset used in our experiments. The top block is widely used datasets found in Scikit-learn~\cite{DBLP:journals/jmlr/PedregosaVGMTGBPWDVPCBPD11}, while the bottom block are datasets found in Espadoto et al.~\cite{DBLP:journals/tvcg/EspadotoMKHT21}.
Datasets marked with a star do not have embeddings available from Espadoto et al.'s supplemental material.
}
\small 
\begin{tabular}{|c c c c|}
\hline
Dataset & Type & Size ($N$) & Dimensionality ($n$) \\
\hline
iris & tabular & 149 & 4 \\
\hline
wine & tabular & 178 & 13 \\
\hline
swiss roll & synthetic & 1500 & 3 \\
\hline
penguins & tabular & 342 & 4 \\
\hline
auto-mpg & multivariate & 391 & 5 \\
\hline
s-curve & synthetic & 1500 & 3 \\
\hline
\hline
bank & multivariate & 2059 & 63 \\
\hline
cifar10 & image & 3250 & 1024 \\
\hline
cnae9 & text & 1080 & 856 \\
\hline
coil20 & image & 1440 & 400 \\
\hline
epileptic & tabular & 5750 & 178 \\
\hline
fashion-mnist & image & 3000 & 784 \\
\hline
fmd & image & 997 & 1536 \\
\hline
har & multivariate & 735 & 561 \\
\hline
hatespeech & text & 3222 & 100 \\
\hline
hiva & tabular & 3076 & 1617 \\
\hline
imdb & text & 3250 & 700 \\
\hline
orl$^*$ & image & 400 & 396 \\
\hline
secom & multivariate & 1567 & 590 \\
\hline
seismic & multivariate & 646 & 24 \\
\hline
sentiment & text & 2748 & 200 \\
\hline
sms & text & 836 & 500 \\
\hline
spambase$^*$ & text & 4601 & 57 \\
\hline
svhn & image & 733 & 1024 \\
\hline
\end{tabular}
\vspace{0.15cm}
\vspace{-0.5cm}

\end{table}

{The datasets for our experiments} 
{cover a wide range of characteristics and complexities.}
They include the well-known \textit{Iris}~\cite{Unwin2021TheID}, \textit{Wine}~\cite{DBLP:journals/pr/AeberhardCV94}, \textit{Swiss Roll}~\cite{DBLP:books/daglib/0029547}, \textit{Palmer Penguins}~\cite{10.1371/journal.pone.0090081}, \textit{Auto MPG}~\cite{DBLP:data/10/Quinlan93}, and \textit{S-Curve}~\cite{DBLP:journals/jmlr/PedregosaVGMTGBPWDVPCBPD11} datasets. In addition to these, we 
{also include} 18 datasets from the recent survey of DR techniques
by Espadoto et al.~\cite{DBLP:journals/tvcg/EspadotoMKHT21}. 
\autoref{tab:datasets} details some statistics about these datasets.
%
The comprehensive corpus of datasets we use allows us to evaluate the behavior of stress under various conditions.
{
Unless otherwise specified, we compute two-dimensional embeddings of these datasets for experiments, and fully replicable code is included in our experimental pipeline. 
}

{
\noindent \textbf{Techniques}
Throughout our experimental section, we consider several widely used DR techniques. Multidimensional Scaling (MDS)~\cite{sammon1969nonlinear} serves as our primary baseline, as it directly optimizes the normalized stress function of Equation~\ref{eq:norm-stress}. We further include the popular neighbor embedding methods t-SNE~\cite{van2008visualizing} and UMAP~\cite{DBLP:journals/corr/abs-1802-03426}, as well as the more classical manifold learning technique Isomap~\cite{doi:10.1126/science.290.5500.2319}. Finally, to establish a lower-bound reference, we consider a uniformly random sample of points in the unit square, referred to as the Random embedding. Note that this embedding contains no meaningful structure and has no correlation with the input data.
}

\noindent \textbf{Metrics}
{
The quality metrics used to assess the chosen DR techniques throughout the following experiments include all metrics described in~\autoref{sec:metrics}. Raw stress is omitted, as we note it is only differs from normalized stress by a constant factor; see~\autoref{tab:metrics}.

We use the implementations of normalized stress and Shepard Goodness from the ZADU library~\cite{DBLP:conf/visualization/JeonCJLHKJS23}. 
We 
{implement} all remaining stress measures based on the definitions provided in~\autoref{sec:stress}, the code for which is made available.}

\subsection{Stress metric consistency}
\begin{figure*}
    \centering
    \includegraphics[width=0.32\linewidth]{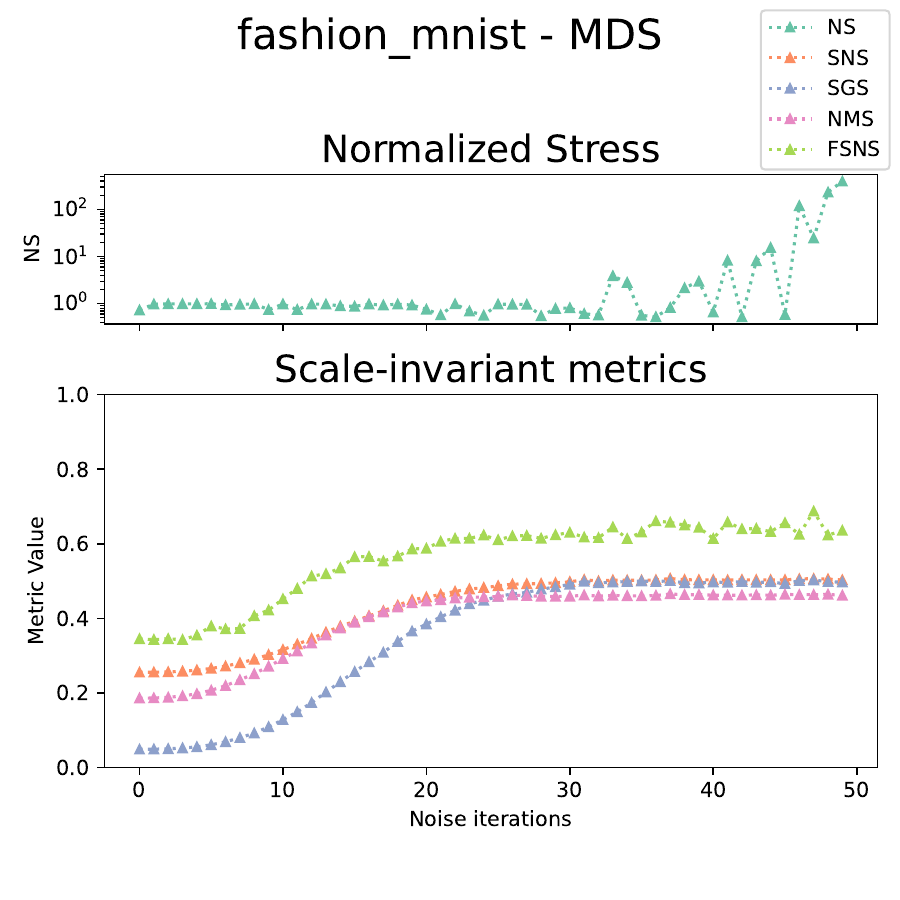}
    \includegraphics[width=0.32\linewidth]{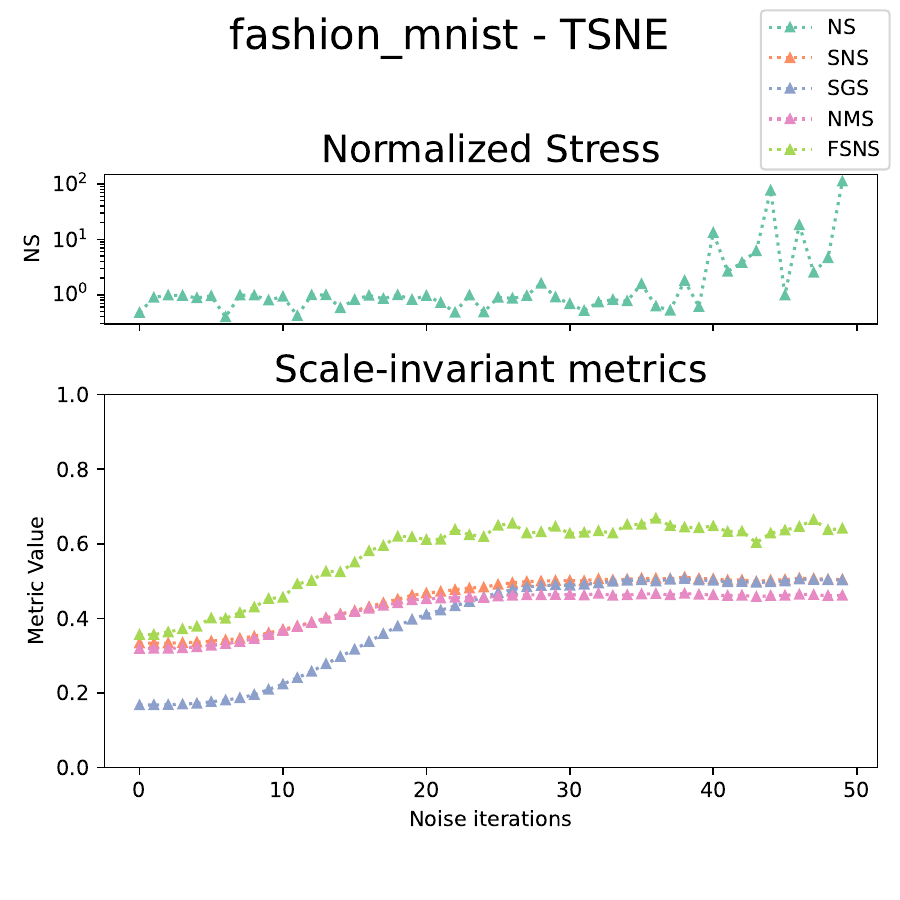}
    \includegraphics[width=0.32\linewidth]{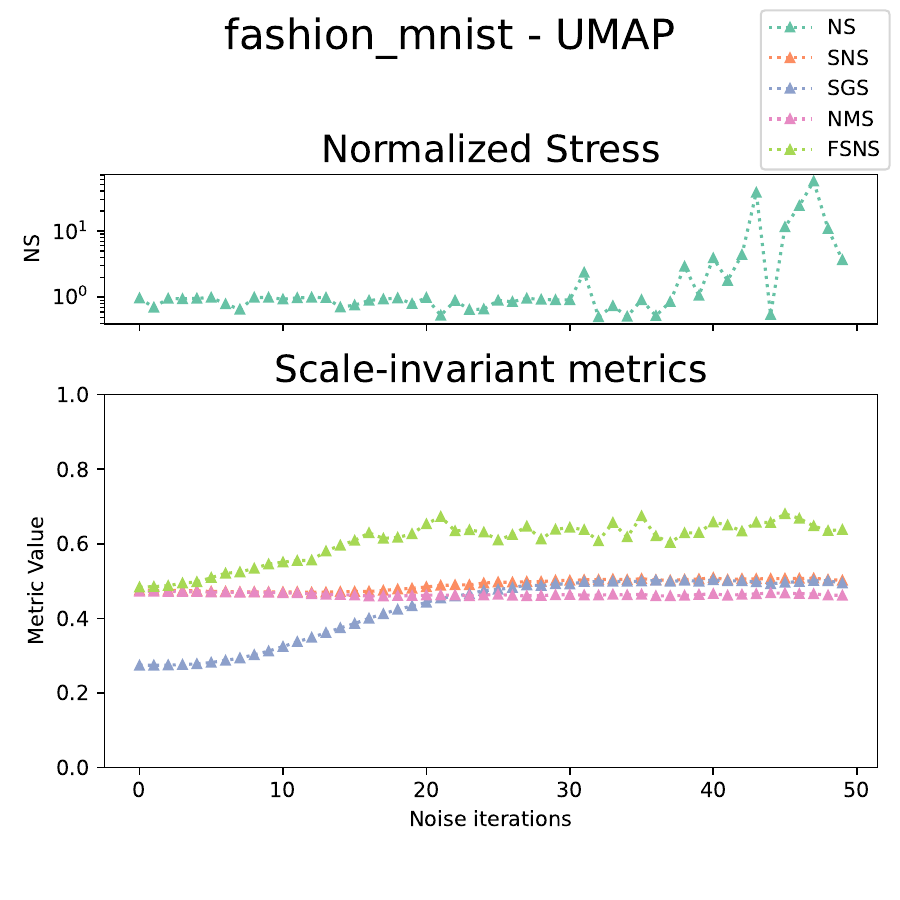}

    \includegraphics[width=0.32\linewidth]{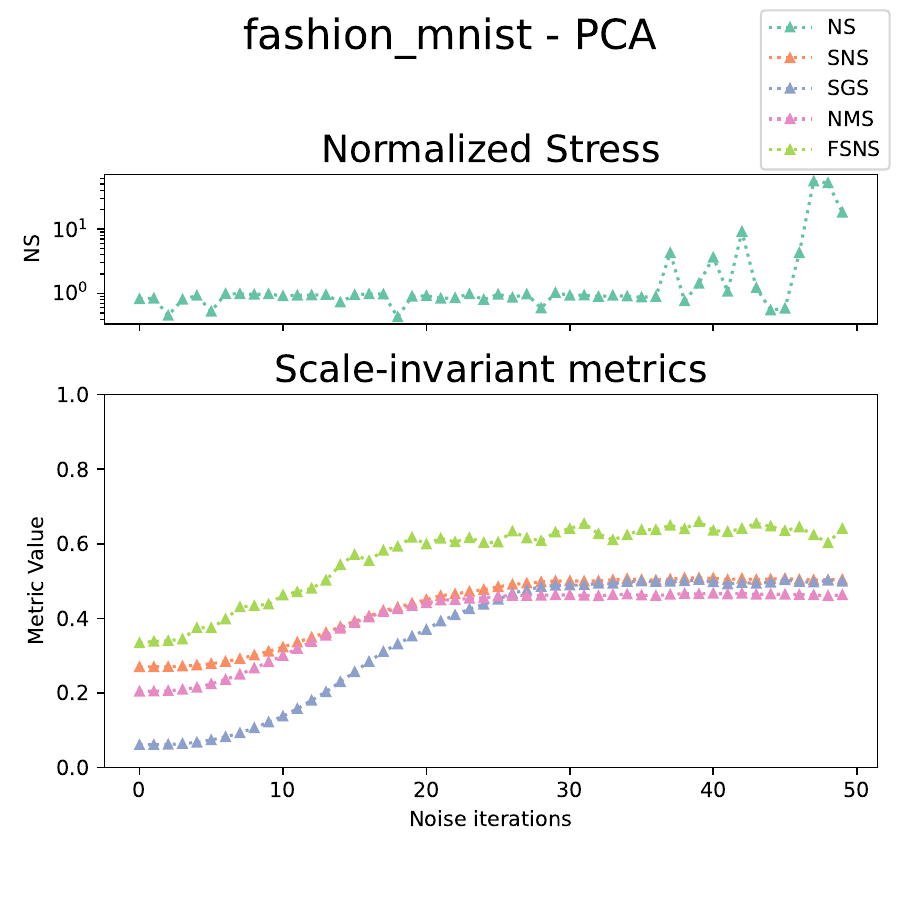}    
    \includegraphics[width=0.32\linewidth]{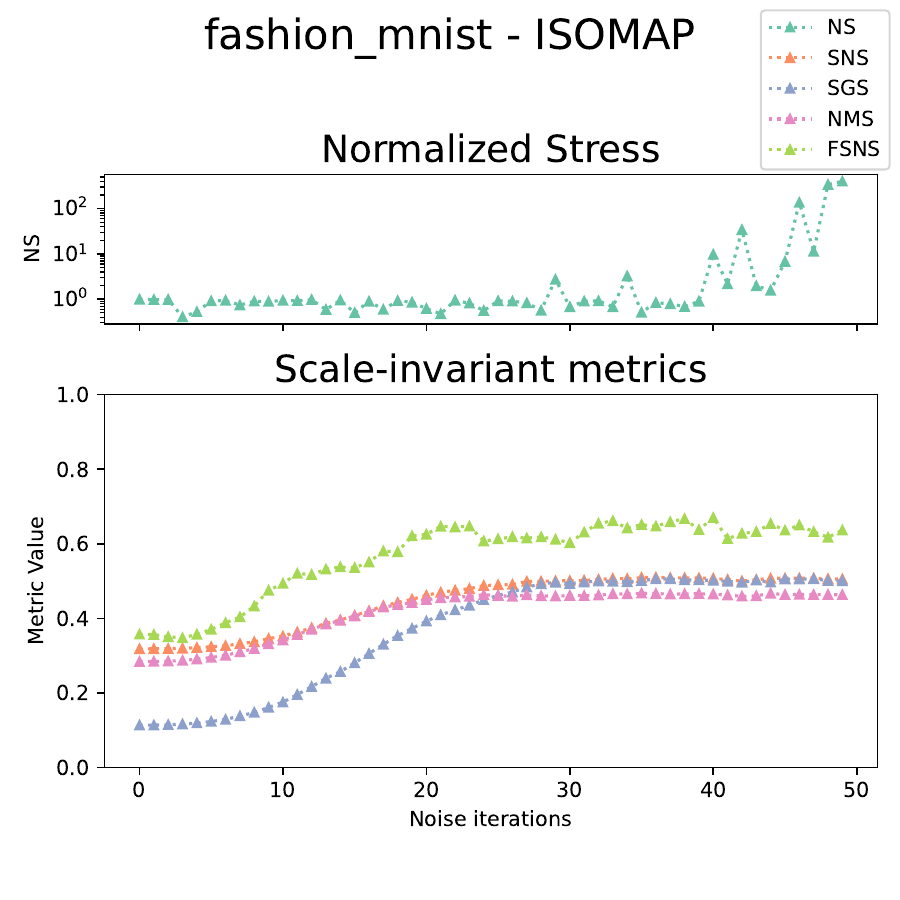}
    \includegraphics[width=0.32\linewidth]{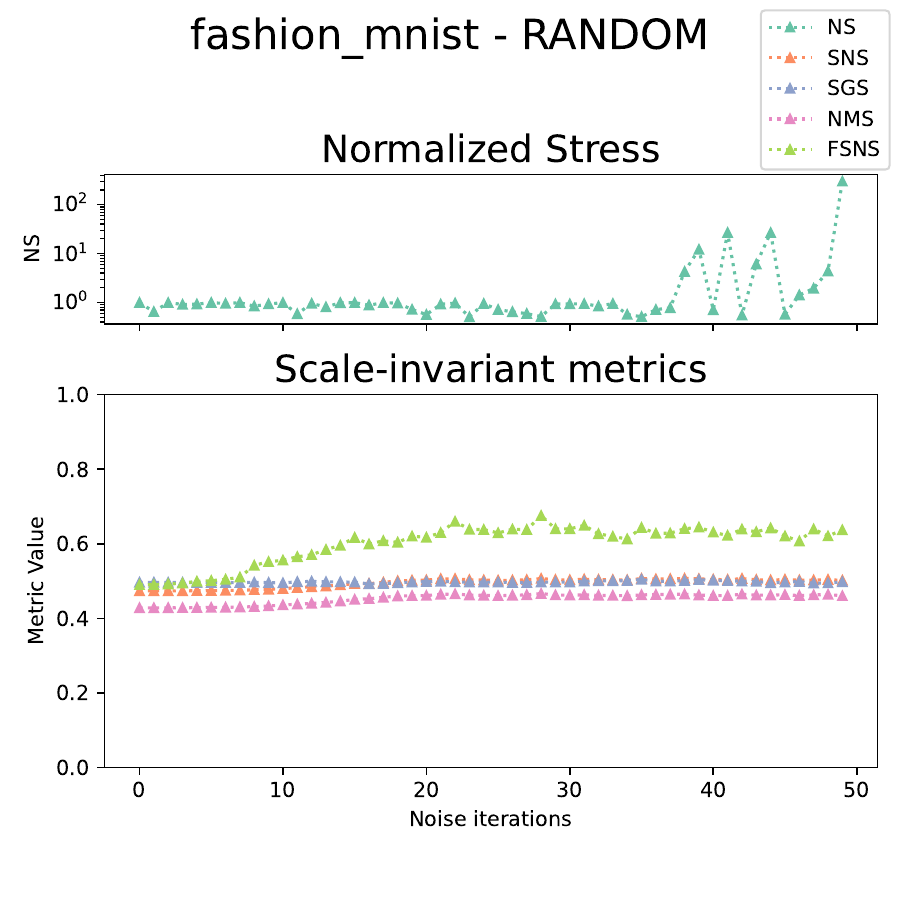}
    \caption{{Each small multiple shows the results of the sensitivity experiment for the Fashion-MNIST dataset across different algorithms. Normalized Stress is shown on top and all scale-invariant measures on the bottom. 
    From a given embedding, isotropic Gaussian noise is repeatedly applied, and at each iteration the embedding is rescaled by a random number between 0.1 and 10. The plots show how each metric behaves as the embedding quality degrades. Normalized stress is plotted separately and at a log scale as the magnitude of changes vary drastically. Note, the values of the metrics should not be compared, only the behavior of the curves. SGS is inverted so that lower is better. Further plots can be found in supplemental material.
    We observe that all scale-invariant metrics behave predictably (stress increases when more noise is added) while the scale-sensitive normalized stress is unpredictable and inconsistent.}}
    \label{fig:ladder-exp}
\end{figure*}

{
We seek first to address \textbf{Q1}: Which stress metrics behave consistently as the quality of an embedding worsens?
To this end, we begin with an embedding, $Y_0$, produced by a DR algorithm assumed to be of high quality. 
We then generate a series of perturbed embeddings $Y_1,\dots Y_{50}$ by successively adding isotropic Gaussian noise of increasing variance. Each perturbation step then yields an embedding of systematically lower quality. To simulate the fact that the absolute scale of DR outputs is typically arbitrary, we randomly rescale each embedding before evaluating each metric. 

Finally, we compare different stress measures across this sequence of embeddings. A metric is consistent if its values exhibit a monotonic or otherwise stable trend with respect to the embedding degradation, regardless of the applied random scaling. 

\noindent 
\textbf{Results}: 
From our results, the overall trend is clear; see Fig.~\ref{fig:ladder-correlation}. The scale-sensitive normalized stress measure is very inconsistent across datasets, often even causing some algorithms to seemingly ``improve'' as the embedding is perturbed. 
Meanwhile, the scale-invariant measures behave in a much more consistent fashion, with values steadily increasing as the embedding worsens (with the exception of the random embedding which, as one would expect, stays roughly constant). 

{
To quantify these trends, we compute the Pearson correlation for each of these individual curves on MDS embeddings, visualized in Fig.~\ref{fig:ladder-correlation-mds}. This correlation quantifies how often the metric tends to increase as the embedding is perturbed further. An ideal correlation value for a metric would be positive one, as this would indicate a perfect trend. However, one should expect some noise for different datasets and some random perturbations will be worse than others just by chance. Still, it should be clear from Fig.~\ref{fig:ladder-correlation-mds} that the scale-invariant measures often achieve correlations near one while normalized stress has significantly lower correlation.
This experiment illustrates the practical risks of relying on a scale-sensitive stress measure, as 
only the scale-invariant measures behave consistently as the embedding quality worsens. 
}

\begin{figure}
    \centering
    \includegraphics[width=0.75\linewidth,height=6cm]{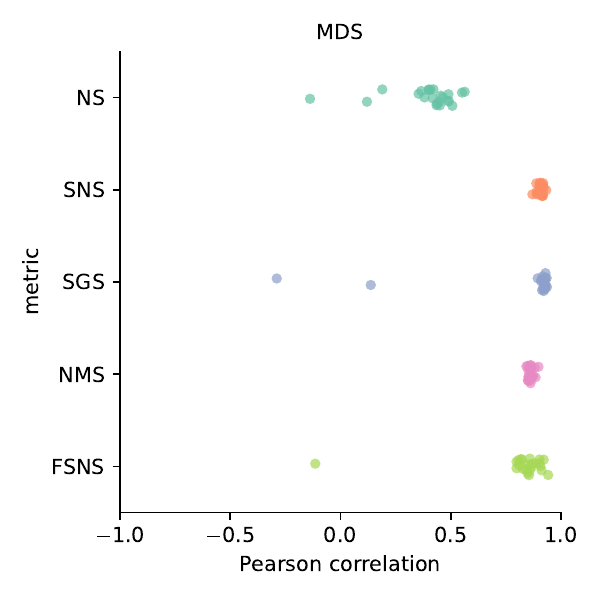}
    \caption{{Pearson correlation between a metric value and the number of noise iterations in the sensitivity experiment, for embeddings produced by MDS. Each point represents a different dataset and its $x$-coordinate is the correlation. A high positive correlation means that as an embedding becomes more noisy, the metric tends to increase while a high negative correlation indicates the opposite. A consistent metric, such as SNS, results in values near +1. }}
    \label{fig:ladder-correlation-mds}
\end{figure}

{
\subsection{Baseline performance}
We next 
address \textbf{Q2} (Is it possible that incorrect conclusions could be drawn by not accounting for scale?) with a second experiment. 
Here we produce ten embeddings for each algorithm-dataset combination to account for different initializations of MDS, t-SNE, UMap, and Random. Note that due to the deterministic, closed-form nature of PCA and Isomap, we have only one algorithm-dataset combination to work with.

We treat the MDS embeddings as a baseline, as MDS explicitly optimizes stress, and thus directly minimizes distance distortion. It should be rare for algorithms that do not explicitly optimize stress to achieve lower values than MDS on any stress-based metric. We quantify violations of this expected order by computing the percentage of cases where the stress of MDS exceeds that of the comparison algorithm; counting the number of times an alternative outperforms MDS under the given stress measure. 

To further highlight the potentially misleading nature of scale-sensitive measures, we repeat the evaluation after rescaling each embedding by a factor of 10. We report both sets of rankings in Table~\ref{tab:alg_less_mds_1x}.

\noindent
\textbf{Results}:
At the default scale, normalized stress generally agrees with the expected order, but notable violations occur. In particular, t-SNE and UMAP are occasionally judged to have lower stress than MDS, despite not optimizing distances directly. By contrast, the scale-invariant measures behave more consistently, rarely finding such reversals. 

The fragility of the normalized stress metric becomes most apparent when the embeddings are rescaled. Just resizing all embeddings by a factor of ten dramatically alters the conclusions: the random embedding, which contains no structure, is reported to outperform MDS in nearly 80\% of cases under normalized stress. Notably, none of the scale-invariant measures produce such contradictions.

We also note that while the order is never reversed between t-SNE and MDS, this could indeed happen.
The default output scale of t-SNE is such that by scaling down (rather than scaling up) we can make it seem like t-SNE outperforms MDS with normalized stress; see Fig.~\ref{fig:teaser}. We choose only one additional  scale to evaluate at (10 times larger than the default), but by considering different scales any embedding could appear to achieve lower normalized stress than MDS.

Taken together, these results demonstrate that normalized stress is an inappropriate measure for algorithm comparisons. While it may appear to function as intended at the default scale, even small changes undermine its interpretability as rescaling 
{shows.} In contrast, scale-invariant measures provide stable, consistent assessments of embedding quality.

}

\begin{table}[ht]
\centering
\caption{{ 
The percentage of trials in which a different algorithm outperforms MDS under the given stress metric at the default or 10 times the output scale. Note that while Normalized Stress (NS) performs as expected at the default scale, increasing the size of the embedding by 10 (changing nothing meaningful about the embedding) causes NS to show obviously incorrect results in most cases, especially exaggerated when evaluating the RANDOM embedding to be better than MDS in 80\% of trials. Meanwhile, the other scale-sensitive measures stay consistent.
}}
\begin{tabularx}{\linewidth}{|p{2.5cm}| X X X X X|}
\hline
& \small{NS} & \small{SGS} & \small{NMS} & \small{SNS} & \small{FSNS} \\ \hline
t-SNE$<$MDS   & \cellcolor[HTML]{E7F0FA} 12.5\% & \cellcolor[HTML]{EEF5FC} 6.7\%  & \cellcolor[HTML]{F7FBFF} 0.0\%  & \cellcolor[HTML]{FFFFFF} 0.4\%  & \cellcolor[HTML]{E7F1FA} 12.1\% \\ 
$10x$ t-SNE$<$MDS   & \cellcolor[HTML]{E7F0FA} 12.5\% & \cellcolor[HTML]{EEF5FC} 6.7\%  & \cellcolor[HTML]{F7FBFF} 0.0\%  & \cellcolor[HTML]{FFFFFF} 0.4\%  & \cellcolor[HTML]{E7F1FA} 12.1\% \\ \hline\hline  
UMAP$<$MDS    & \cellcolor[HTML]{E1EDF8} 16.7\% & \cellcolor[HTML]{F0F6FD} 5.4\%  & \cellcolor[HTML]{F7FBFF} 0.0\%  & \cellcolor[HTML]{F5FAFE} 1.2\%  & \cellcolor[HTML]{FFFFFF} 0.0\%  \\ 
$10x$ UMAP$<$MDS  & \cellcolor[HTML]{C6DBEF} 37.5\% & \cellcolor[HTML]{F0F6FD} 5.4\%  & \cellcolor[HTML]{F7FBFF} 0.0\%  & \cellcolor[HTML]{F5FAFE} 1.2\%  & \cellcolor[HTML]{FFFFFF} 0.0\%  \\ \hline\hline 
PCA$<$MDS     & \cellcolor[HTML]{F2F7FD} 4.2\%  & \cellcolor[HTML]{DCEAF6} 20.4\% & \cellcolor[HTML]{E1EDF8} 16.7\% & \cellcolor[HTML]{E4EFF9} 14.2\% & \cellcolor[HTML]{DCE9F6} 20.8\% \\ 
$10x$ PCA$<$MDS     & \cellcolor[HTML]{58A1CF} 83.3\% & \cellcolor[HTML]{DCEAF6} 20.4\% & \cellcolor[HTML]{E1EDF8} 16.7\% & \cellcolor[HTML]{E4EFF9} 14.2\% & \cellcolor[HTML]{DCE9F6} 20.8\% \\ \hline\hline
Isomap$<$MDS  & \cellcolor[HTML]{D6E6F4} 25.0\% & \cellcolor[HTML]{E7F0FA} 12.5\% & \cellcolor[HTML]{F7FBFF} 0.0\%  & \cellcolor[HTML]{F2F7FD} 4.2\%  & \cellcolor[HTML]{EEF5FC} 6.7\%  \\ 
$10x$ Isomap$<$MDS  & \cellcolor[HTML]{C6DBEF} 37.5\% & \cellcolor[HTML]{E7F0FA} 12.5\% & \cellcolor[HTML]{F7FBFF} 0.0\%  & \cellcolor[HTML]{F2F7FD} 4.2\%  & \cellcolor[HTML]{EEF5FC} 6.7\%  \\ \hline\hline
RND$<$MDS     & \cellcolor[HTML]{FFFFFF} 0.0\%  & \cellcolor[HTML]{FFFFFF} 0.0\%  & \cellcolor[HTML]{FFFFFF} 0.0\%  & \cellcolor[HTML]{FFFFFF} 0.0\%  & \cellcolor[HTML]{F2F7FD} 4.6\%  \\ 
$10x$ RND$<$MDS     & \cellcolor[HTML]{61A7D2} 79.6\% & \cellcolor[HTML]{F7FBFF} 0.0\%  & \cellcolor[HTML]{FFFFFF} 0.0\%  & \cellcolor[HTML]{FFFFFF} 0.0\%  & \cellcolor[HTML]{F2F7FD} 4.6\%  \\ \hline
\end{tabularx}
\label{tab:alg_less_mds_1x}
\end{table}

\subsection{Validating Prior Results}\label{sec:experiment-B}

{ It is also important to demonstrate how reevaluating prior studies using scale-invariant stress measures rather than the scale-sensitive normalized stress can invalidate some prior results. Here, we aim to address \textbf{Q3} by revisiting the data from the quantitative survey of DR methods by Espadoto et al.~\cite{DBLP:journals/tvcg/EspadotoMKHT21}.} 
Specifically, we take 5 precomputed embeddings (by MDS, t-SNE, LLE, ISOMAP, and UMAP) for 16  of the 18 
datasets made available by Espadoto et al. 
{
We cannot use two of the datasets, marked with stars in \autoref{tab:datasets}, as 90\% of the ORL dataset is null vectors, and the Spambase embeddings are not provided and so cannot be used in this experiment.
}

Each of these 16 embeddings 
{is} finely tuned with hyperparameters to achieve good results, and the embeddings are at different scales.
We then compute two metrics for each of these embeddings: the normalized stress used in the survey evaluation and the scale-normalized stress proposed in~\autoref{sec:SNS}.

\autoref{fig:ns-vs-sns-rankings}
compares the rankings produced by normalized stress {(NS)} and scale-normalized stress {(SNS)} for the techniques and datasets in Espadoto et al.~\cite{DBLP:journals/tvcg/EspadotoMKHT21}.
The first thing that stands out is that MDS is consistently ranked the best regardless of the metric,  which is consistent with the fact that MDS directly optimizes normalized stress. However, the order of the other four techniques changes considerably -- in fact
there is no dataset for which normalized stress {(NS)} agrees with scale-normalized stress {(SNS)}.
Another visualization of this data which highlights high level patterns can be found in supplemental material.

Normalized stress {(NS)} places ISO as second-best approximately 33\% of the time and LLE as second-best 40\% of the time. Meanwhile, the scale-normalized stress {(SNS)} metric ranks t-SNE as the second best 80\% of the time.
When considering the techniques that both metrics rank as the least effective, normalized stress {(NS)} ranks t-SNE last 80\% of the time, while scale-normalized stress {(SNS)} ranks LLE last approximately 73\% of the time.
This 
begins to answer \textbf{Q3} and
calls into question the validity of results based on normalized stress as an evaluation metric.

{Another important observation in ~\autoref{fig:ns-vs-sns-rankings} is that scale-normalized stress (SNS) never ranks UMAP above t-SNE. This contrasts with the commonly held belief that UMAP preserves global distances better than t-SNE, and is instead consistent with recent findings by van der Hoorn et al.~\cite{diehl2025reviewer}.
}

\begin{figure}
    \centering
    \includegraphics[width=\linewidth]{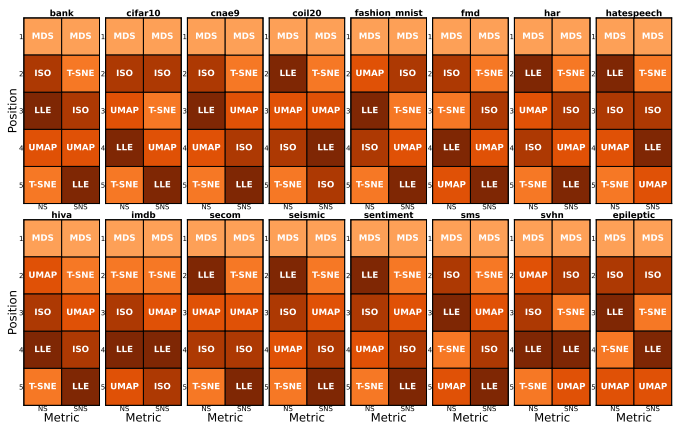}
    \caption{The difference in embedding (MDS, t-SNE, UMAP, LLE, and ISO) orders between normalized stress and scale-normalized stress for each dataset. The first column of each diagram depicts the order given by NS, while the second column depicts the order given by SNS.}
    \label{fig:ns-vs-sns-rankings}
\end{figure}

\section{KL Divergence Experiments}

We again 
demonstrate how often and how severely scale may affect a quality metric, but now looking only at the KL divergence instead of stress. Additionally, our experiments show that KL divergence provides a metric that matches one's expectations once scale is accounted for.


\subsection{Experiment Description}
We continue to use the same 24 (6 well-known, and 18 found in Espadoto et al.~\cite{DBLP:journals/tvcg/EspadotoMKHT21}) datasets and DR techniques described in \autoref{sec:stress-exp-desc} with the KL divergence metrics described in the previous section.
Each quality metric is implemented and available open source; see \autoref{sec:supplemental_materials}. The high-dimensional distribution $P$ is the same for a given dataset across all measures. We set the perplexity parameter to the value 30 (recommended by van der Maaten and Hinton~\cite{van2008visualizing}) for all datasets.


\subsection{Empirical Results: KL Experiment 2A}

We evaluate the projections with the various KL divergence metrics using the t-SNE, MDS, and RND algorithms.
We expect the order to be t-SNE$<$MDS$<$RND, since t-SNE directly optimizes KL divergence and we expect MDS to perform better than RND. \autoref{tab:pct-table-kl} clearly displays the scale-sensitive nature of the t-SNE-defined KL divergence. The orders change significantly when changing the scale from 1 ($KL_1$) to 10 ($KL_{10}$).

Among the scale-invariant metrics, 
SNKL performs the best, capturing the expected order over 96\% of the time, while FSKL does not perform as well (as expected since the scales chosen are not informed by the embedding).
In our experiments, SNKL only 
{gives} an unexpected ordering in
8 of the 10 runs of
the Penguins dataset, where we 
{obtain} the ordering of MDS$<$t-SNE$<$RND. Here, we notice that the difference in SNKL between MDS and t-SNE 
{is} relatively small:
MDS 
{achieves} an average score of 1.97 across the 10 runs, while t-SNE 
{obtains} a value of 2.01.


In ~\autoref{tab:corr-of-scores-kl}, we see the Spearman rank correlation between the scores given by each of our metrics. 
Here, we observe a high correlation between $KL$ at the default scale, and at 10 times the default scale. This can be explained by the fact that, unlike with stress which varies quadratically with scale, KL divergence varies by a much smaller amount. We also observe low correlation between FSKL and all other metrics, since FSKL produces incorrect orders where the others do not.

\begin{table}[ht]
\centering
\caption{Percentage of each possible ordering of the three projections (rows) given by each metric (columns). The first row (t-SNE, MDS, RND) is the expected local preservation ordering. Metrics with high values on this row most often capture this ordering. The columns of $KL_1$ and $KL_{10}$ refer to the Student's $t$ KL divergence at $\alpha=1$ and $\alpha=10$.}
\begin{tabularx}{\linewidth}{|p{2.65cm}| X X X X|}
\hline
& \small{SNKL} & \small{FSKL} & \small{$KL_1$} & \small{$KL_{10}$} \\ \hline
TSNE$<$MDS$<$RND & \cellcolor[HTML]{3d8dc4} 96.67\% & \cellcolor[HTML]{5aa2cf} 82.92\% & \cellcolor[HTML]{4695c8} 91.67\% & \cellcolor[HTML]{7db8da} 68.33\%  \\ \hline 

TSNE$<$RND$<$MDS & \cellcolor[HTML]{f7fbff} 0.00\% & \cellcolor[HTML]{ffffff} 0.00\% & \cellcolor[HTML]{ffffff} 0.00\% & \cellcolor[HTML]{f2f8fd} 3.75\%  \\ \hline 

MDS$<$TSNE$<$RND & \cellcolor[HTML]{f3f8fe} 3.33\% & \cellcolor[HTML]{e0ecf8} 17.08\% & \cellcolor[HTML]{ffffff} 0.00\% & \cellcolor[HTML]{e3eef8} 15.42\%   \\ \hline 

MDS$<$RND$<$TSNE & \cellcolor[HTML]{ffffff} 0.00\% & \cellcolor[HTML]{ffffff} 0.00\% & \cellcolor[HTML]{ecf4fb} 8.33\% & \cellcolor[HTML]{e7f0fa} 12.50\%   \\ \hline 

RND$<$TSNE$<$MDS & \cellcolor[HTML]{ffffff} 0.00\% & \cellcolor[HTML]{ffffff} 0.00\% & \cellcolor[HTML]{ffffff} 0.00\% & \cellcolor[HTML]{ffffff} 0.00\%  \\ \hline 

RND$<$MDS$<$TSNE & \cellcolor[HTML]{ffffff} 0.00\% & \cellcolor[HTML]{ffffff} 0.00\% & \cellcolor[HTML]{ffffff} 0.00\% & \cellcolor[HTML]{ffffff} 0.00\%  \\ \hline 
\end{tabularx}
\label{tab:pct-table-kl}
\end{table}

\begin{table}[ht]
\centering
\caption{Spearman rank correlation between the scores of the KL divergence metrics}
\begin{tabularx}{\linewidth}{|p{2cm}| X X X X |}
\hline
& \small{SNKL} & \small{FSKL} & \small{$KL_1$} & \small{$KL_{10}$} \\ \hline 
SNKL & \cellcolor[HTML]{77b5d9} 1.0 & \cellcolor[HTML]{e3eef9} 0.51 & \cellcolor[HTML]{8abfdd} 0.94 & \cellcolor[HTML]{8abfdd} 0.94\\ \hline 
FSKL & \cellcolor[HTML]{f7fbff} - & \cellcolor[HTML]{77b5d9} 1.0 & \cellcolor[HTML]{dfebf7} 0.54 & \cellcolor[HTML]{ffffff} 0.51\\ \hline 
$KL_1$ & \cellcolor[HTML]{ffffff} - & \cellcolor[HTML]{ffffff} - & \cellcolor[HTML]{77b5d9} 1.0 & \cellcolor[HTML]{9fcae1} 0.87\\ \hline 
$KL_{10}$ & \cellcolor[HTML]{ffffff} - & \cellcolor[HTML]{ffffff} - & \cellcolor[HTML]{ffffff} - & \cellcolor[HTML]{77b5d9} 1.0\\ \hline
\end{tabularx}
\newline
\label{tab:corr-of-scores-kl}
\end{table}

}

\section{Discussion}

We 
{show} the need to account for scale when using stress-based measures to validate DR results. Otherwise, the results can be misleading; for example, when using  normalized stress~\cite{DBLP:journals/tvcg/EspadotoMKHT21,DBLP:journals/tvcg/NonatoA19,DBLP:conf/igarss/ChenCG06}
, it appears as though a  random, uncorrelated plot of the data does a better job at distance preservation than t-SNE (and sometimes even MDS). Intuitively, this should not be true, and visually, it clearly is not the case; see~\autoref{fig:teaser}. 
Using a metric called ``normalized stress'' seems like a good idea when evaluating the quality of a new DR method. However, as we 
{show} in this paper, ``normalized stress'' is very similar to ``raw stress,'' and both of them are severely affected by scaling. 

{
The same holds for applying t-SNE's optimization function, KL divergence, as a quality metric. If scale is not taken into account, the results are at best unreliable and at worst  misleading. 
}

One should be careful when comparing the outputs from different DR methods, which often come at very different scales. 
The proposed scale-invariant metrics SNS and SNKL are not only more in line with the expected results; they are more consistent and cannot be easily manipulated (e.g., by selectively scaling some outputs).
{ 
We recommend SNS and SNKL over other scale-invariant options for their explainability and ease of implementation. 
}
{
}






\noindent \textbf{Limitations: }
While the experiments described here provide insights into the behavior of different 
metrics used to evaluate  DR techniques, there are many limitations.
We base our experimental validation on selecting a limited number of algorithms and 
a limited number of metrics.
The experiments 
{utilize} a fairly small number of datasets as well. While we believe this provides a reasonable base for comparison, 
including more DR techniques, more datasets, and more quality  metrics~\cite{DBLP:conf/vissym/SeifertSK10,ed0a397fae134f218fa4fd3998ced318,DBLP:journals/tvcg/KorenC04} would increase the robustness of our findings.

In 
our baseline performance experiments, we 
{operate} under the expectation that there is a ``correct order'' when considering how well  DR techniques preserve qualities like distances or neighbors in their embeddings. However plausible the hypothesis that optimized embeddings should score better than 
non-optimized embeddings, a better-grounded experiment 
{can} verify the effectiveness of quality metrics.
Finally,
our experiments
do not account for potential variations in the performance of DR techniques due to hyperparameter settings or initialization conditions, which can influence the quality of DR techniques~\cite{DBLP:conf/vluds/EngelHH11}.

{ We also note that while the KL divergence of t-SNE can be used as a quality metric, it should only be done in select circumstances. It is best used as a diagnostic tool for neighbor embedding techniques, and should not be used to compare algorithms with different similarity models without care. Its primary value in this work is to illustrate how even a widely used objective function is subject to the same scale-sensitivity as stress.}

\noindent \textbf{Conclusion: }
In this paper, we demonstrate the significant impact of scaling for DR measures such as stress and KL divergence. The empirical evidence indicates that the scale-sensitivity problem affects real-world datasets, and taking scale into account may alter previous evaluations.
To address this, we propose scale-invariant quality metrics.



\section*{Supplemental Materials}
\label{sec:supplemental_materials}
Below are links to supplemental materials available online.

\footnotemark[1]{\href{https://kiransmelser.github.io/normalized-stress-is-not-normalized}{\texttt{kiransmelser\discretionary{}{.}{.}github\discretionary{}{.}{.}io\discretionary{/}{}{/}normalized\discretionary{}{-}{-}stress\discretionary{}{-}{-}is\discretionary{}{-}{-}not\discretionary{}{-}{-}normalized}}} \label{sup-1}

\footnotemark[2]{\href{https://github.com/KiranSmelser/stress-metrics-and-scale}{\texttt{kiransmelser\discretionary{}{.}{.}github\discretionary{}{.}{.}io\discretionary{/}{}{/}stress\discretionary{}{-}{-}metrics\discretionary{}{-}{-}and\discretionary{}{-}{-}scale}}} \label{sup-2}


\bibliographystyle{abbrv-doi-hyperref-narrow}

\bibliography{references}


\appendix

\section*{Supplemental Material}

 We provide more information about the behavior of KL divergence and give more examples of how both normalized stress and KL divergence behave with respect to scale. \autoref{appendix:A} further analytically describes the behavior of KL divergence as the scale varies. \autoref{appendix:B} gives a simple example where KL divergence has a globabl maximum at $\alpha = 0$, and monotonically decreases to the right. 
\autoref{appendix:C} gives more context to the Gaussian kernel variant of KL divergence. \autoref{appendix:D} describes the pair of tables found in \autoref{tab:rerun-table-NS}, illustrating further how much the rankings change between normalized and scale-normalized stress. 
\autoref{appendix:E} gives a brief introduction to the many subsequent plots illlustrating how normalized stress and KL divergence behave as scale changes on many datasets. 
{\color{blue}

}

There are also supplemental material available online. The first is a web application which contains embeddings and curves (in the style of \autoref{fig:supplemental-orl} for all datasets, found here: \href{https://kiransmelser.github.io/normalized-stress-is-not-normalized}{\texttt{kiransmelser\discretionary{}{.}{.}github\discretionary{}{.}{.}io\discretionary{/}{}{/}normalized\discretionary{}{-}{-}stress\discretionary{}{-}{-}is\discretionary{}{-}{-}not\discretionary{}{-}{-}normalized}}. This is intended as a ``proof by example'' of why accounting for scale is important for quality metrics. Secondly, our repository contains all code to rerun the experiments described in the paper from scratch: pulling datasets from the web, computing embeddings, and computing metrics. This is found at the following link: \href{https://github.com/KiranSmelser/stress-metrics-and-scale}{\texttt{kiransmelser\discretionary{}{.}{.}github\discretionary{}{.}{.}io\discretionary{/}{}{/}stress\discretionary{}{-}{-}metrics\discretionary{}{-}{-}and\discretionary{}{-}{-}scale}}. Our normalized stress implementation can also be found in the Zadu Python library~\cite{DBLP:conf/visualization/JeonCJLHKJS23}.

\section{KL Divergence}
\label{appendix:A}
Here we include additional derivations for the behavior of KL divergence with respect to scale. 
For the following sections, let {$d_{i,j} = d(x_i,x_j)$ }serve as a shorthand for the high-dimensional input distances.

\subsection{Derivatives}
We first derive the derivatives of KL divergence with respect to the scaling parameter $\alpha$. 
The first derivative is given by the following formula:

\begin{equation}
    \label{eq:kl-first-derivative}
     \sum_{i \neq j} p_{ij} \times 
     \left(
         \dfrac{2 \alpha d_{ij}^2}{(1 + \alpha^2 d_{ij}^2)}
         - \dfrac
         { \sum_{k \neq l} \dfrac{2 \alpha d_{kl}^2}{(1 + \alpha^2  d_{kl}^2)^{2}} }
         {\sum_{k \neq l} (1 + \alpha^2 d_{ij}^2)^{-1}}
    \right)
\end{equation}
Notice that both numerators in the bracket become $0$ when $\alpha = 0$. 
The second derivative is unfortunately no less cluttered:

\begin{multline}
    \label{eq:kl-second-derivative}
    \sum_{i \neq j} p_{ij} \times 
    \left[
        \dfrac{2d_{ij}^2}{1 + \alpha^2 d_{ij}^2}
        - \dfrac{(2 \alpha d_{ij}^2)^2}{(1 + \alpha^2 d_{ij}^2)^2}\right. \\
        \left.
        - \dfrac{1}{\sum_{k \neq l} (1 + \alpha^2 d_{ij}^2)^{-1}}
        \times \sum_{k \neq l}
            \left( 
                \dfrac{2 d_{kl}^2}{(1 + \alpha^2 d_{ij}^2)^{2}}
                - \dfrac{2 (2 \alpha d_{kl}^2)^2}{(1 + \alpha^2 d_{ij}^2)^{3}}
            \right) \right.              \\
        \left. - \left(
            \sum_{k \neq l} \dfrac{2 \alpha d_{kl}^2}{(1 + \alpha^2 d_{kl}^2)^2}
          \right)^2
        \times \dfrac{1}{\left( \sum_{k \neq l} (1 + \alpha^2 d_{ij}^2)^{-1} \right)^2}
    \right]
\end{multline}


\subsection{Behavior at scale 0}

The KL divergence as a function of $\alpha$ starts at zero at a constant point that only depends on the high-dimensional dataset. By substituting $\alpha = 0$ into the KL divergence, we get the following value at zero:

\[
    \sum_{i \neq j} p_{ij} \times \left( \log p_{ij} + \log N (N - 1) \right)
\]

It can be seen from \autoref{eq:kl-first-derivative} that the first derivative at $\alpha = 0$ has the value $0$. Therefore, we expect to see an extremum at zero. In order to see whether we have a local minimum or a local maximum, we look at the second derivative, which can be simplified to:

\begin{multline}
    \label{eq:kl_2nd_diff_at_0}
    \sum_{i \neq j} 2 p_{ij} \times \left[ d_{ij}^2 - \dfrac{1}{N (N-1)} \sum_{k \neq l} d_{kl}^2 \right]
    = \sum_{i \neq j} 2 p_{ij} \times \left( d_{ij}^2 - \overline{d^2}\right)
\end{multline}

Here, $\overline{d^2}$ represents the mean {squared} pairwise distance between points in the embedding.

For our purposes, we can think of \autoref{eq:kl_2nd_diff_at_0} as a weighted average of the mean deviation of the squared distances, where the weights are given by the probabilities $p_{ij}$. Loosely speaking, we would expect \autoref{eq:kl_2nd_diff_at_0} to be positive (and therefore for us to have a minimum at 0) when high $p_{ij}$ values correspond to high $d_{ij}^2$ (i.e., where $\left( d_{ij}^2 - \overline{d^2}\right)$ would be positive). If the opposite is true, then we're giving higher weights to instances where $\left( d_{ij}^2 - \overline{d^2}\right)$ is negative, and therefore the second derivative will be negative (local maximum).

But $p_{ij}$ represents a similarity measure, so if high $p_{ij}$s correspond to high $d_{ij}^2$s, that would mean that our embedding is quite terrible! Therefore, in broad strokes, we would only expect to see a minimum at 0 for embeddings generated by extremely poor algorithms, and for algorithms that are actually performant and therefore worthy of comparing through this metric at all, we would see a maximum at 0.

We 
\textcolor{blue}{observe} experimentally that if there is a local minimum at 0, then this is also the global minimum; KL divergence increases monotonically from 0.

\subsection{Behavior as scale tends to infinity}

Here, let us look at the value the individual $q_{ij}(\alpha)$ probabilities take at infinity.

{
First, we reformulate $q_{ij}(\alpha)$ as follows:
\begin{multline*}
    q_{ij}(\alpha)
    = \dfrac{(1 + \alpha^2 d_{ij}^2)^{-1}}{\sum_{k \neq l} (1 + \alpha^2 d_{kl}^2)^{-1}}\\
    = \dfrac{\alpha^2(\alpha^{-2} + d_{ij}^2)^{-1}}{\sum_{k \neq l} \alpha^2 (\alpha^{-2} +  d_{kl}^2 )^{-1}}\\
    = \dfrac{(\alpha^{-2} + d_{ij}^2)^{-1}}{\sum_{k \neq l} (\alpha^{-2} +  d_{kl}^2 )^{-1}}  
\end{multline*}

It is now easy to see that the limit of $q_{ij}(\alpha)$ (and therefore $Q(\alpha)$) approaches a finite value as $\alpha$ approaches infinity:

\begin{equation}
    \label{eq:q_ij-at-infty}
    q_{ij}(\infty)
    = \lim_{\alpha \to \infty} q_{ij}(\alpha)
    = \dfrac{(\alpha^{-2} + d_{ij}^2)^{-1}}{\sum_{k \neq l} (\alpha^{-2} +  d_{kl}^2 )^{-1}}  
    = \dfrac{d_{ij}^{-2}}{\sum_{k \neq l} d_{kl}^{-2}}
\end{equation}


}

Therefore, since all other variables in the KL divergence are constant with respect to $\alpha$ (\autoref{eq:kl-div-short}), we can conclude that KL divergence approaches a horizontal asymptote that is particular to the embedding (unlike stress, which grows arbitrarily with $\alpha$).
\newline



\noindent \textbf{KL Divergence at Infinity ($KL_\infty$)}
\label{sec:kl-inf}

{

If we look closely at \autoref{eq:q_ij-at-infty}, we observe that taking $Q$ at scale $\infty$ is equivalent to replacing the Student's $t$ kernel with the inverse square kernel. Thus it is possible to define a new metric, $KL_{\infty}$, which has a inverse square kernel in $Q$ instead of Student's $t$ kernel (and is guaranteed to be scale-invariant since it is equivalent to choosing a specific scale in the original t-SNE KL divergence):

\begin{equation*}
    KL_\infty(X,Y) = KL(P, Q_{(\infty)})
\end{equation*}

However, in our experiments we 
\textcolor{blue}{find} that this metric does not perform well. A probable reason for this is that much like FSKL (\autoref{sec:fskl}), we are again arbitrarily choosing a scale uninformed by the embedding.
}



It turns out that taking this limit is equivalent to replacing the Student's $t$ kernel in the low-dimensional probability distribution by an inverse square kernel. That is, $q_{ij}$ is redefined as follows:
\begin{equation*}
    \label{eq:q_ij-invsq-kernel}
    q_{ij}
    = \dfrac{||x_i - x_j||^{-2}}{\sum_{k \neq l} ||x_k - x_l||^{-2}}
\end{equation*}


It should be noted that, unlike the Student's $t$ kernel, the inverse square kernel is subject to floating point error for small input distances, so in practice we use a fixed small value when necessary.

\section{Approaching Asymptote from Below}
\label{appendix:C}

While KL divergence as a function of scale behaved as in \autoref{fig:kl-wine} in all our experiments, we observe that this does not always have to be the case. While KL divergence generally approaches its horizontal asymptote from below, we construct the following example where KL divergence approaches its horizontal asymptote from above.

We consider a dataset of 3 samples, whose high-dimensional probability distribution we set as follows:

\[
P = \begin{bmatrix}
0 & 0.2 & 0.1 \\
0.2 & 0 & 0.2 \\
0.1 & 0.2 & 0
\end{bmatrix}
\]

For this dataset, we create the following embedding:

\[
Y = \left\{ 
\begin{bmatrix} 0 \\ 0 \end{bmatrix}, 
\begin{bmatrix} 0 \\ 1 \end{bmatrix}, 
\begin{bmatrix} 1 \\ 1 \end{bmatrix} 
\right\}
\]

The following graph for the corresponding KL divergence function clearly depicts KL divergence approaching its horizontal asymptote of 0 from above. (That this asymptote is indeed 0 can be verified by calculating $KL_\infty$ for the above embedding.)

\includegraphics[width=\linewidth]{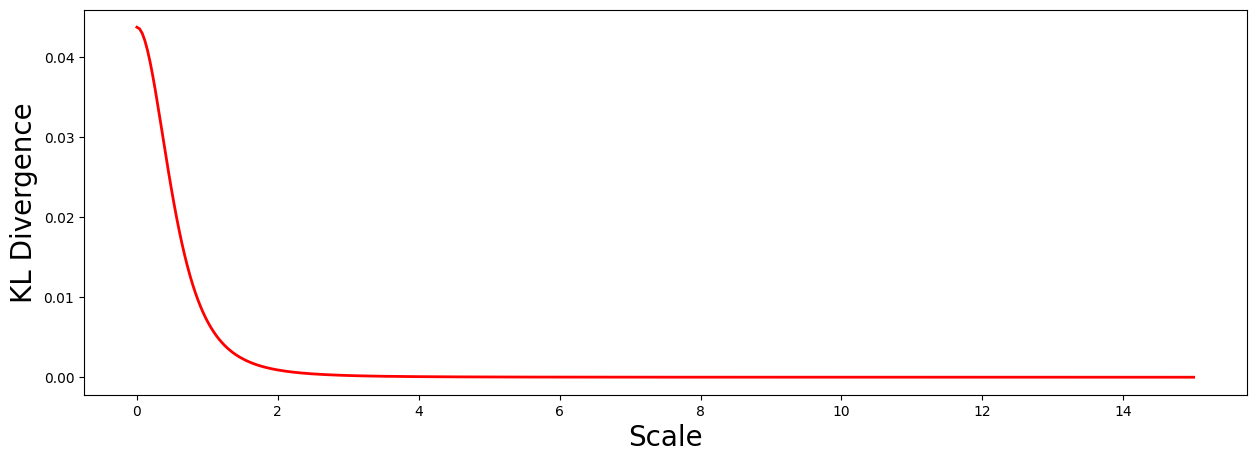}

\section{Gaussian}
\label{appendix:B}

{
This section explores the mathematical characteristics of KL divergence with a Gaussian kernel in $Q$ as a function of scale.}



{

\subsection{Behavior at scale 0}

At $\alpha = 0$, the distances in the low-dimensional space are effectively scaled to 0, which implies:
\begin{equation*}
    q_{ij}^G(0) = \frac{\exp(0)}{\sum_{k \neq l} \exp(0)} = \frac{1}{N(N-1)},
\end{equation*}
where $N$ is the total number of points. This is the same point that t-SNE's KL divergence starts at.

\subsection{Behavior as scale tends to infinity}

First, we introduce the following term:
\begin{equation*}
    d = \min\limits_{i, j} ||y_i - y_j ||^2
\end{equation*}

Now, we rewrite $q_{ij}^{G}(\alpha)$ as follows:
\begin{equation*}
    q_{ij}^{G}(\alpha) = \frac{\exp (-\alpha^2 ||y_i - y_j||^2)}{\sum_{k\neq l} \exp (-\alpha^2 ||y_k - y_l||^2)}
    = \frac{\exp (\alpha^2 (d - ||y_i - y_j||^2))}{\sum_{k\neq l} \exp (\alpha^2 (d - ||y_k - y_l||^2))}
\end{equation*}

All terms $\exp (\alpha^2 (d - ||y_k - y_l||^2))$ approach either zero as $\alpha \to \infty$ if $d < ||y_k - y_l||^2$, and 1 otherwise. Therefore, the limit at infinity looks as follows:

\begin{equation*}
    q_{ij}(\infty)
    = \lim_{\alpha \to \infty} q_{ij}(\alpha)
    = 
    \begin{cases}
        1 / n\_min, & \text{if} ||y_i - y_j||^2 = d  \\
        0, & \text{otherwise}
    \end{cases}
\end{equation*}
Here, $n\_min$ is the number of pairs $i, j$ where $||y_i - y_j||^2 = d$.

}

\section{Change in Ranking}
\label{appendix:D}

In \autoref{sec:experiment-B}, we demonstrate that the rankings with regards to stress 
\textcolor{blue}{change} for every embedding in the Espadoto et al.~\cite{DBLP:journals/tvcg/EspadotoMKHT21} evaluation when using scale-normalized stress instead of the standard normalized stress. We show in \autoref{tab:rerun-table-NS} an accompanying figure. While \autoref{fig:ns-vs-sns-rankings} shows that the rankings change somewhat for every dataset, the table here demonstrates that the global behavior also shifts. Most notably, when using scale-normalized stress the ranking of t-SNE improves from 5 to 2 on many datasets. Conventional wisdom in the field would state that t-SNE is poor at preserving global structure, but here we see it actually performs relatively well compared to its competitors, with carefully tuned hyperparameters.

\begin{table*}[ht]
    \centering
    \caption{The order given by normalized stress (left)  and scale-normalized stress (right) on embeddings of datasets (rows). Each embedding (MDS, t-SNE, UMAP, LLE, and ISO) is according to metric value, with lower ranks having lower stress. We see a change in overall pattern between the two rankings. Notably, t-SNE improves from mostly rank 5 to mostly rank 2.}
    \small
    
\begin{minipage}{0.48\linewidth}
\centering

    \begin{tabularx}{\linewidth}{|p{1.6cm}| X X X X p{.9cm}|}
    \multicolumn{6}{c}{\textbf{NS Ranking}}\\
    \hline 
& MDS & ISO & LLE & UMAP & T-SNE\\ \hline
bank & \cellcolor[HTML]{FFFFFF} 1 & \cellcolor[HTML]{D2EBF5} 2 & \cellcolor[HTML]{A5D6EA} 3 & \cellcolor[HTML]{78C2E0} 4 & \cellcolor[HTML]{4BAED6} 5\\ \hline
cifar10 & \cellcolor[HTML]{FFFFFF} 1 & \cellcolor[HTML]{D2EBF5} 2 & \cellcolor[HTML]{78C2E0} 4 & \cellcolor[HTML]{A5D6EA} 3 & \cellcolor[HTML]{4BAED6} 5\\ \hline
cnae9 & \cellcolor[HTML]{FFFFFF} 1 & \cellcolor[HTML]{D2EBF5} 2 & \cellcolor[HTML]{A5D6EA} 3 & \cellcolor[HTML]{78C2E0} 4 & \cellcolor[HTML]{4BAED6} 5\\ \hline
coil20 & \cellcolor[HTML]{FFFFFF} 1 & \cellcolor[HTML]{78C2E0} 4 & \cellcolor[HTML]{D2EBF5} 2 & \cellcolor[HTML]{A5D6EA} 3 & \cellcolor[HTML]{4BAED6} 5\\ \hline
epileptic & \cellcolor[HTML]{FFFFFF} 1 & \cellcolor[HTML]{D2EBF5} 2 & \cellcolor[HTML]{A5D6EA} 3 & \cellcolor[HTML]{4BAED6} 5 & \cellcolor[HTML]{78C2E0} 4\\ \hline
fashionmnist & \cellcolor[HTML]{FFFFFF} 1 & \cellcolor[HTML]{78C2E0} 4 & \cellcolor[HTML]{A5D6EA} 3 & \cellcolor[HTML]{D2EBF5} 2 & \cellcolor[HTML]{4BAED6} 5\\ \hline
fmd & \cellcolor[HTML]{FFFFFF} 1 & \cellcolor[HTML]{D2EBF5} 2 & \cellcolor[HTML]{78C2E0} 4 & \cellcolor[HTML]{4BAED6} 5 & \cellcolor[HTML]{A5D6EA} 3\\ \hline
har & \cellcolor[HTML]{FFFFFF} 1 & \cellcolor[HTML]{78C2E0} 4 & \cellcolor[HTML]{D2EBF5} 2 & \cellcolor[HTML]{A5D6EA} 3 & \cellcolor[HTML]{4BAED6} 5\\ \hline
hatespeech & \cellcolor[HTML]{FFFFFF} 1 & \cellcolor[HTML]{A5D6EA} 3 & \cellcolor[HTML]{D2EBF5} 2 & \cellcolor[HTML]{78C2E0} 4 & \cellcolor[HTML]{4BAED6} 5\\ \hline
hiva & \cellcolor[HTML]{FFFFFF} 1 & \cellcolor[HTML]{A5D6EA} 3 & \cellcolor[HTML]{78C2E0} 4 & \cellcolor[HTML]{D2EBF5} 2 & \cellcolor[HTML]{4BAED6} 5\\ \hline
imdb & \cellcolor[HTML]{FFFFFF} 1 & \cellcolor[HTML]{A5D6EA} 3 & \cellcolor[HTML]{78C2E0} 4 & \cellcolor[HTML]{4BAED6} 5 & \cellcolor[HTML]{D2EBF5} 2\\ \hline
secom & \cellcolor[HTML]{FFFFFF} 1 & \cellcolor[HTML]{78C2E0} 4 & \cellcolor[HTML]{D2EBF5} 2 & \cellcolor[HTML]{A5D6EA} 3 & \cellcolor[HTML]{4BAED6} 5\\ \hline
seismic & \cellcolor[HTML]{FFFFFF} 1 & \cellcolor[HTML]{A5D6EA} 3 & \cellcolor[HTML]{D2EBF5} 2 & \cellcolor[HTML]{78C2E0} 4 & \cellcolor[HTML]{4BAED6} 5\\ \hline
sentiment & \cellcolor[HTML]{FFFFFF} 1 & \cellcolor[HTML]{A5D6EA} 3 & \cellcolor[HTML]{D2EBF5} 2 & \cellcolor[HTML]{78C2E0} 4 & \cellcolor[HTML]{4BAED6} 5\\ \hline
sms & \cellcolor[HTML]{FFFFFF} 1 & \cellcolor[HTML]{D2EBF5} 2 & \cellcolor[HTML]{A5D6EA} 3 & \cellcolor[HTML]{4BAED6} 5 & \cellcolor[HTML]{78C2E0} 4\\ \hline
svhn & \cellcolor[HTML]{FFFFFF} 1 & \cellcolor[HTML]{A5D6EA} 3 & \cellcolor[HTML]{78C2E0} 4 & \cellcolor[HTML]{D2EBF5} 2 & \cellcolor[HTML]{4BAED6} 5\\ \hline
\end{tabularx}

\end{minipage}%
\hfill
\begin{minipage}{0.49\linewidth}
\centering

\begin{tabularx}{\linewidth}{|p{1.6cm}| X X X X p{.9cm}|}
    \multicolumn{6}{c}{\textbf{SNS Ranking}}\\    
    \hline 
& MDS & ISO & LLE & UMAP & T-SNE\\ \hline
bank & \cellcolor[HTML]{FFFFFF} 1 & \cellcolor[HTML]{A5D6EA} 3 & \cellcolor[HTML]{4BAED6} 5 & \cellcolor[HTML]{78C2E0} 4 & \cellcolor[HTML]{D2EBF5} 2\\ \hline
cifar10 & \cellcolor[HTML]{FFFFFF} 1 & \cellcolor[HTML]{D2EBF5} 2 & \cellcolor[HTML]{4BAED6} 5 & \cellcolor[HTML]{78C2E0} 4 & \cellcolor[HTML]{A5D6EA} 3\\ \hline
cnae9 & \cellcolor[HTML]{FFFFFF} 1 & \cellcolor[HTML]{78C2E0} 4 & \cellcolor[HTML]{4BAED6} 5 & \cellcolor[HTML]{A5D6EA} 3 & \cellcolor[HTML]{D2EBF5} 2\\ \hline
coil20 & \cellcolor[HTML]{FFFFFF} 1 & \cellcolor[HTML]{4BAED6} 5 & \cellcolor[HTML]{78C2E0} 4 & \cellcolor[HTML]{A5D6EA} 3 & \cellcolor[HTML]{D2EBF5} 2\\ \hline
epileptic & \cellcolor[HTML]{FFFFFF} 1 & \cellcolor[HTML]{D2EBF5} 2 & \cellcolor[HTML]{78C2E0} 4 & \cellcolor[HTML]{4BAED6} 5 & \cellcolor[HTML]{A5D6EA} 3\\ \hline
fashionmnist & \cellcolor[HTML]{FFFFFF} 1 & \cellcolor[HTML]{D2EBF5} 2 & \cellcolor[HTML]{4BAED6} 5 & \cellcolor[HTML]{78C2E0} 4 & \cellcolor[HTML]{A5D6EA} 3\\ \hline
fmd & \cellcolor[HTML]{FFFFFF} 1 & \cellcolor[HTML]{A5D6EA} 3 & \cellcolor[HTML]{4BAED6} 5 & \cellcolor[HTML]{78C2E0} 4 & \cellcolor[HTML]{D2EBF5} 2\\ \hline
har & \cellcolor[HTML]{FFFFFF} 1 & \cellcolor[HTML]{A5D6EA} 3 & \cellcolor[HTML]{4BAED6} 5 & \cellcolor[HTML]{78C2E0} 4 & \cellcolor[HTML]{D2EBF5} 2\\ \hline
hatespeech & \cellcolor[HTML]{FFFFFF} 1 & \cellcolor[HTML]{A5D6EA} 3 & \cellcolor[HTML]{78C2E0} 4 & \cellcolor[HTML]{4BAED6} 5 & \cellcolor[HTML]{D2EBF5} 2\\ \hline
hiva & \cellcolor[HTML]{FFFFFF} 1 & \cellcolor[HTML]{78C2E0} 4 & \cellcolor[HTML]{4BAED6} 5 & \cellcolor[HTML]{A5D6EA} 3 & \cellcolor[HTML]{D2EBF5} 2\\ \hline
imdb & \cellcolor[HTML]{FFFFFF} 1 & \cellcolor[HTML]{4BAED6} 5 & \cellcolor[HTML]{78C2E0} 4 & \cellcolor[HTML]{A5D6EA} 3 & \cellcolor[HTML]{D2EBF5} 2\\ \hline
secom & \cellcolor[HTML]{FFFFFF} 1 & \cellcolor[HTML]{78C2E0} 4 & \cellcolor[HTML]{4BAED6} 5 & \cellcolor[HTML]{A5D6EA} 3 & \cellcolor[HTML]{D2EBF5} 2\\ \hline
seismic & \cellcolor[HTML]{FFFFFF} 1 & \cellcolor[HTML]{78C2E0} 4 & \cellcolor[HTML]{4BAED6} 5 & \cellcolor[HTML]{A5D6EA} 3 & \cellcolor[HTML]{D2EBF5} 2\\ \hline
sentiment & \cellcolor[HTML]{FFFFFF} 1 & \cellcolor[HTML]{78C2E0} 4 & \cellcolor[HTML]{4BAED6} 5 & \cellcolor[HTML]{A5D6EA} 3 & \cellcolor[HTML]{D2EBF5} 2\\ \hline
sms & \cellcolor[HTML]{FFFFFF} 1 & \cellcolor[HTML]{78C2E0} 4 & \cellcolor[HTML]{4BAED6} 5 & \cellcolor[HTML]{A5D6EA} 3 & \cellcolor[HTML]{D2EBF5} 2\\ \hline
svhn & \cellcolor[HTML]{FFFFFF} 1 & \cellcolor[HTML]{D2EBF5} 2 & \cellcolor[HTML]{78C2E0} 4 & \cellcolor[HTML]{4BAED6} 5 & \cellcolor[HTML]{A5D6EA} 3\\ \hline
\end{tabularx}

\end{minipage}

    \label{tab:rerun-table-NS}
    \vspace{-0.5cm}
\end{table*}

\section{Extra images}
\label{appendix:E}
We show a representative example of the effect of scale on datasets in our benchmark here. All examples can be found in our open source repository. Note that the absolute values of the small-multiple axes are not comparable. 

{
Additionally, we show additional experimental data from the sensitivity experiment.
}

\begin{figure*}[ht]{
 \centering

 \includegraphics[width=.49\linewidth]{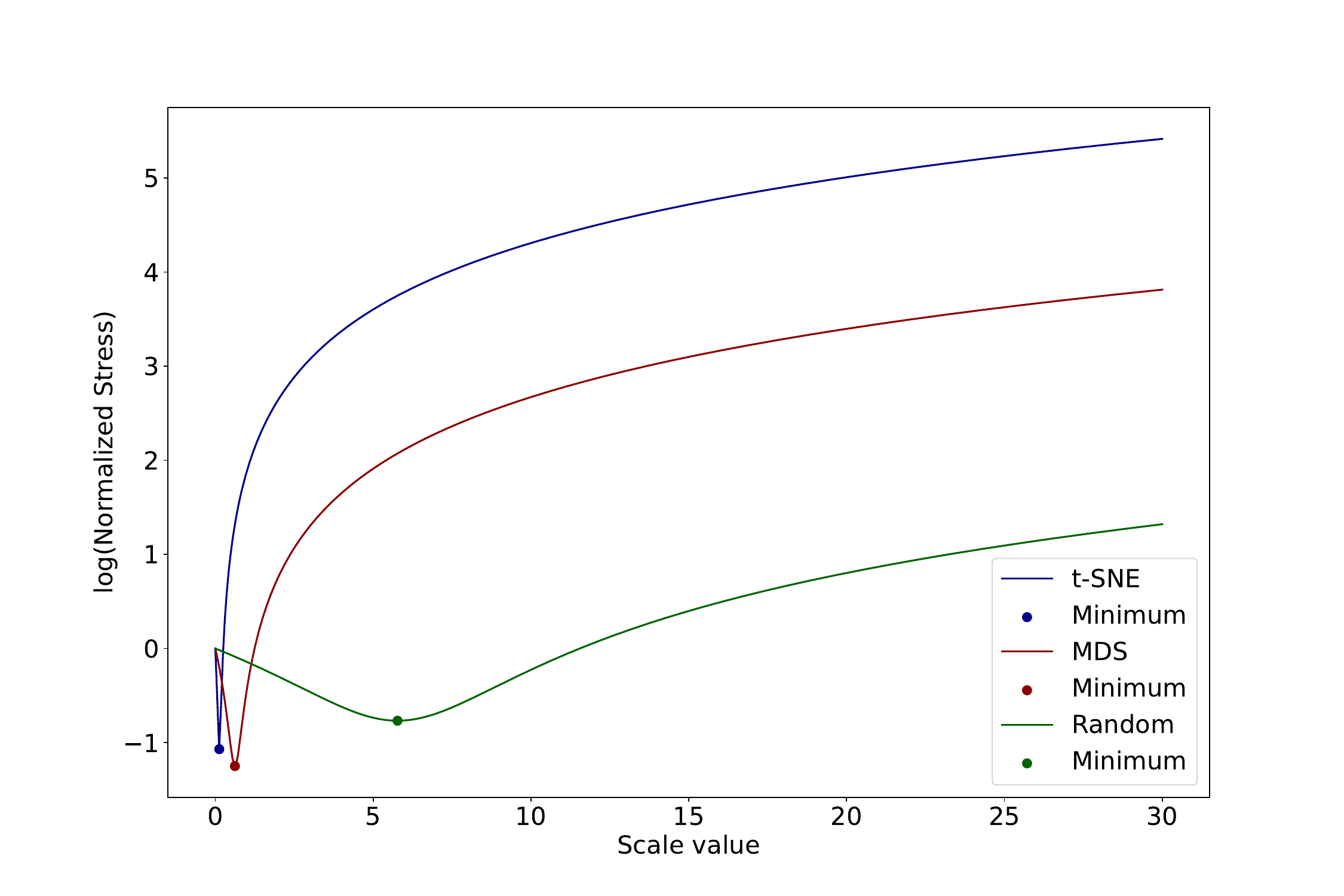}
 \includegraphics[width=0.49\linewidth]{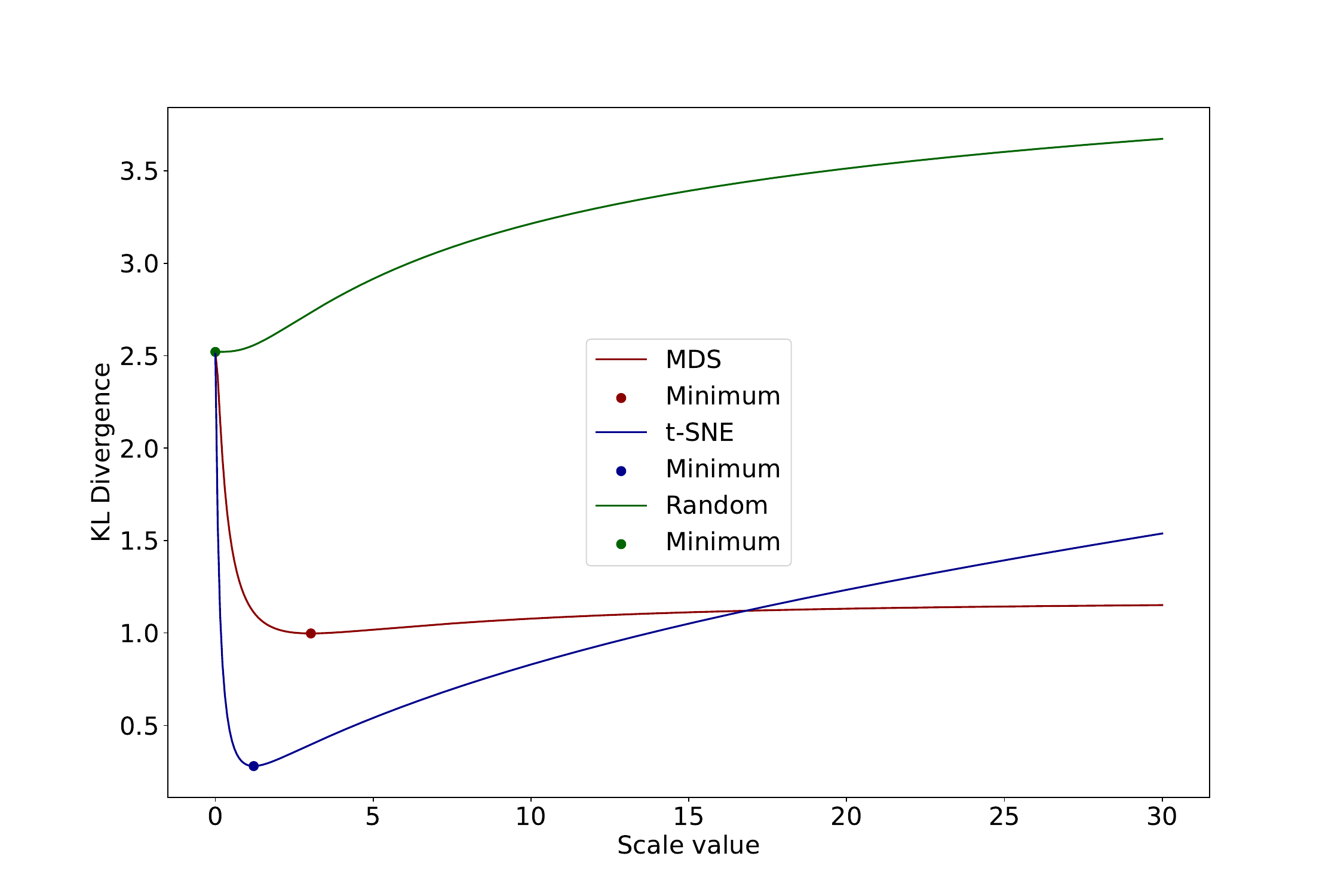}

 \includegraphics[width=0.32\linewidth,height=5cm]{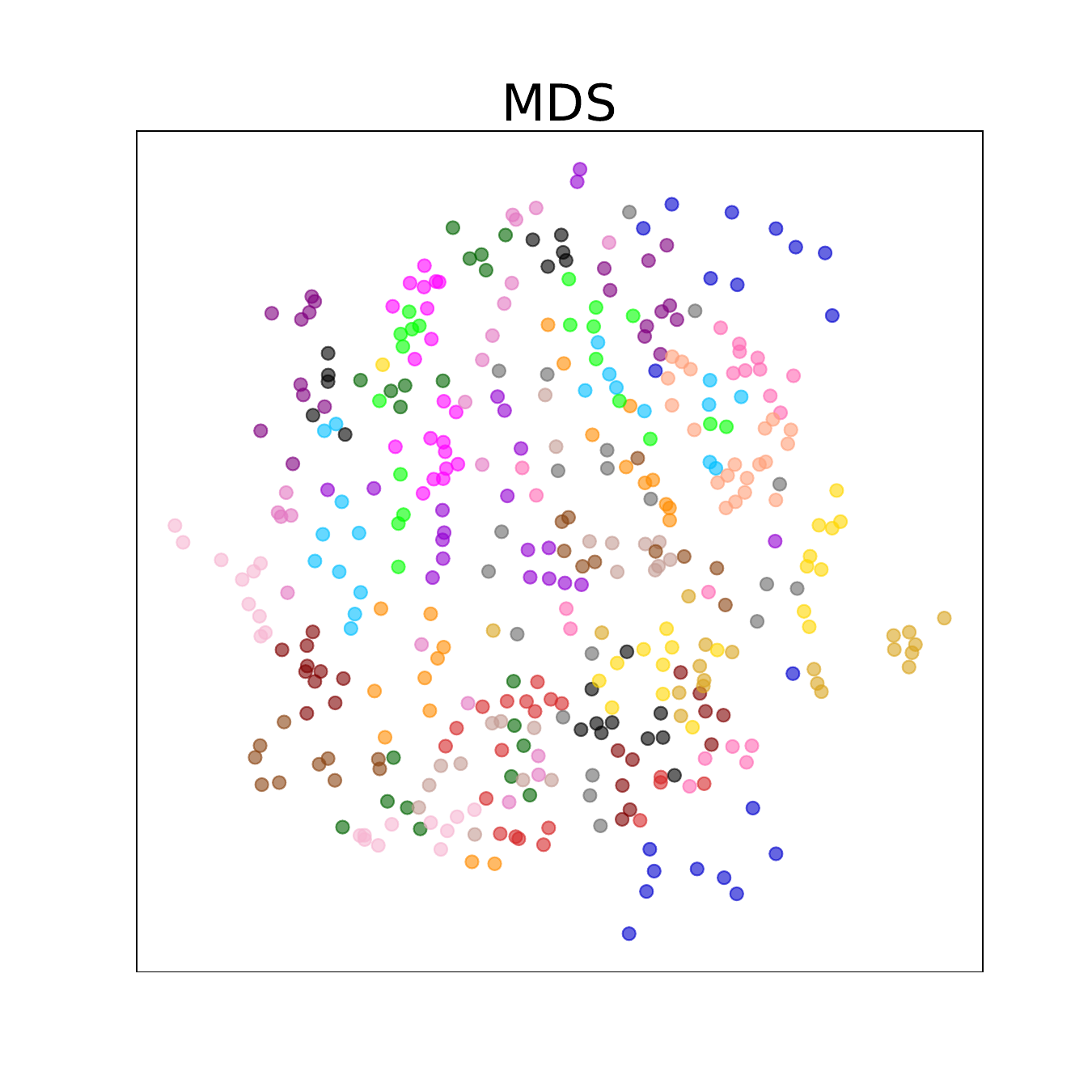}
 \includegraphics[width=0.32\linewidth,height=5cm]{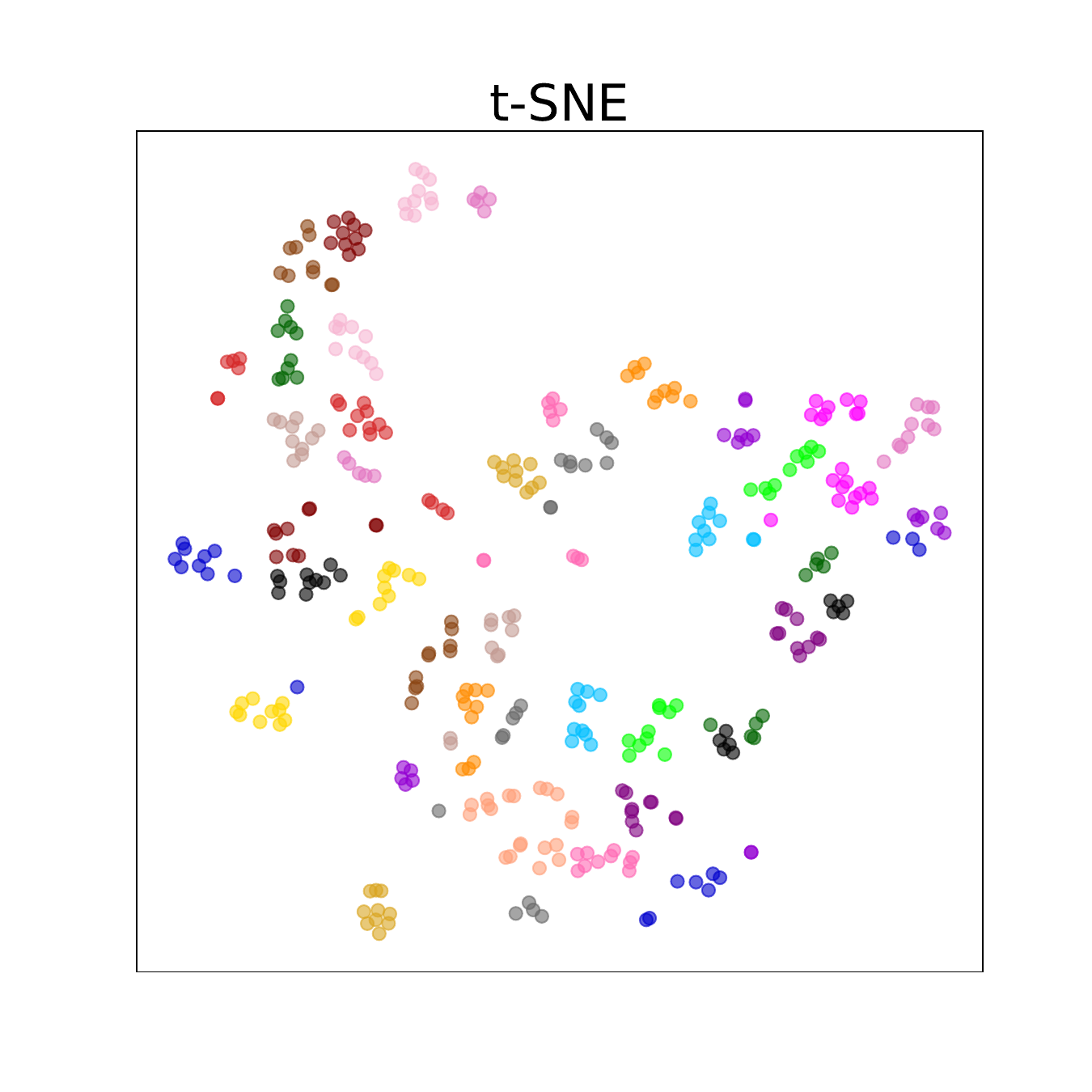} 
 \includegraphics[width=0.32\linewidth,height=5cm]{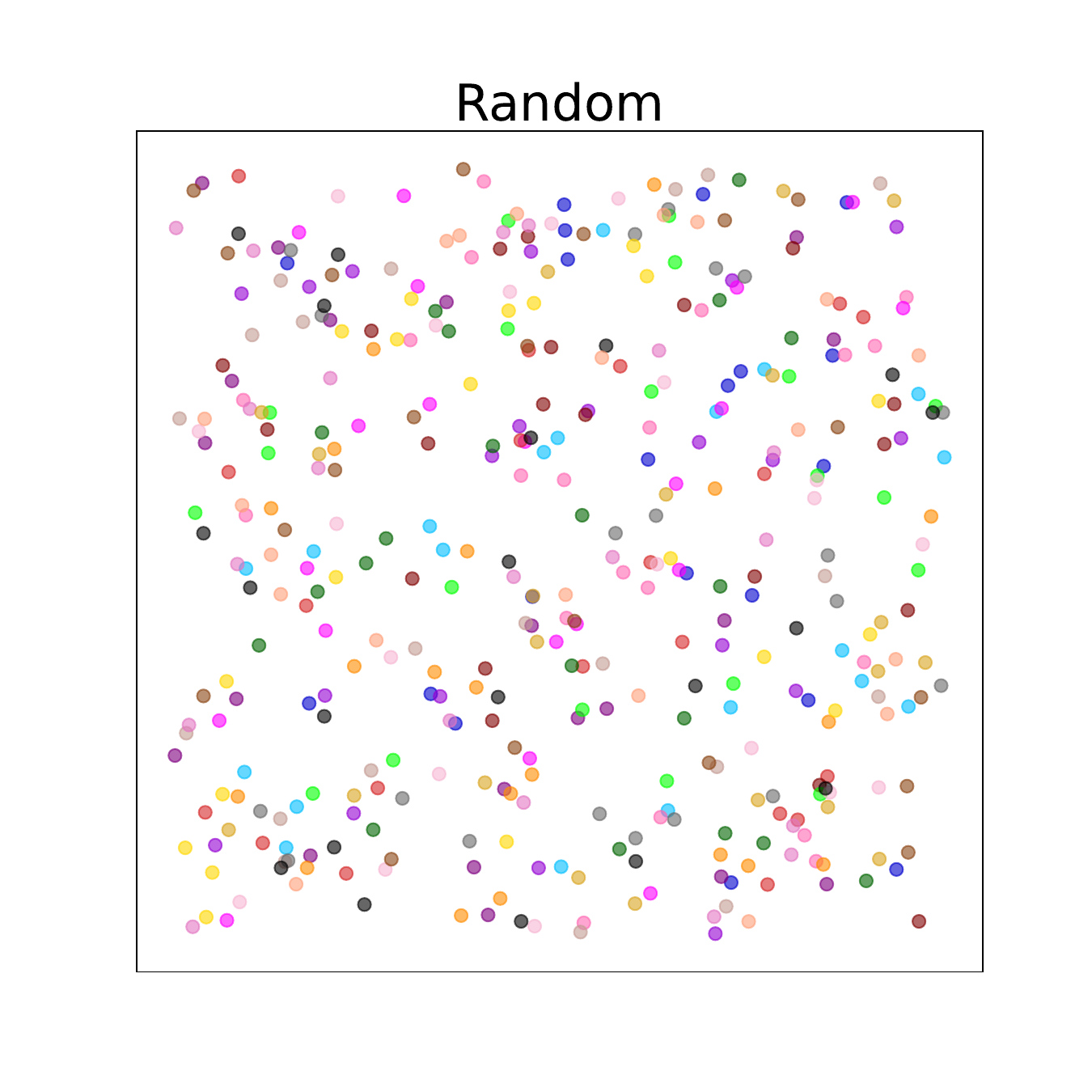}
 
\caption{
        MDS, t-SNE, and Random embeddings of the ORL dataset from left to right (bottom). 
        The plots (top) show the variation of Normalized Stress (top-left) and KL Divergence (top-right) with scale.
    }
    \label{fig:supplemental-orl}
}
\end{figure*}

\begin{figure*}[ht]{
 \centering

 \includegraphics[width=.49\linewidth]{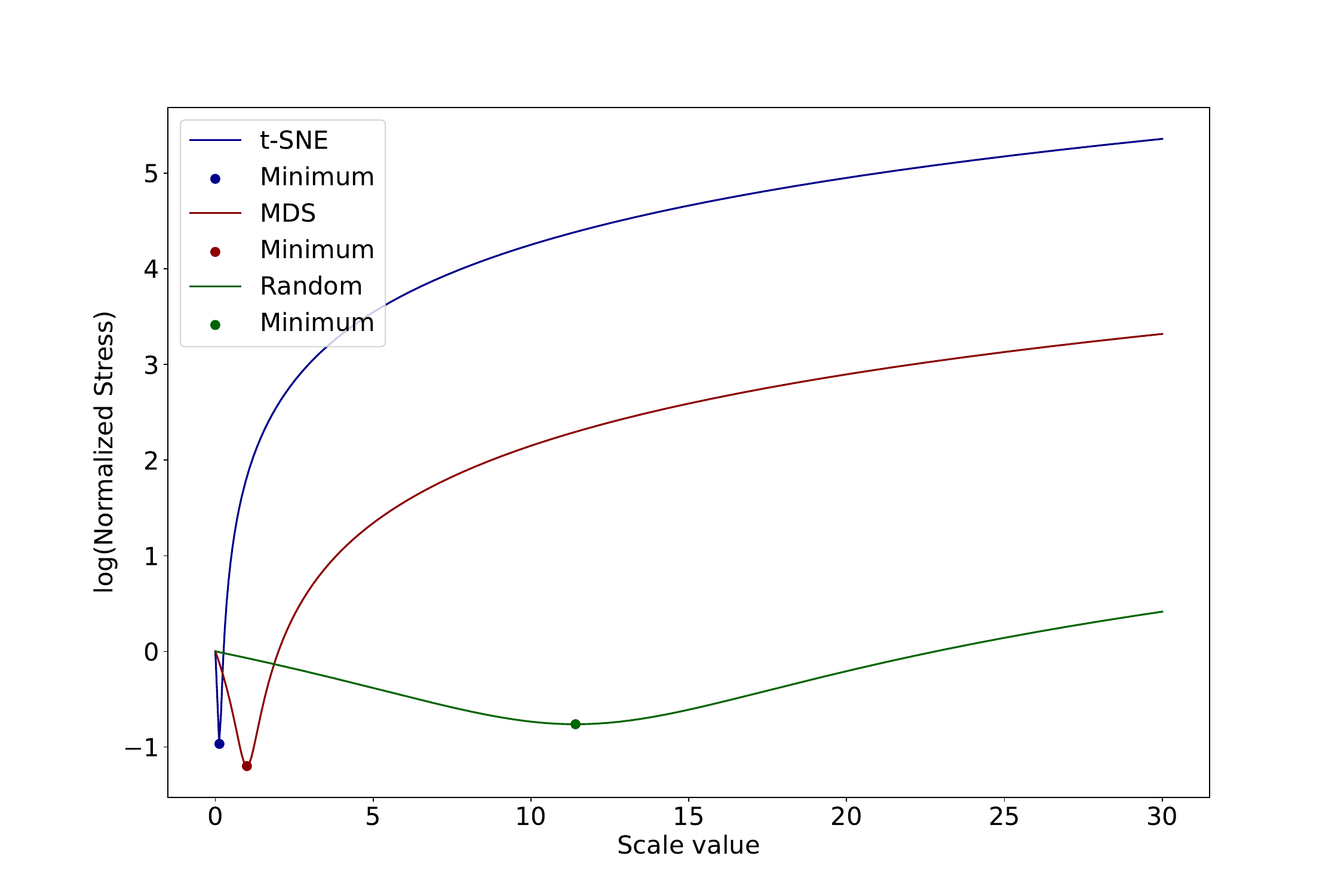}
 \includegraphics[width=0.49\linewidth]{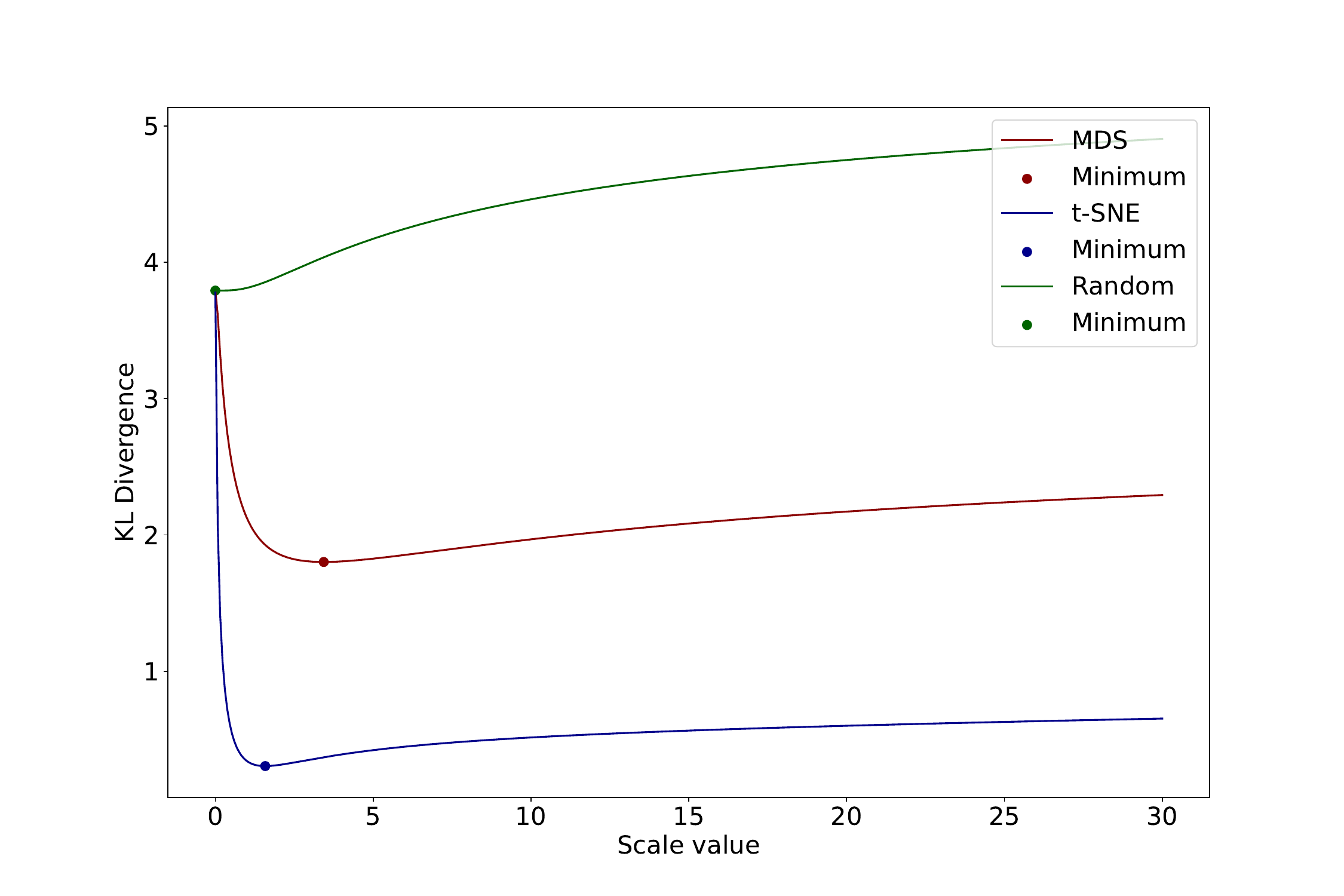}

 \includegraphics[width=0.32\linewidth,height=5cm]{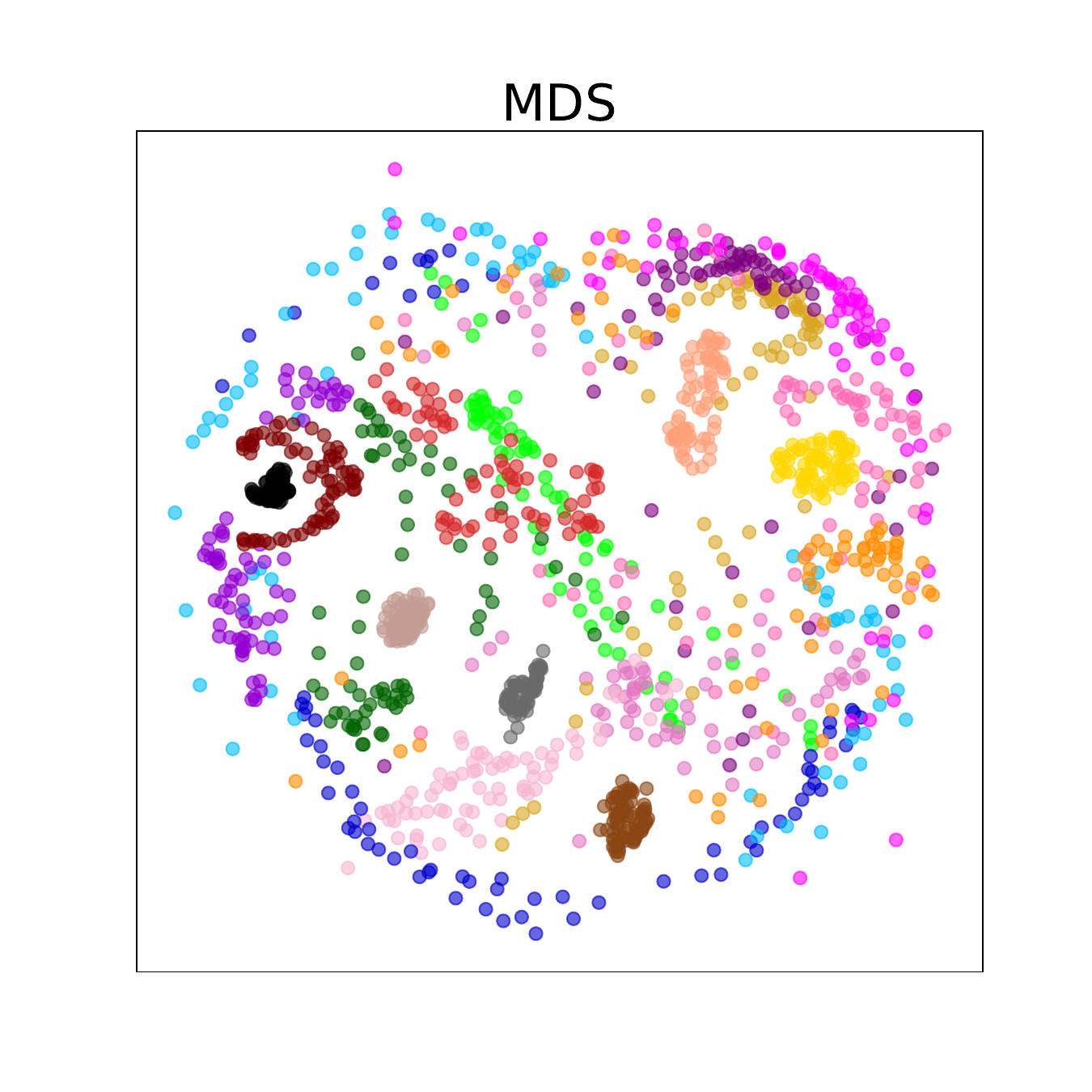}
 \includegraphics[width=0.32\linewidth,height=5cm]{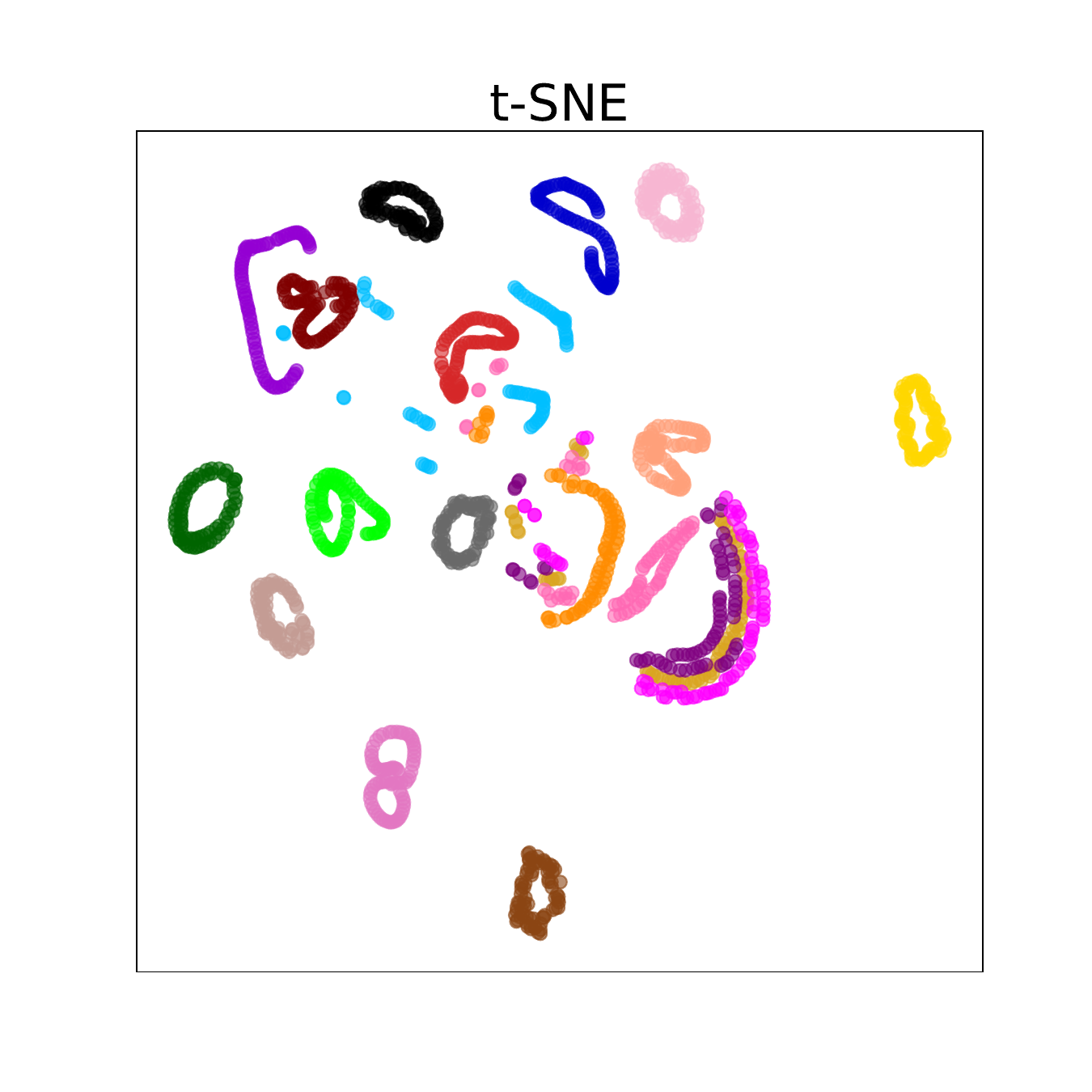} 
 \includegraphics[width=0.32\linewidth,height=5cm]{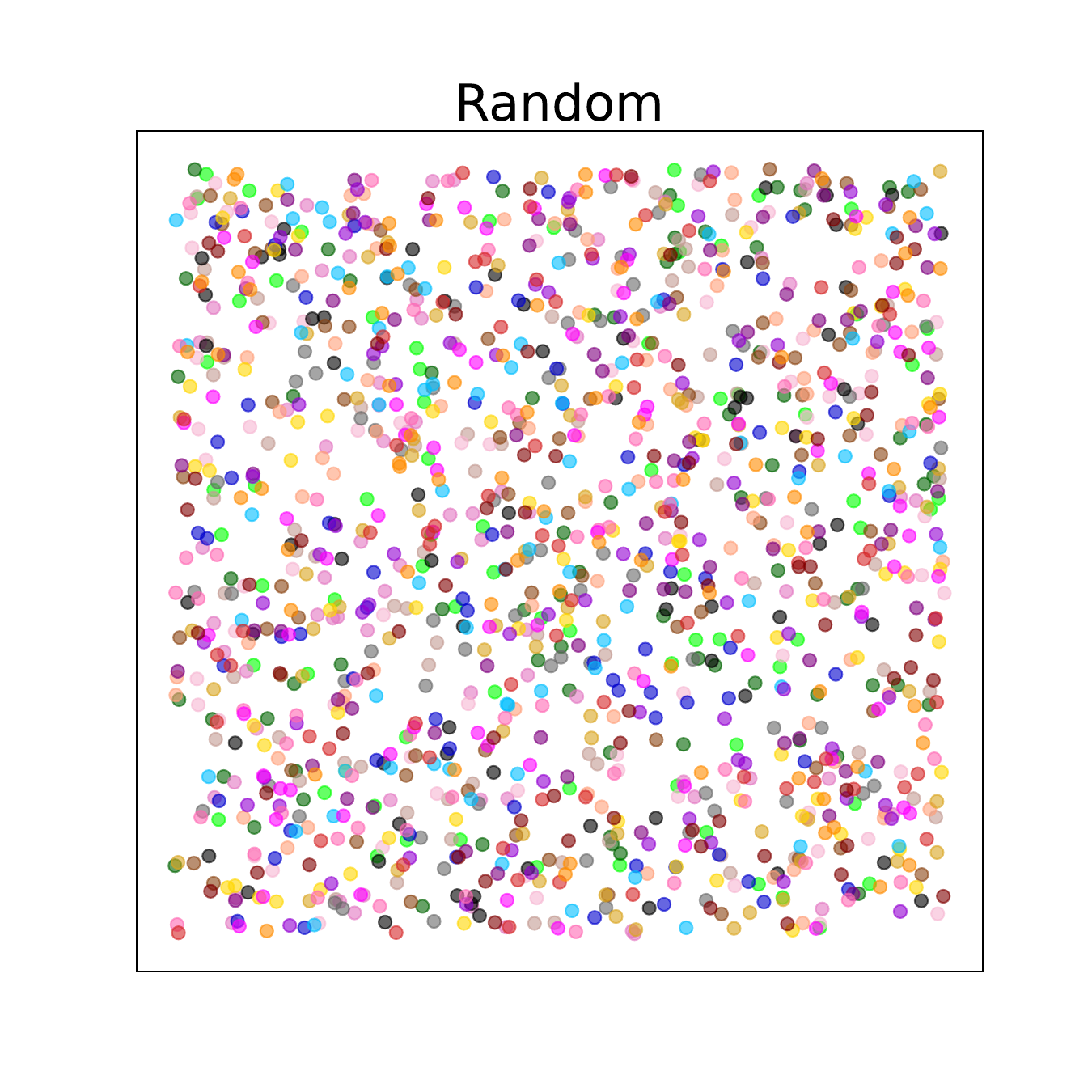}
\caption{
        MDS, t-SNE, and Random embeddings of the COIL-20 dataset from left to right (bottom). 
        The plots (top) show the variation of Normalized Stress (top-left) and KL Divergence (top-right) with scale.
    }
    \label{fig:supplemental-coil20}
}
\end{figure*}




\begin{figure*}[ht]{
 \centering

 \includegraphics[width=.49\linewidth]{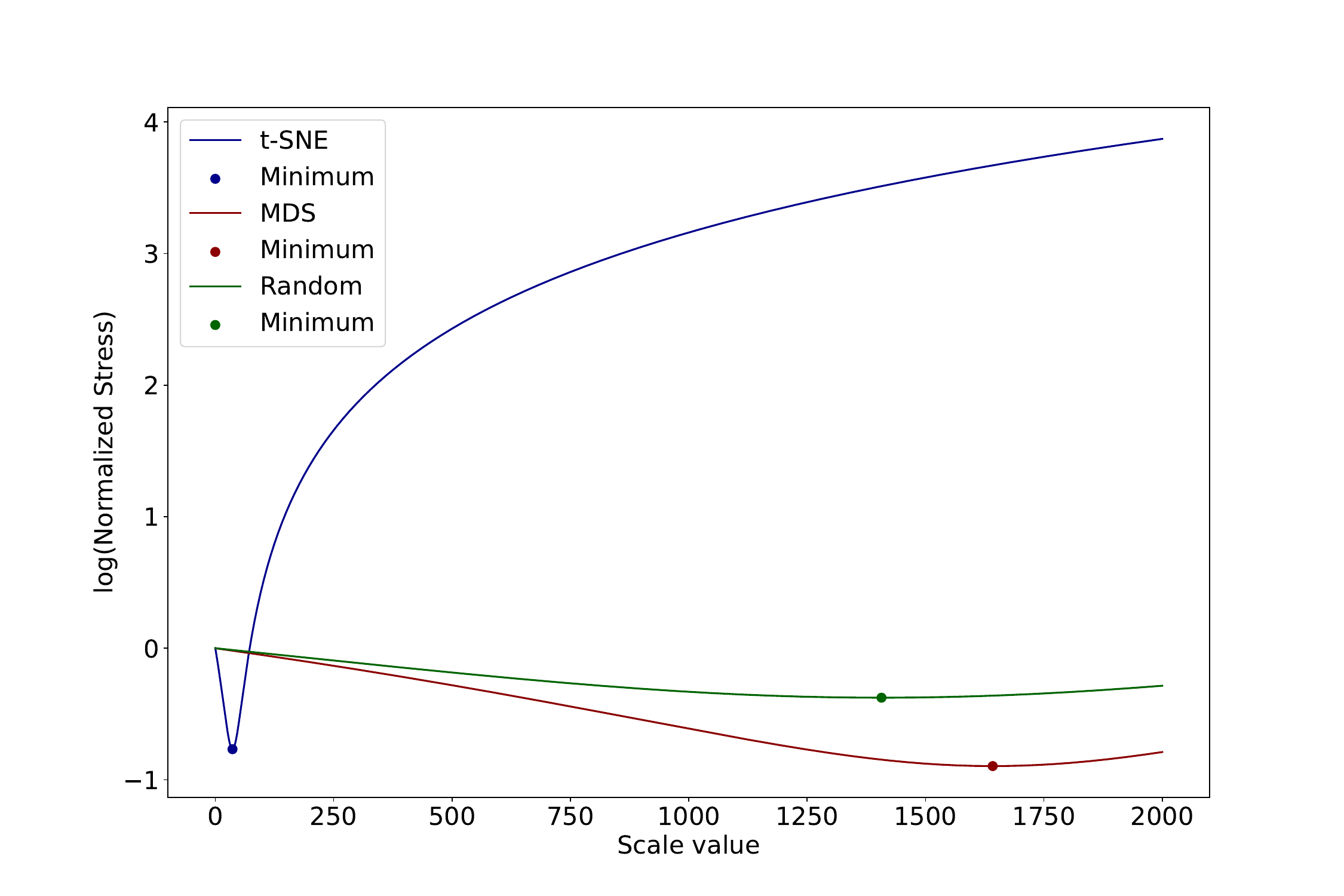}
 \includegraphics[width=0.49\linewidth]{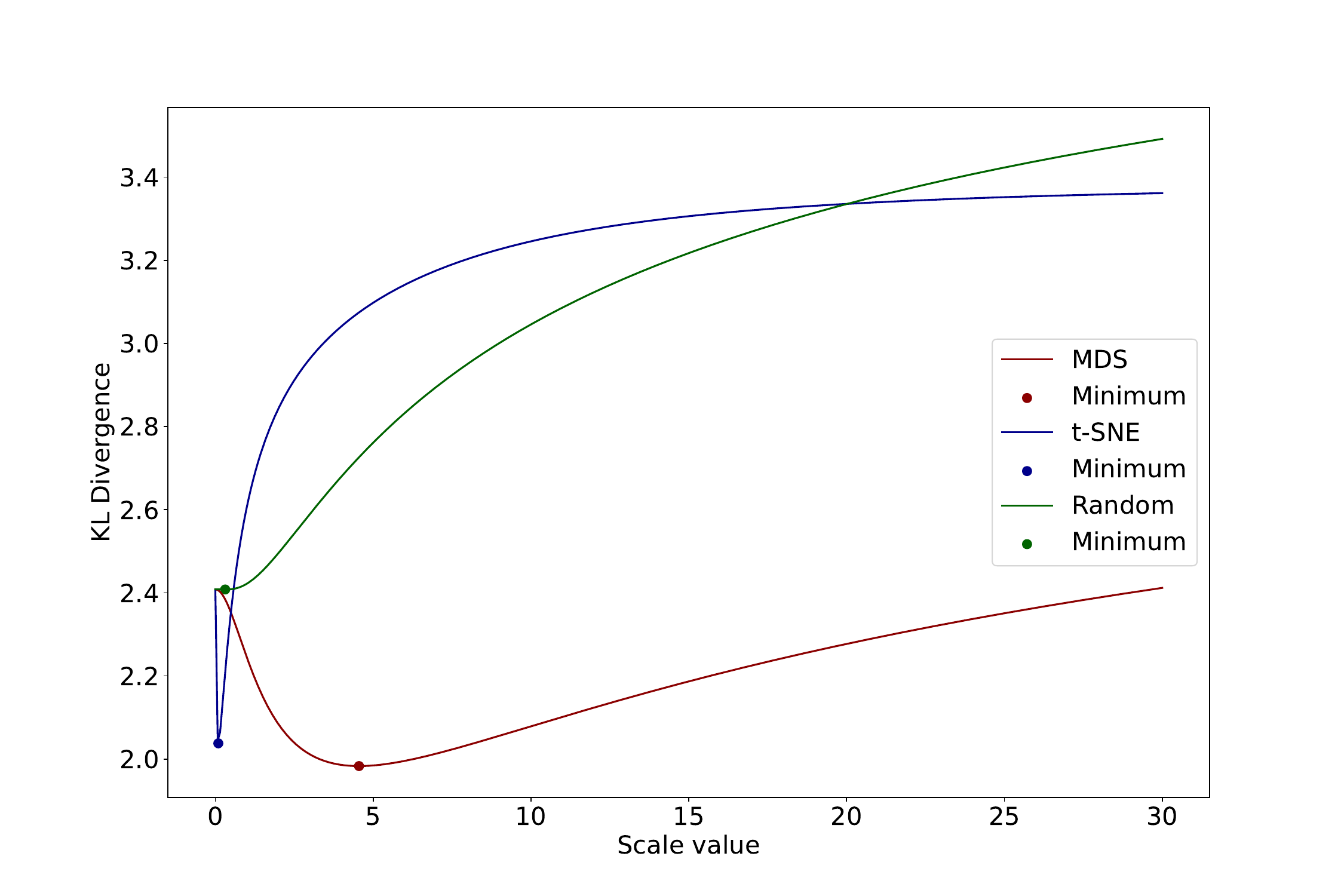}

 \includegraphics[width=0.32\linewidth,height=5cm]{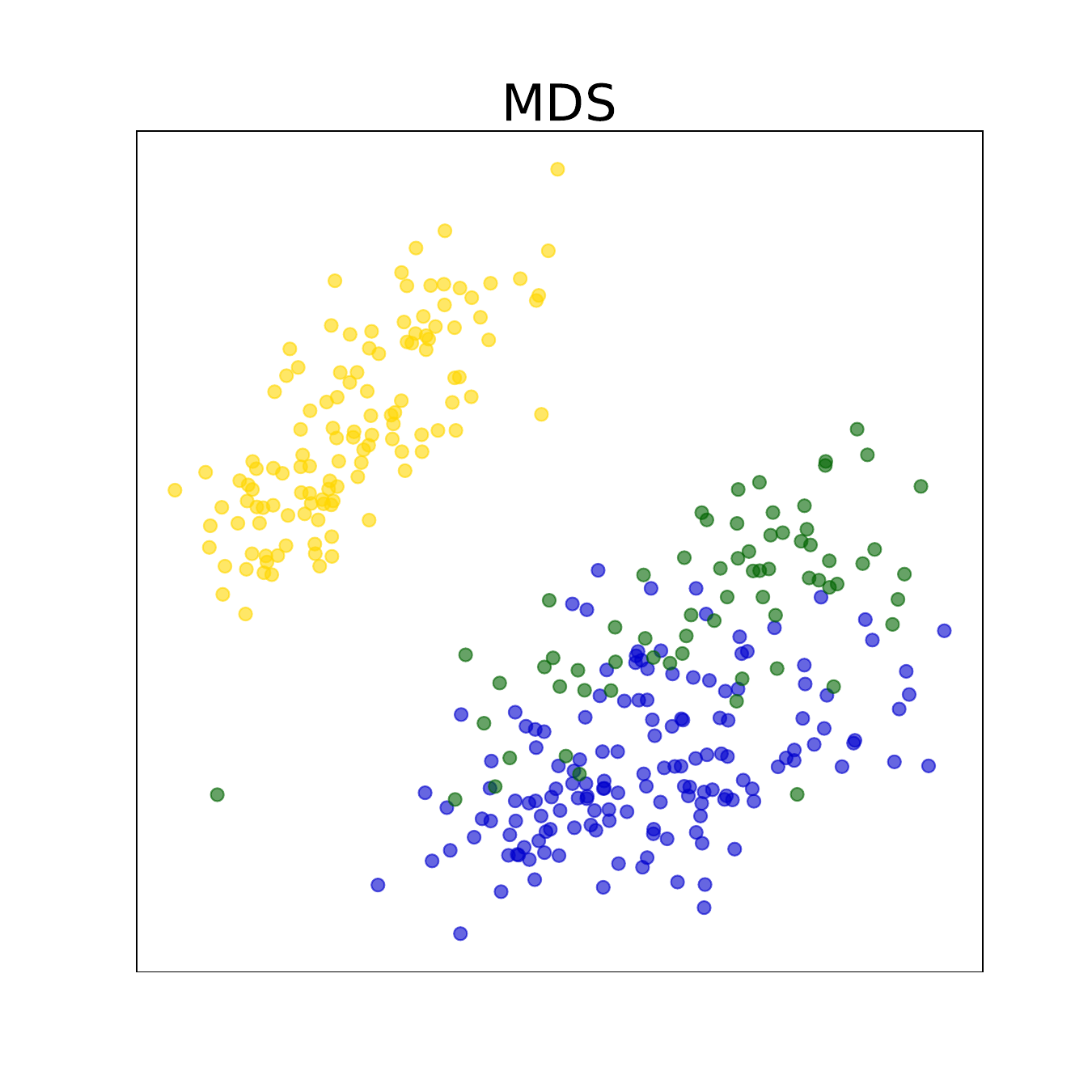}
 \includegraphics[width=0.32\linewidth,height=5cm]{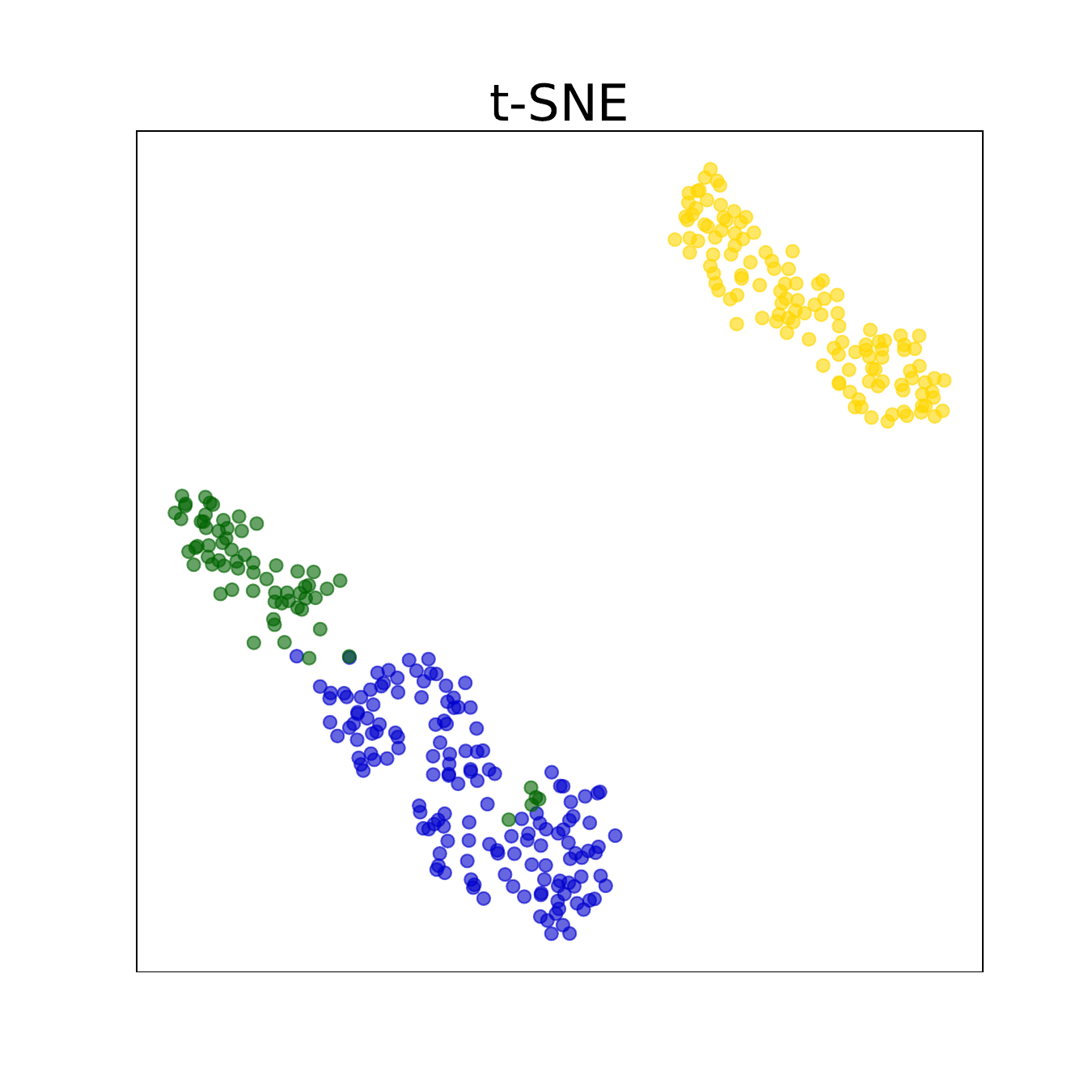} 
 \includegraphics[width=0.32\linewidth,height=5cm]{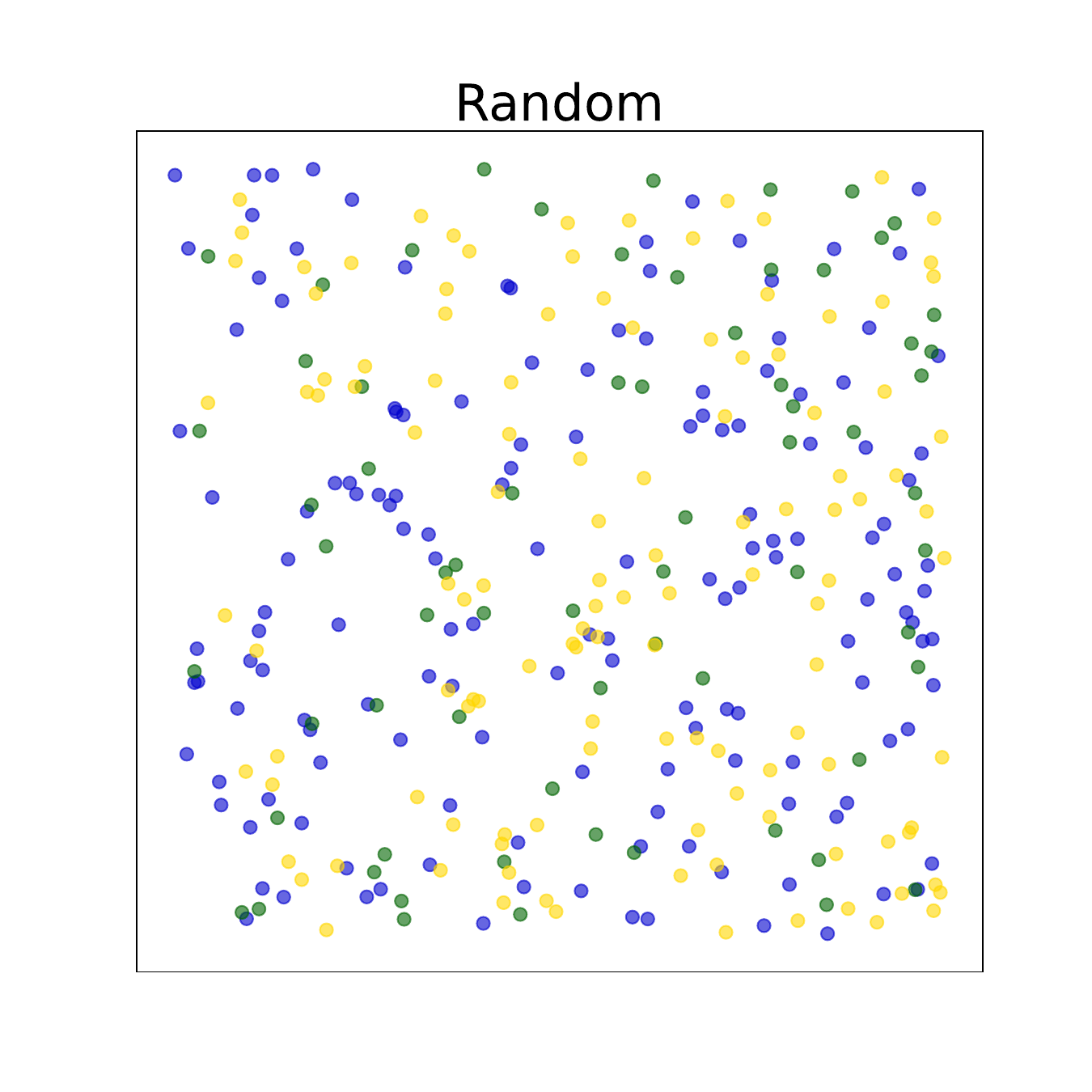}
\caption{
        MDS, t-SNE, and Random embeddings of the Palmer's Penguins dataset from left to right (bottom). 
        The plots (top) show the variation of Normalized Stress (top-left) and KL Divergence (top-right) with scale.
    }
    \label{fig:supplemental-penguins}
}
\end{figure*}

\begin{figure*}[ht]{
 \centering

 \includegraphics[width=.49\linewidth]{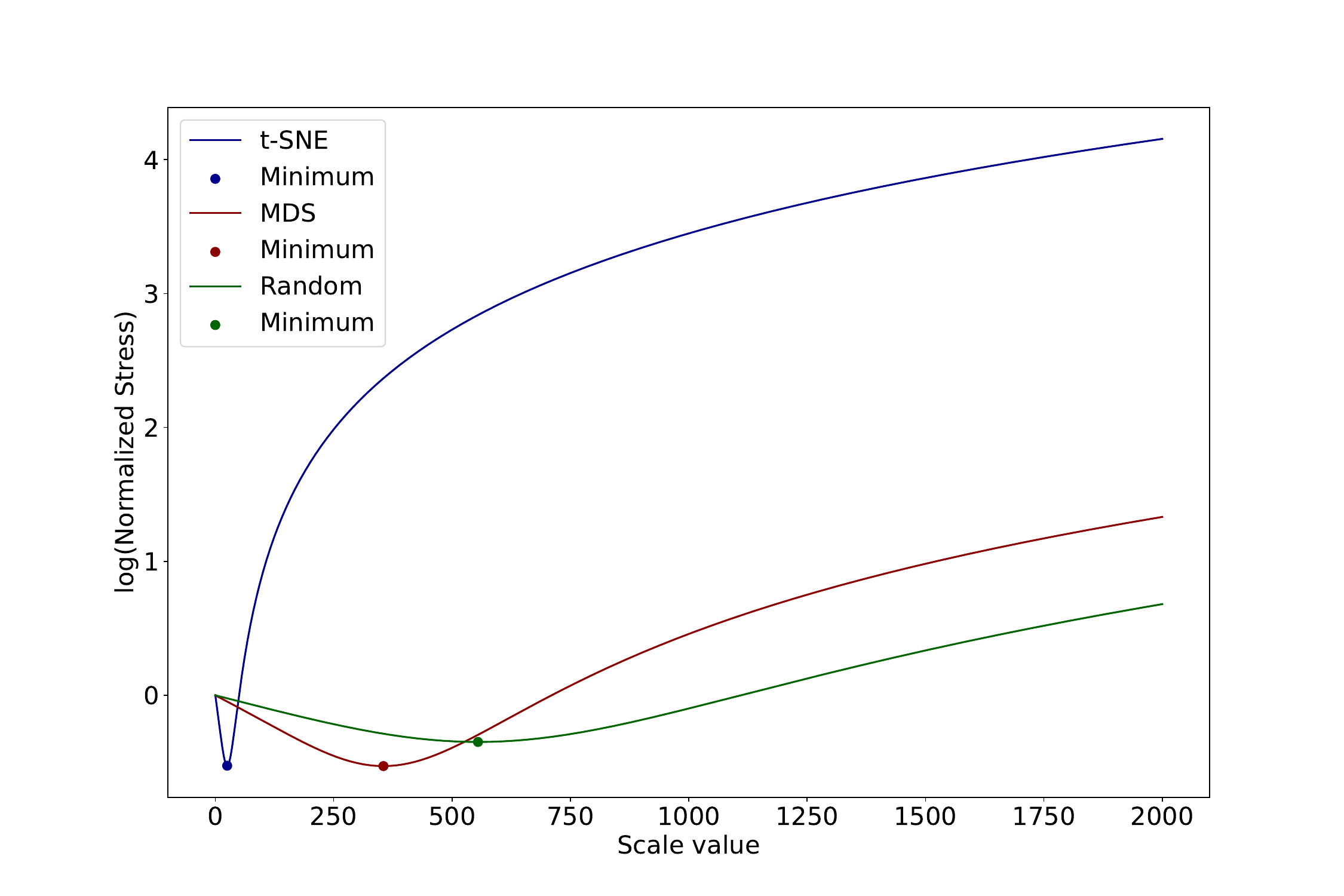}
 \includegraphics[width=0.49\linewidth]{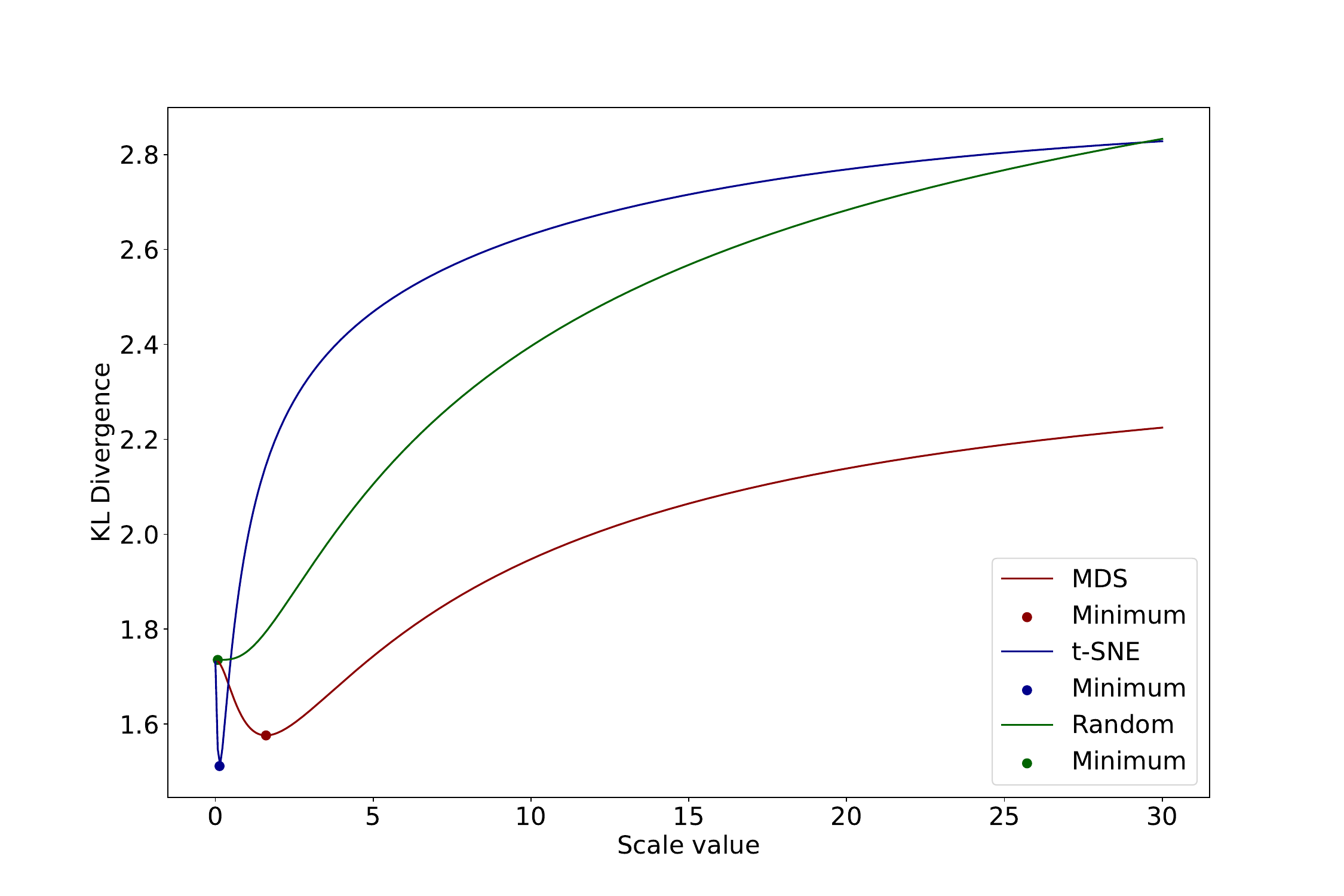}

 \includegraphics[width=0.32\linewidth,height=5cm]{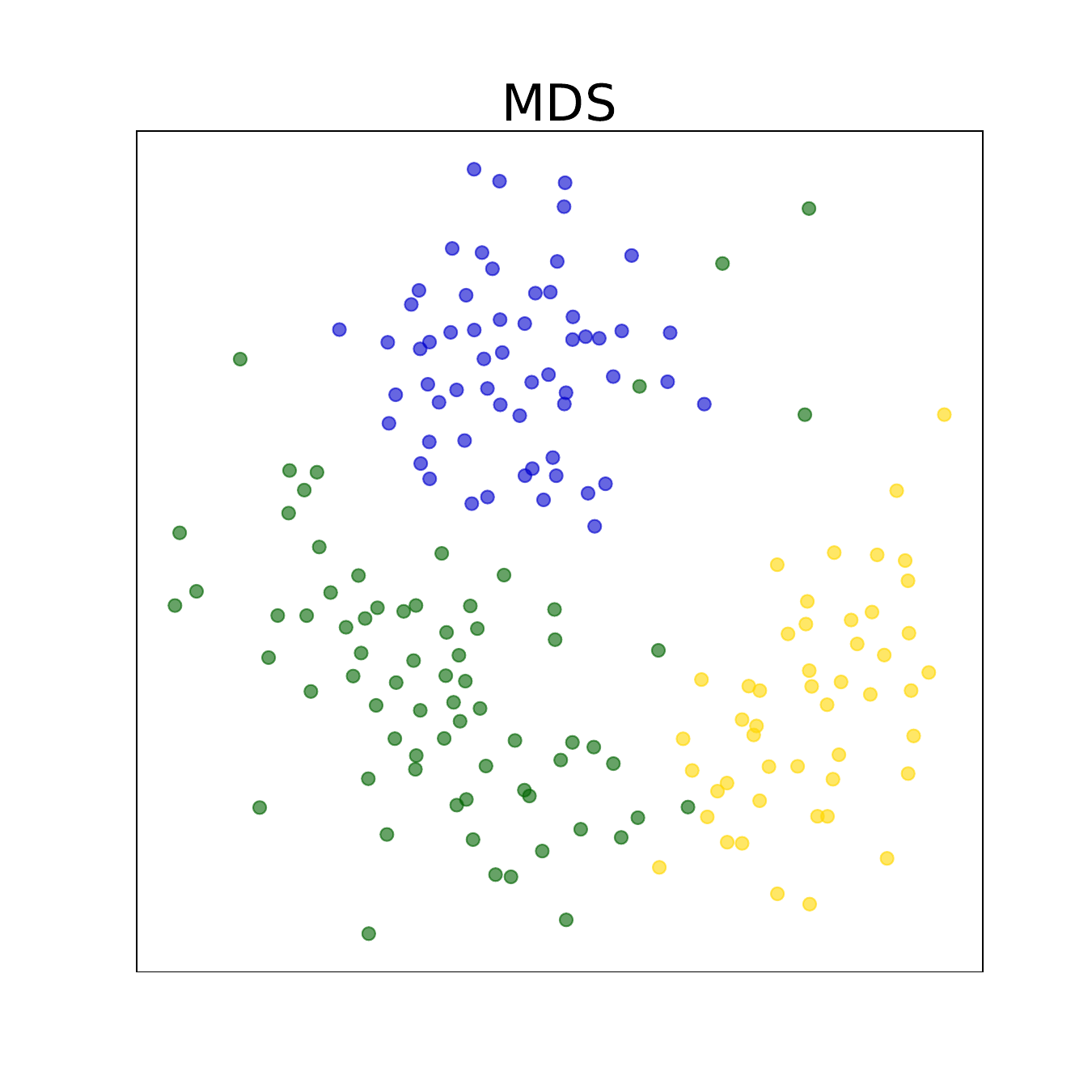}
 \includegraphics[width=0.32\linewidth,height=5cm]{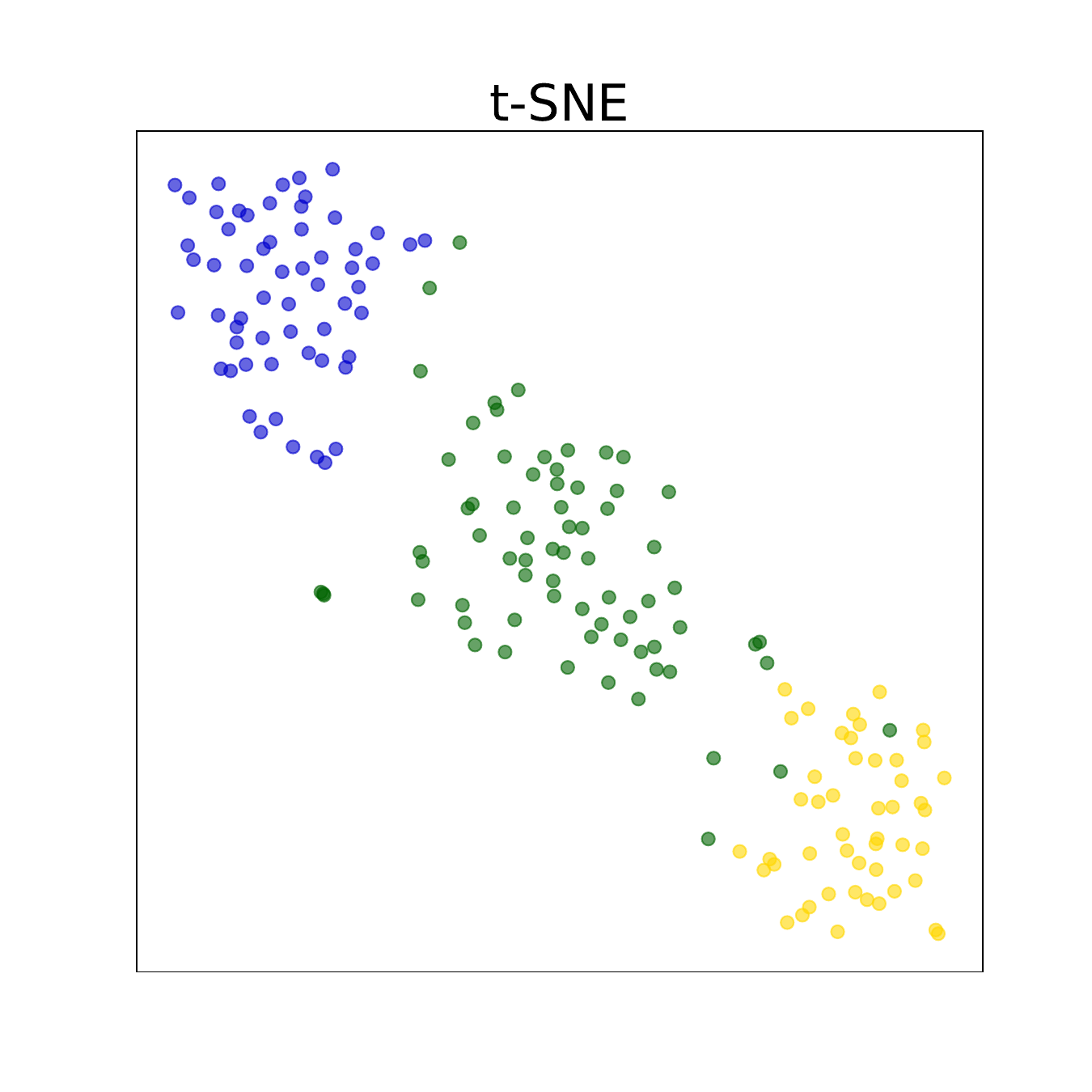} 
 \includegraphics[width=0.32\linewidth,height=5cm]{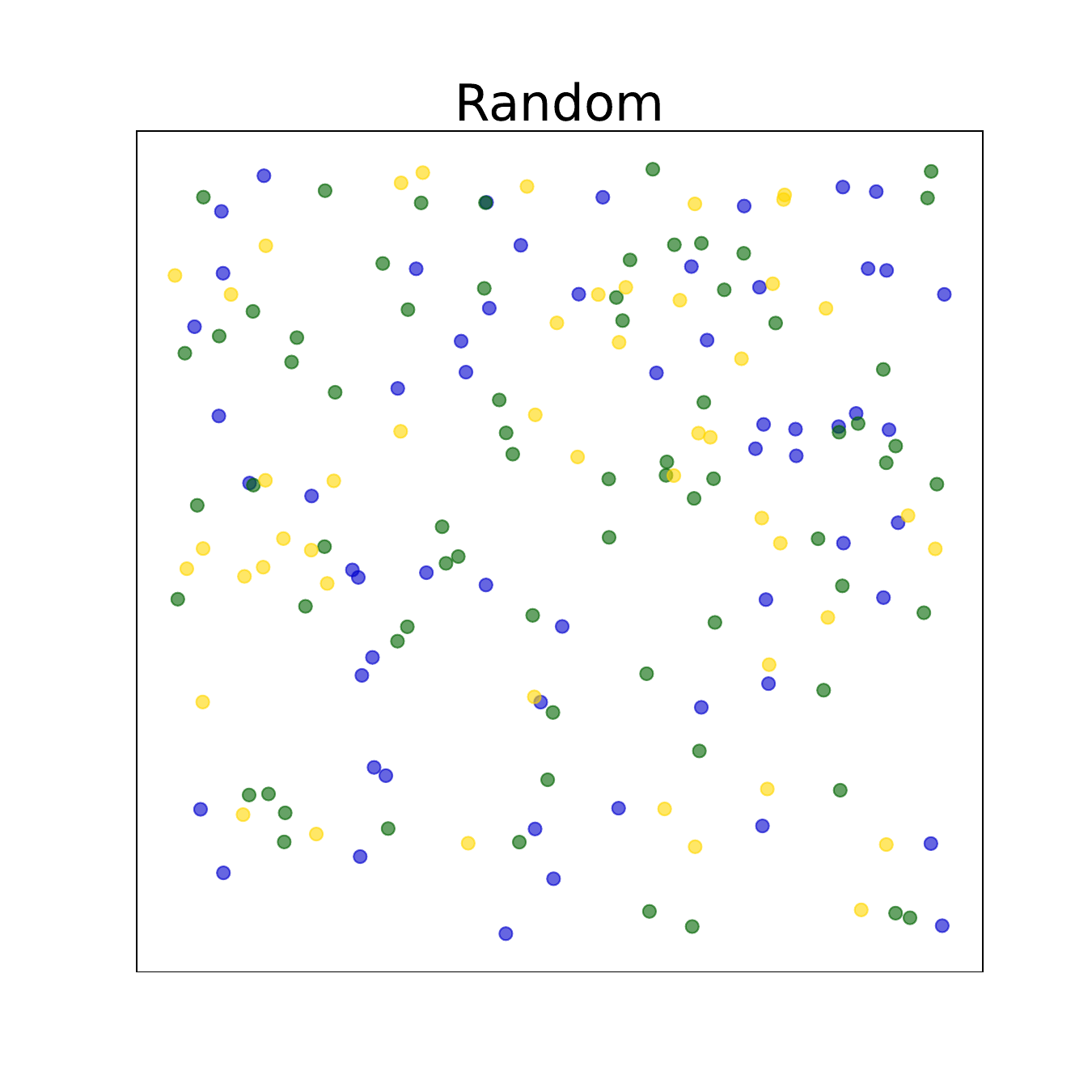}
\caption{
        MDS, t-SNE, and Random embeddings of the Wine dataset from left to right (bottom). 
        The plots (top) show the variation of Normalized Stress (top-left) and KL Divergence (top-right) with scale.
    }
    \label{fig:supplemental-wine}
}
\end{figure*}

\begin{figure*}[ht]{
 \centering

 \includegraphics[width=.49\linewidth]{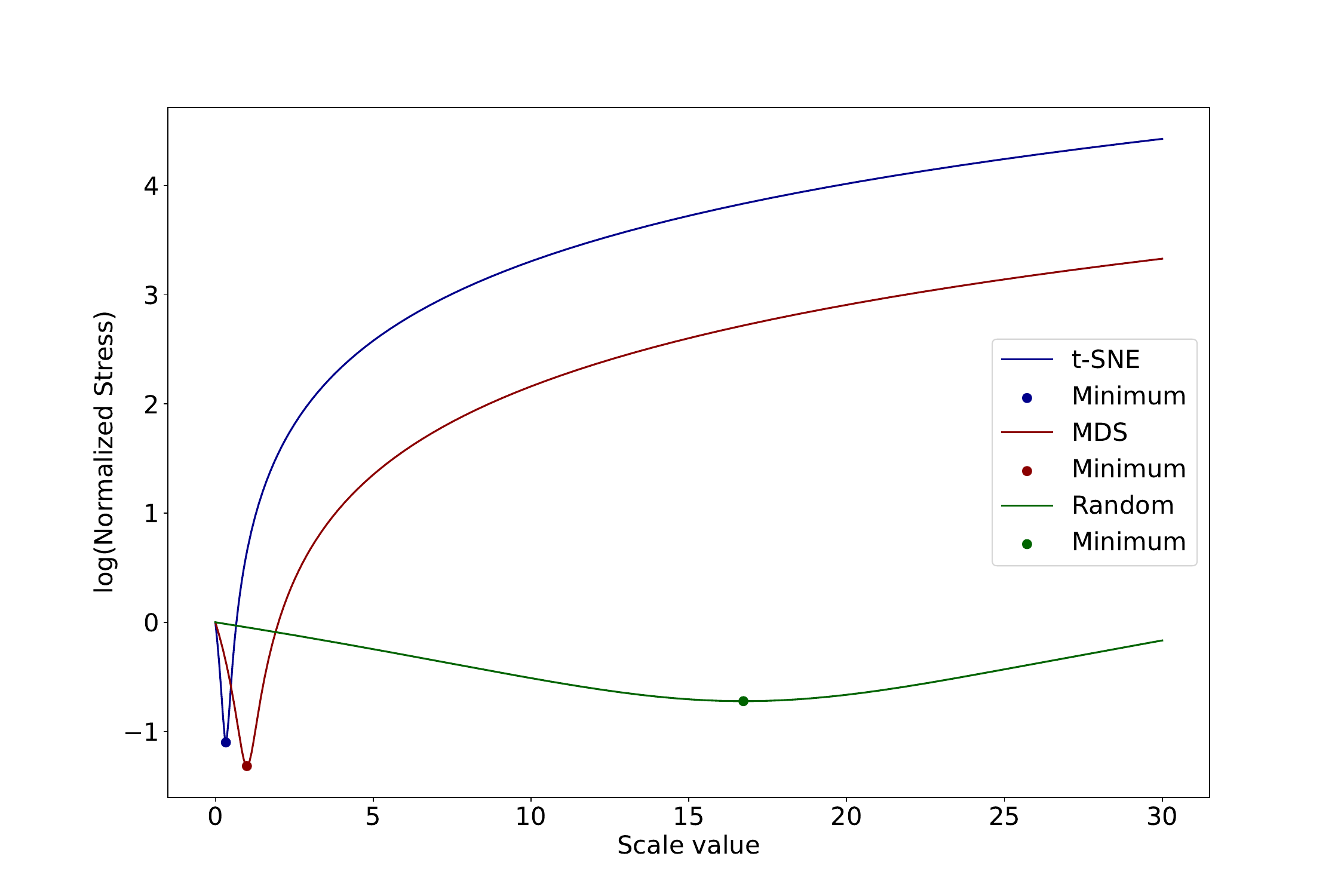} \includegraphics[width=0.49\linewidth]{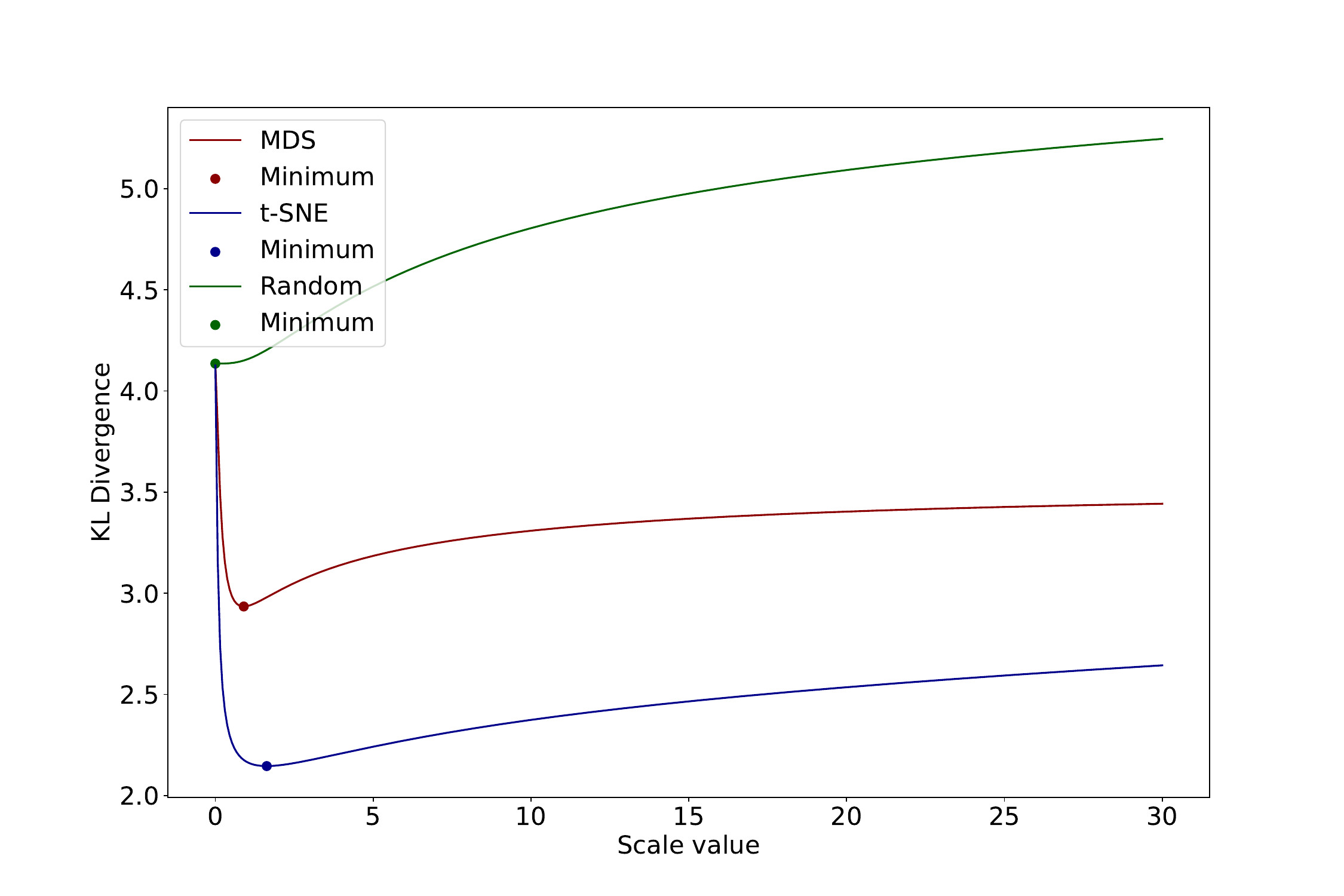}

 \includegraphics[width=0.32\linewidth,height=5cm]{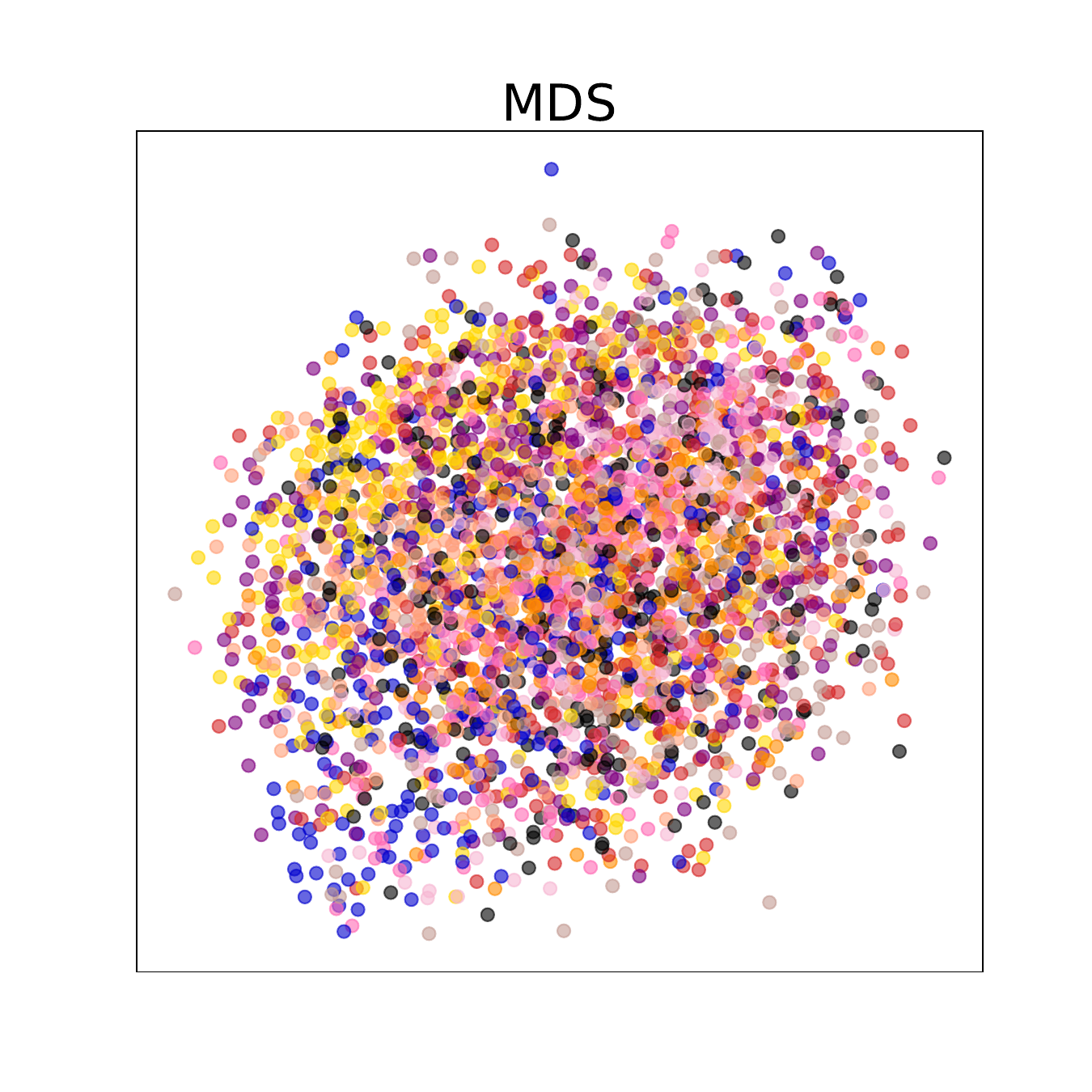}
 \includegraphics[width=0.32\linewidth,height=5cm]{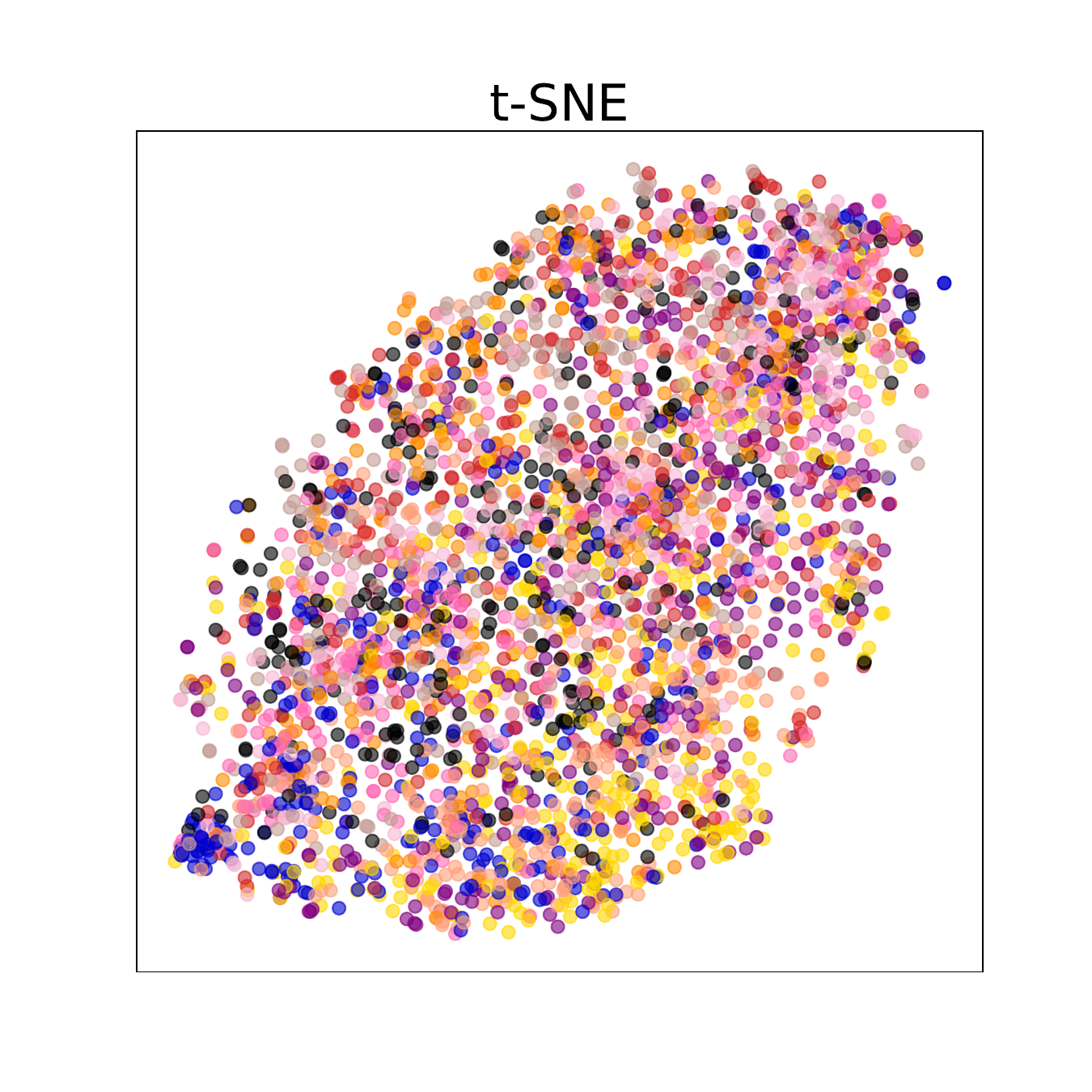} 
 \includegraphics[width=0.32\linewidth,height=5cm]{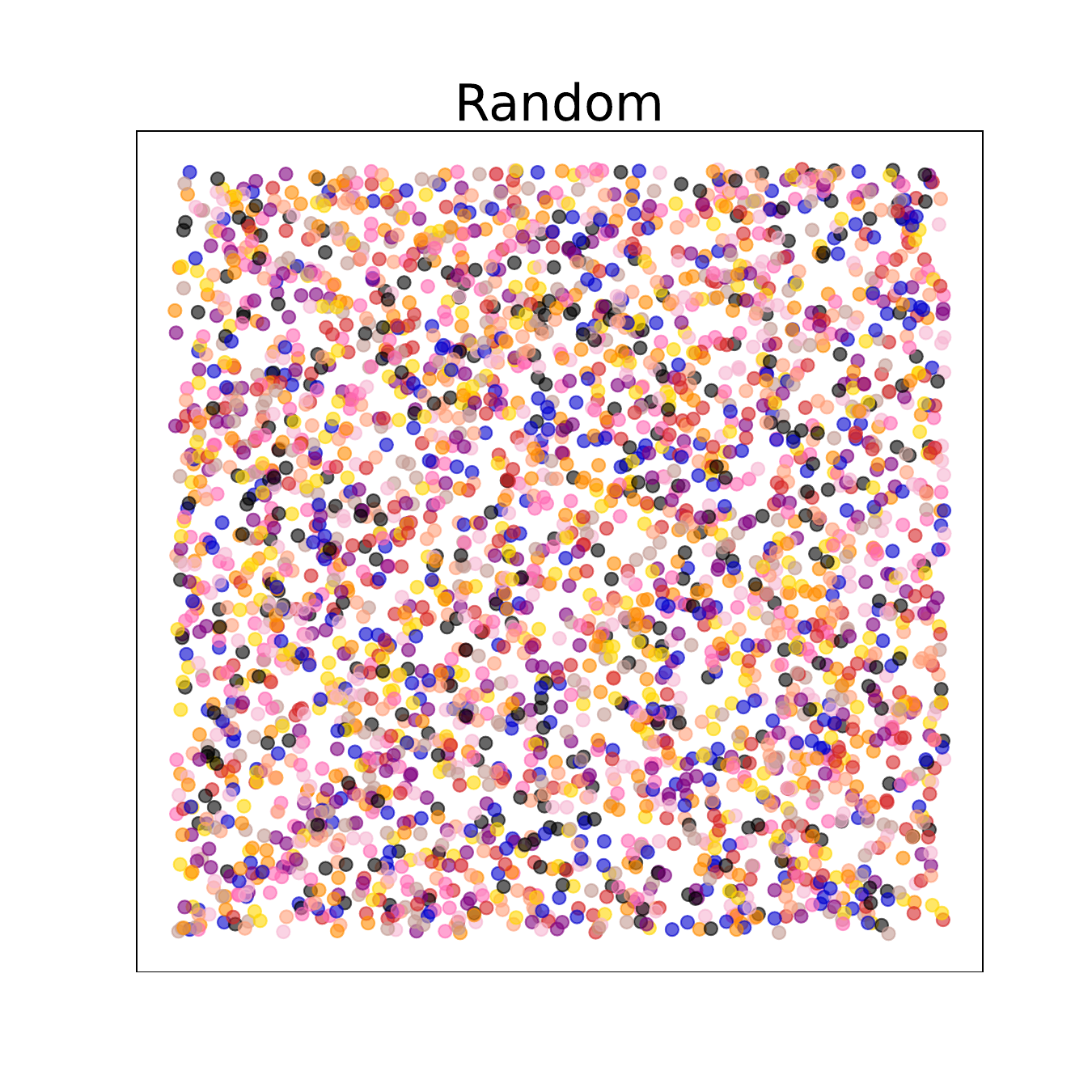}
\caption{
        MDS, t-SNE, and Random embeddings of the CIFAR-10 dataset from left to right (bottom). 
        The plots (top) show the variation of Normalized Stress (top-left) and KL Divergence (top-right) with scale.
    }
    \label{fig:supplemental-cifar10}
}
\end{figure*}

\begin{figure*}[ht]{
 \centering

 \includegraphics[width=.49\linewidth]{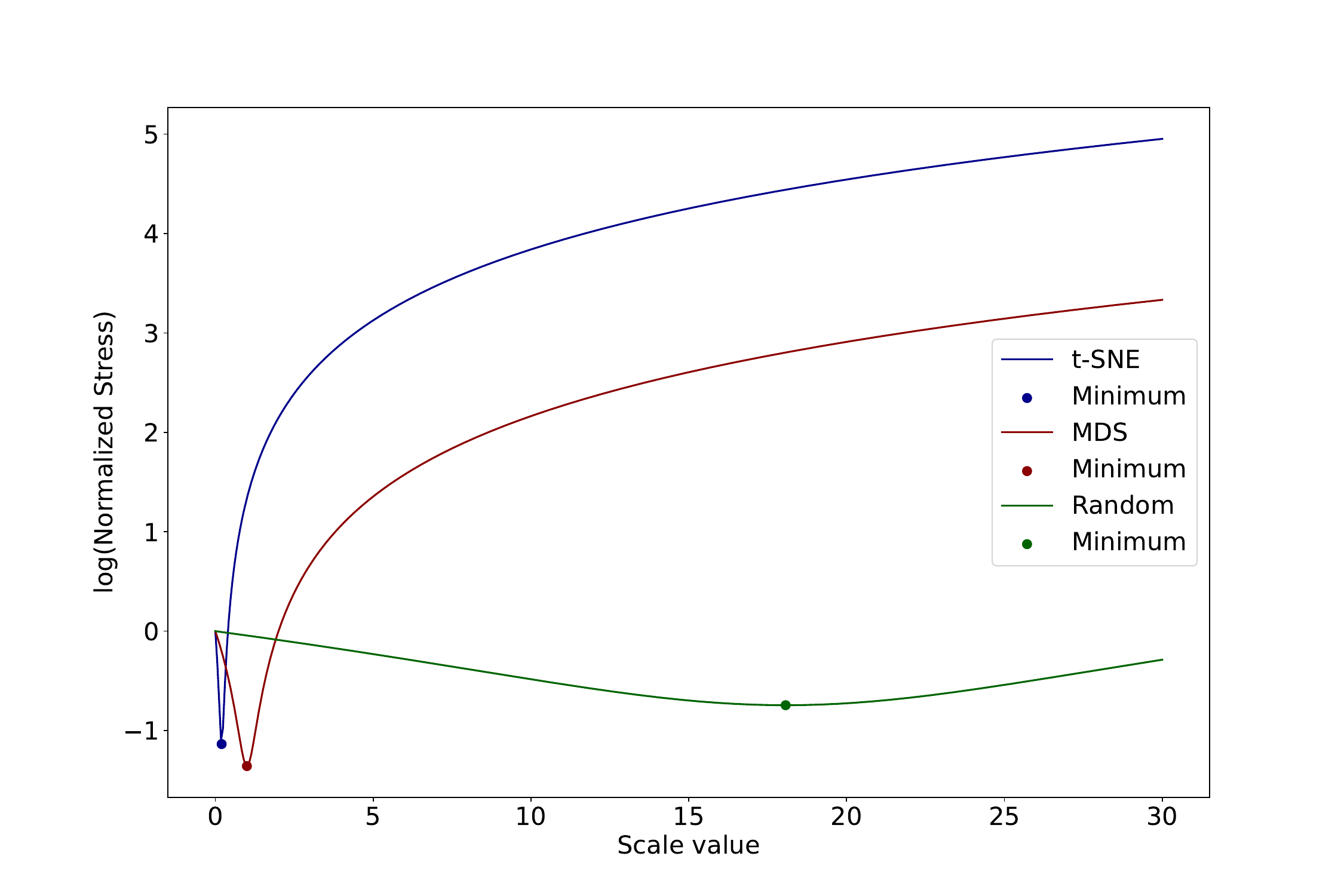} \includegraphics[width=0.49\linewidth]{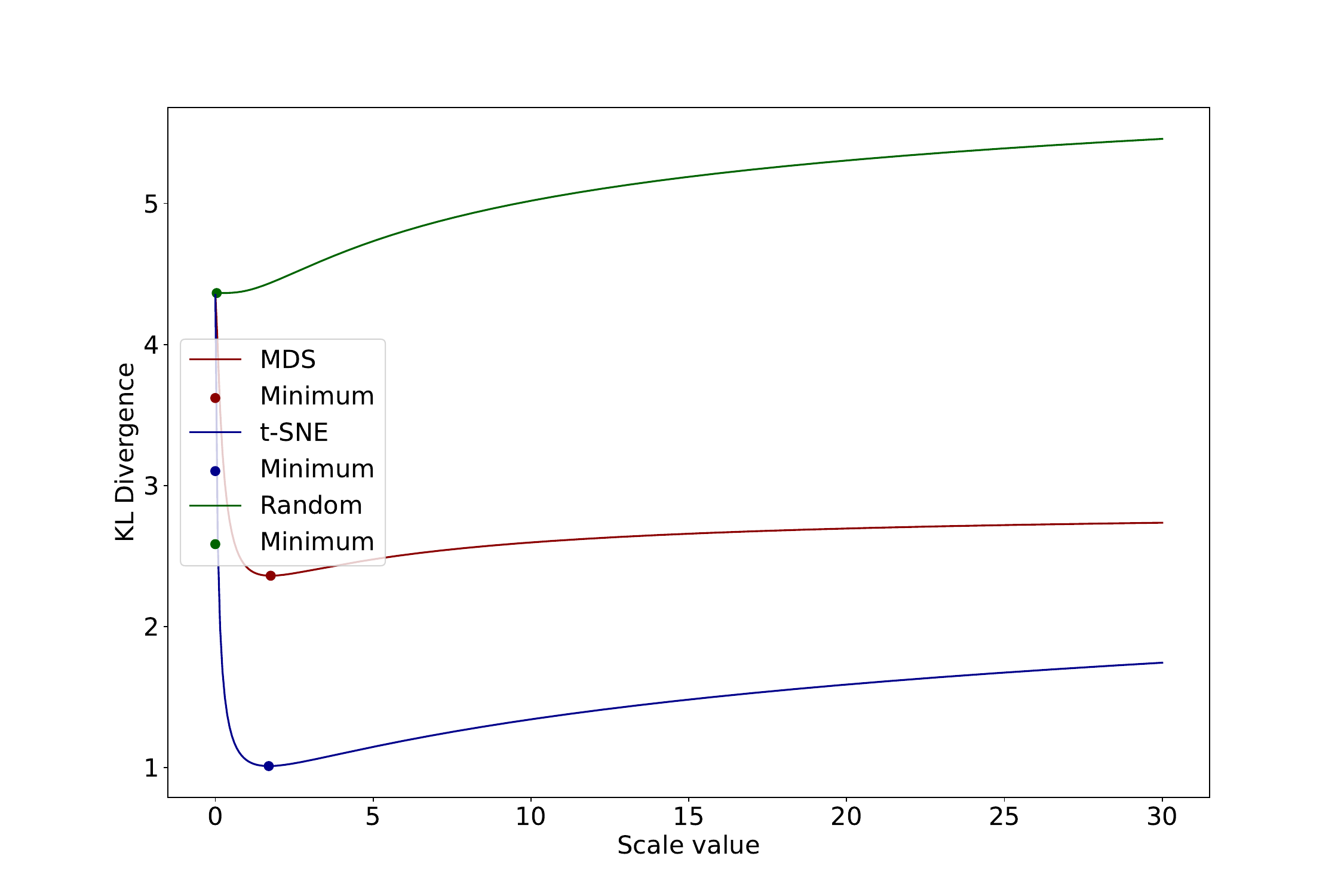}

 \includegraphics[width=0.32\linewidth,height=5cm]{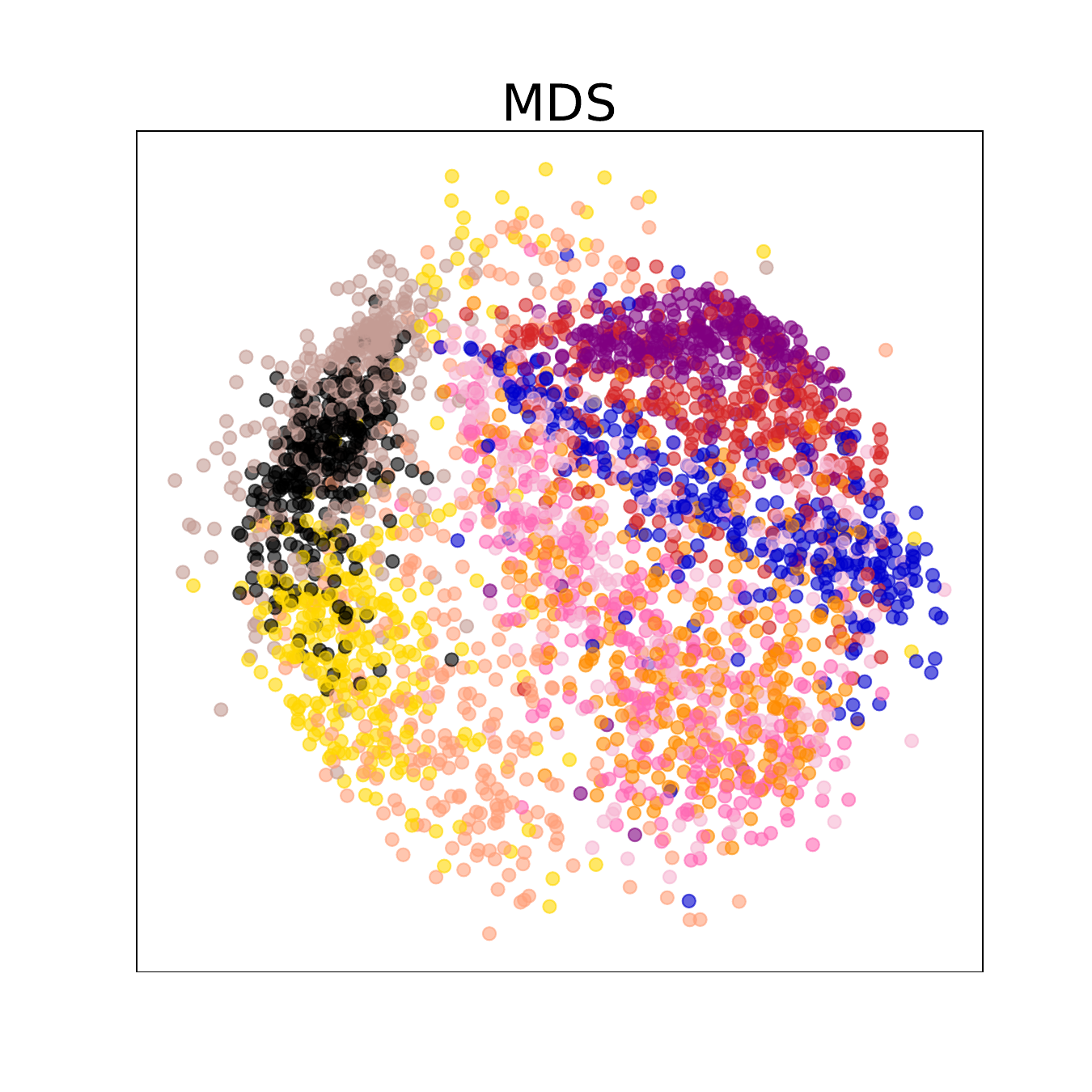}
 \includegraphics[width=0.32\linewidth,height=5cm]{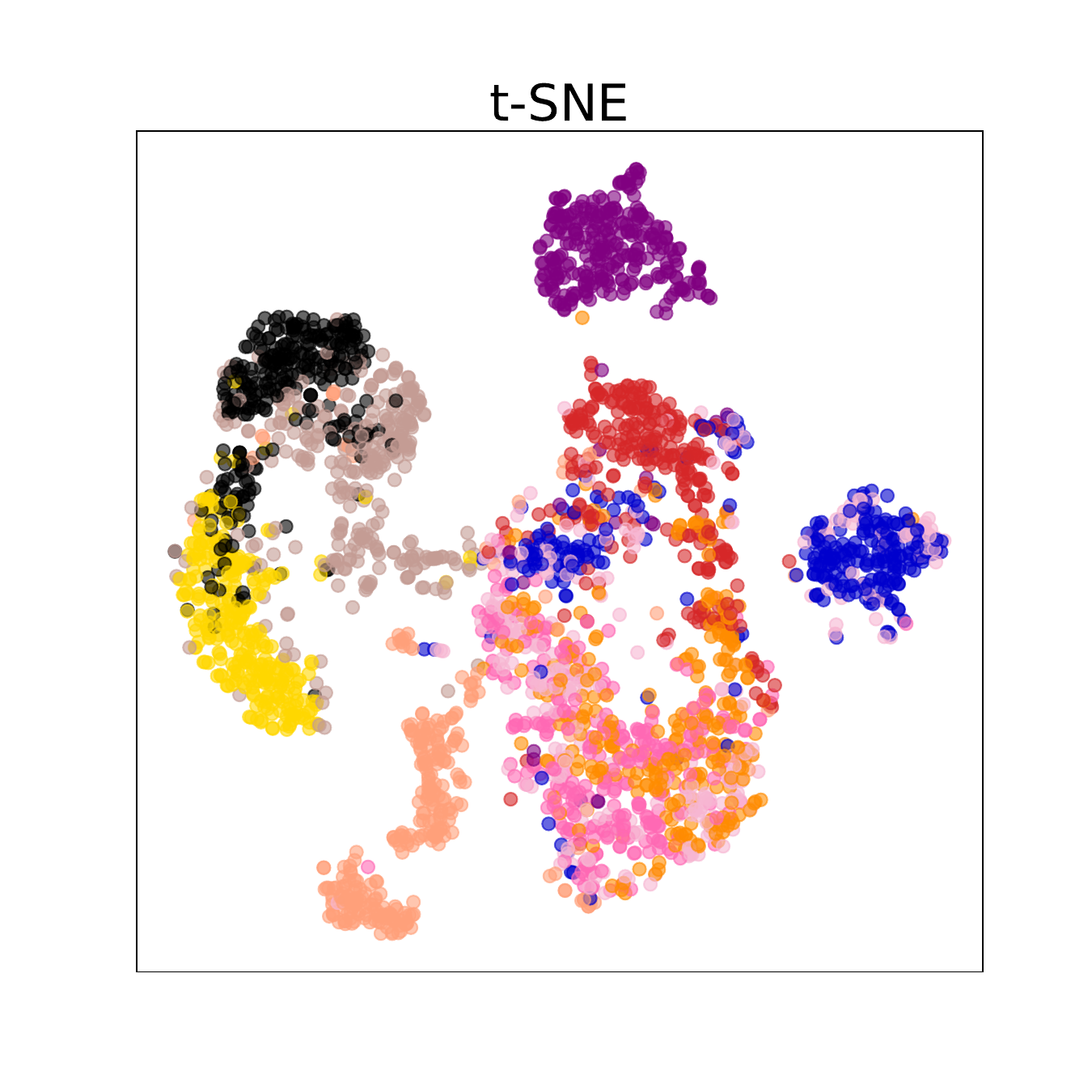} 
 \includegraphics[width=0.32\linewidth,height=5cm]{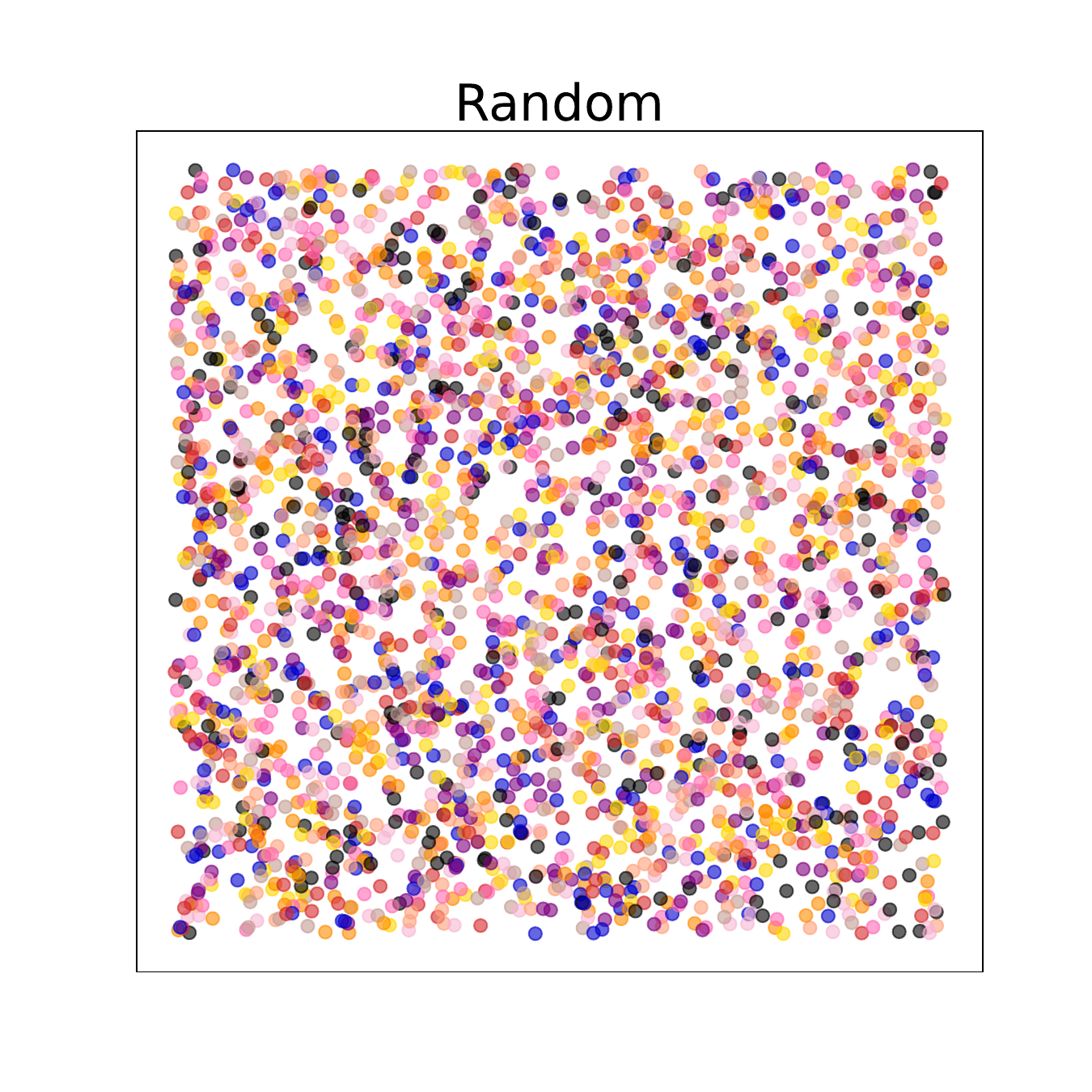}
\caption{
        MDS, t-SNE, and Random embeddings of the Fashion MNIST dataset from left to right (bottom). 
        The plots (top) show the variation of Normalized Stress (top-left) and KL Divergence (top-right) with scale.
    }
    \label{fig:supplemental-fashion}
}
\end{figure*}

\begin{figure}
    \centering
    \includegraphics[width=\linewidth]{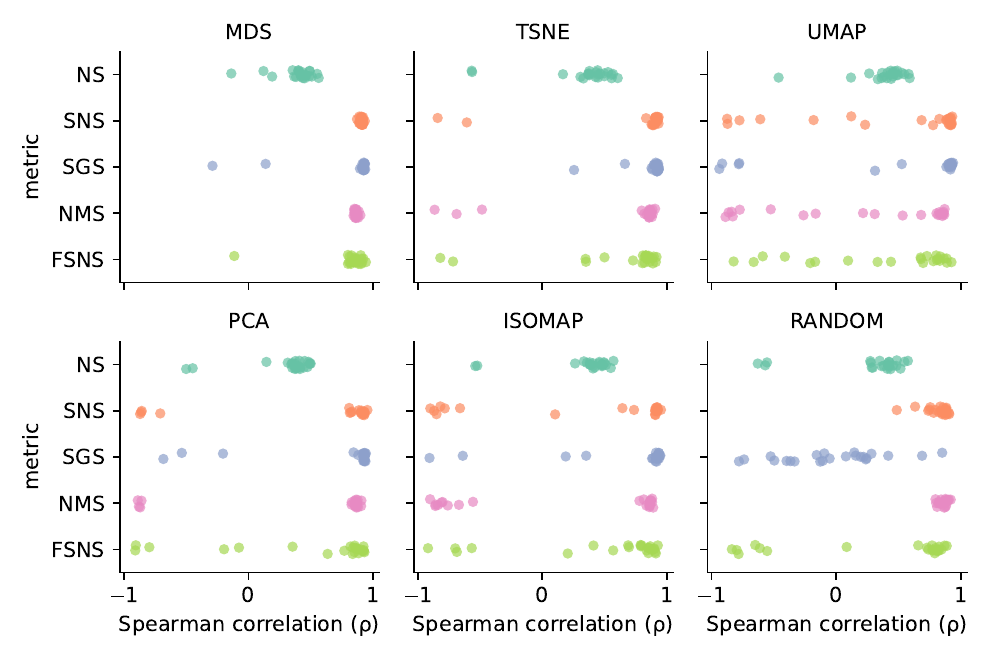}
    \caption{{Pearson correlation for each metric value against the number of noise iterations in the sensitivity experiment. Each point is an embedding generated by the indicated algorithm. A high positive correlation means that as an embedding becomes more noisy, the metric tends to increase while a high negative correlation indicates the opposite. A consistent metric would have values near (+)1. }}
    \label{fig:ladder-correlation}
\end{figure}

\begin{figure*}
    \centering
    \includegraphics[width=0.32\linewidth]{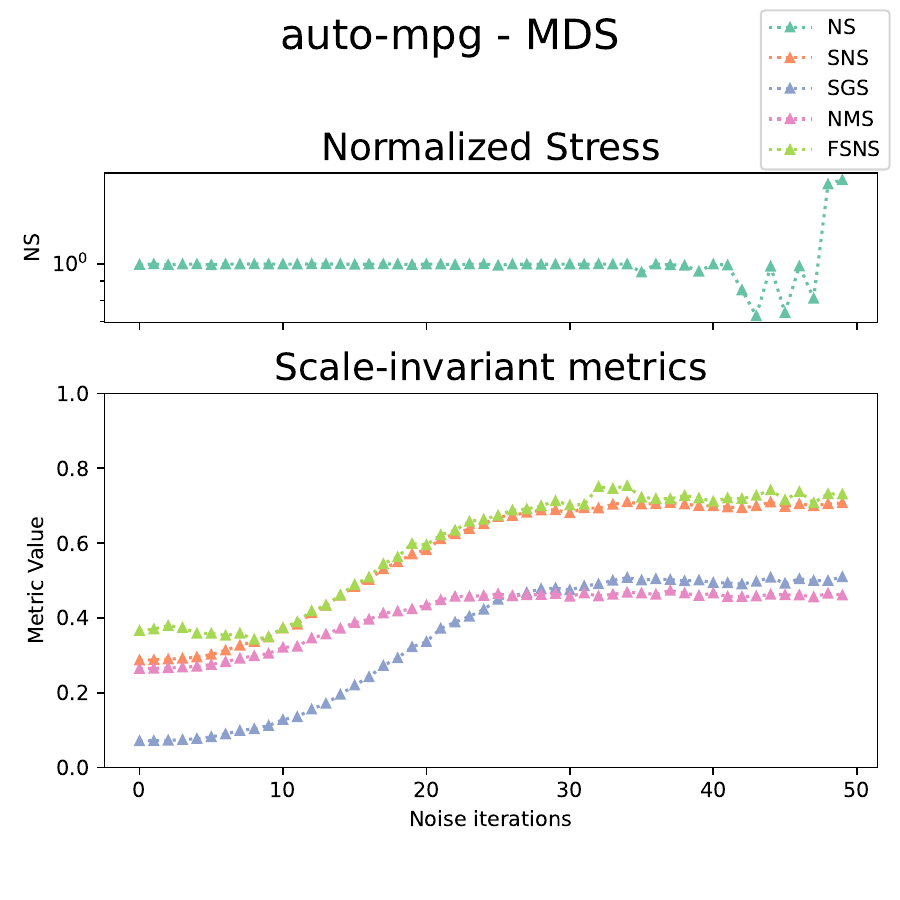}
    \includegraphics[width=0.32\linewidth]{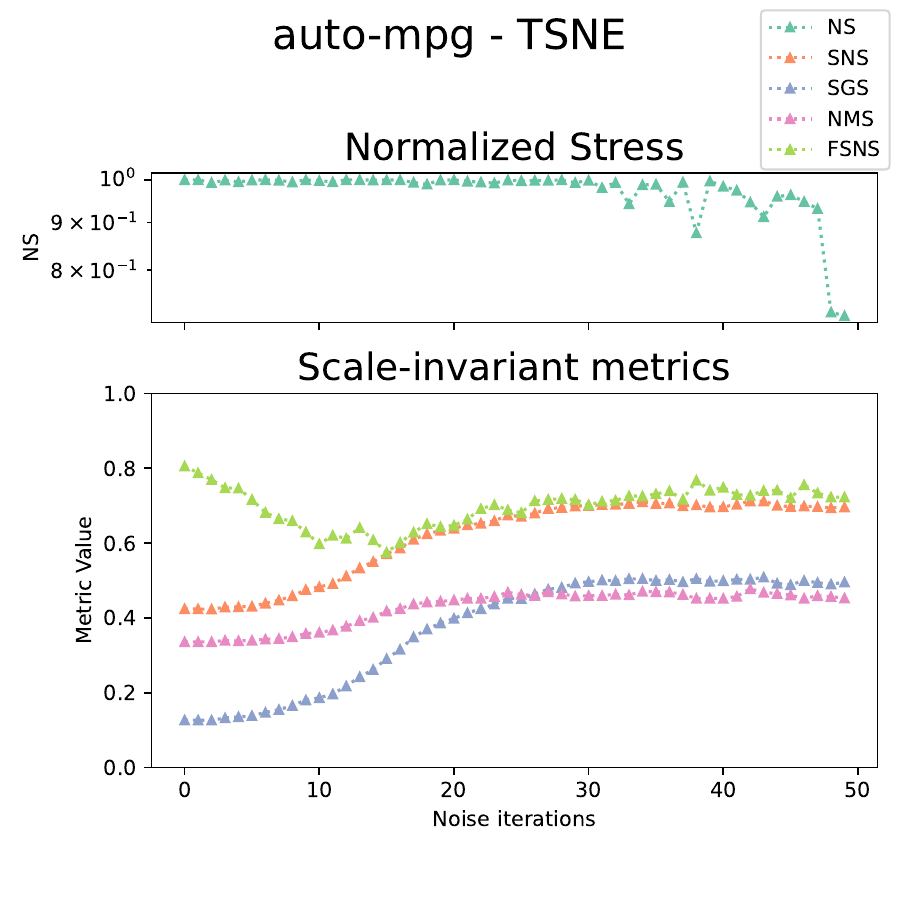}
    \includegraphics[width=0.32\linewidth]{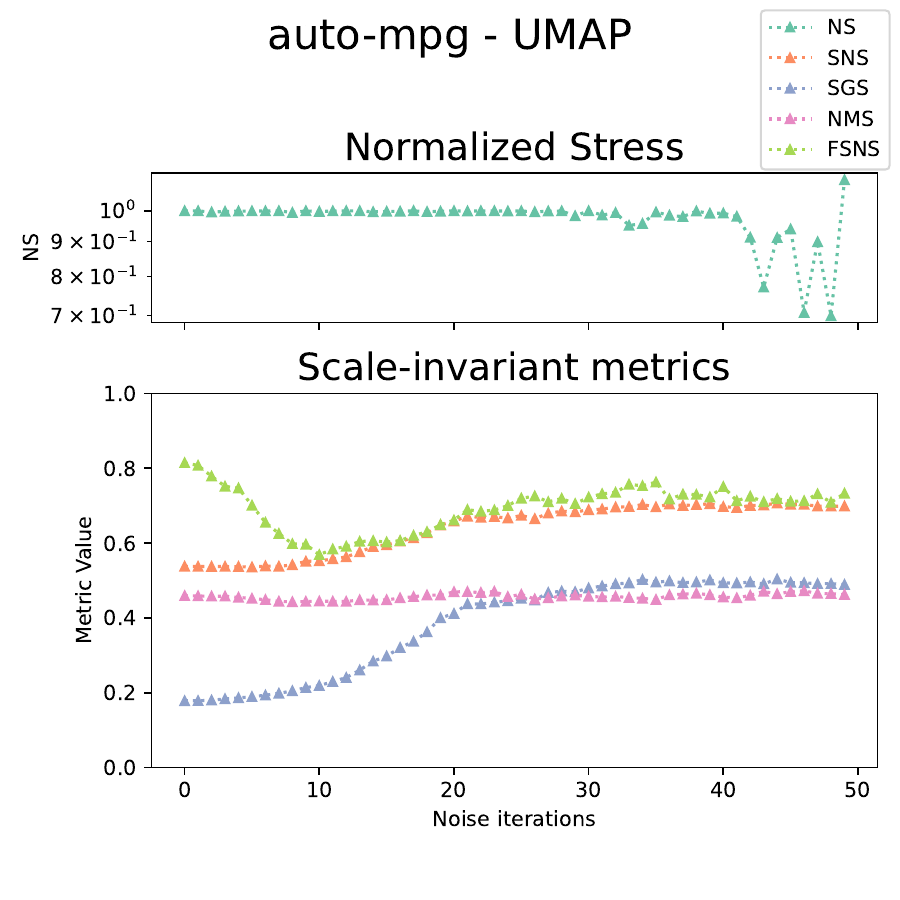}

    \includegraphics[width=0.32\linewidth]{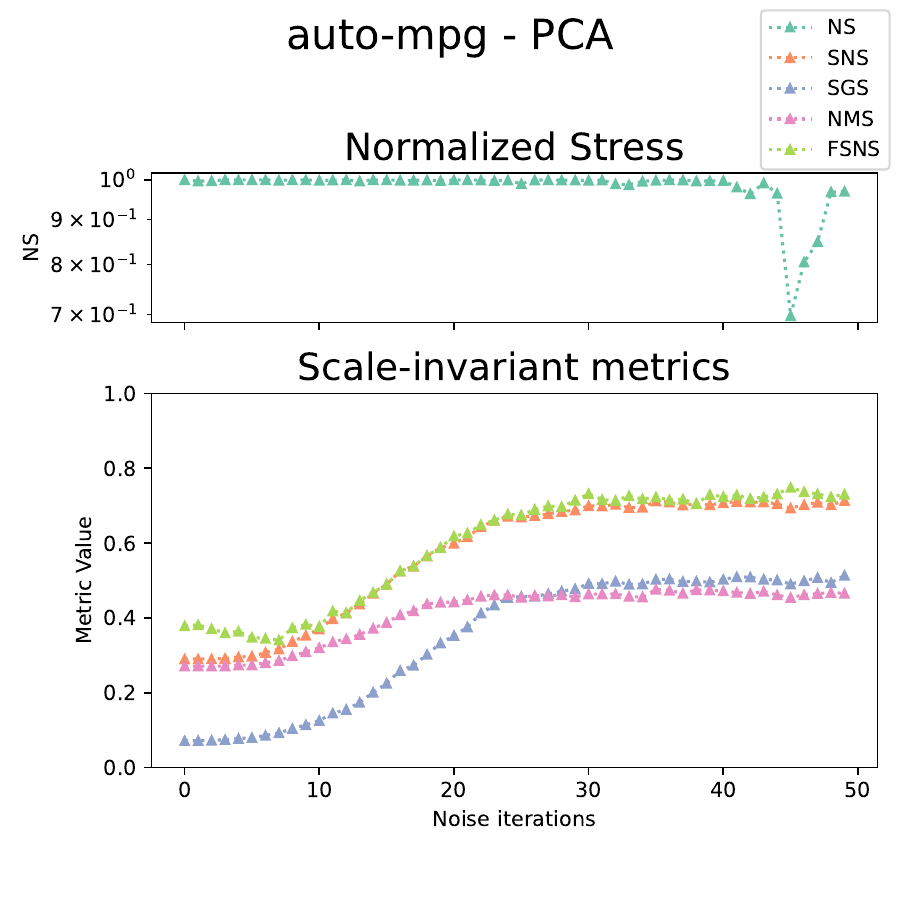}    
    \includegraphics[width=0.32\linewidth]{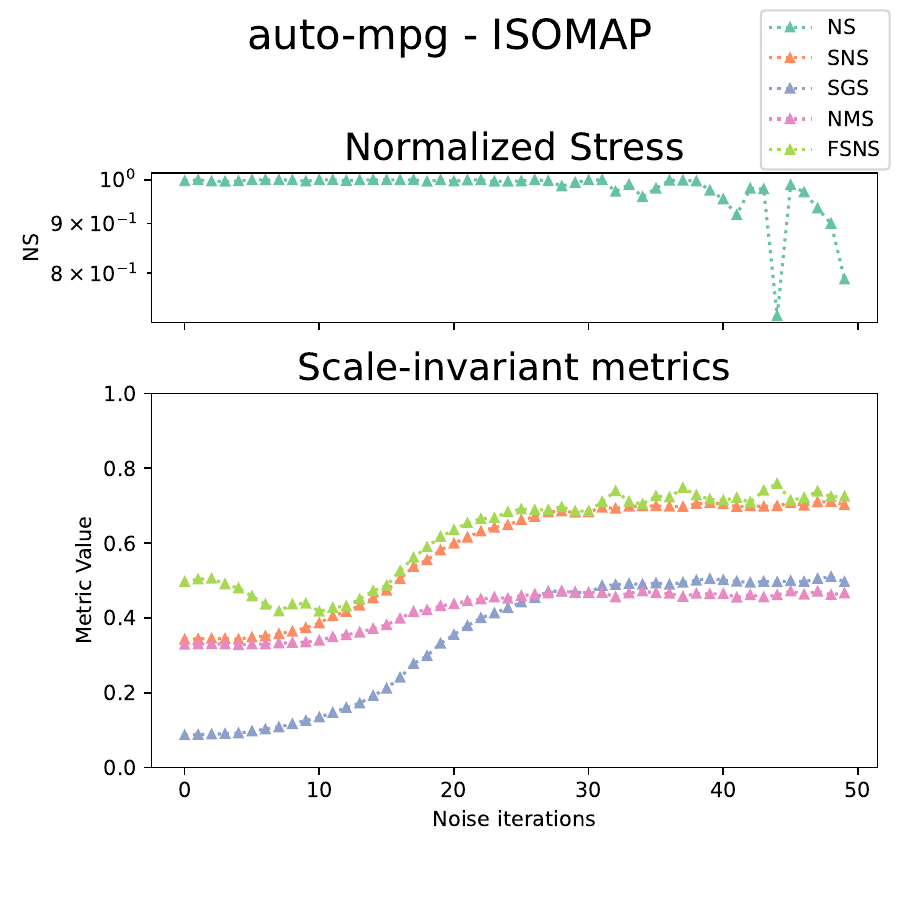}
    \includegraphics[width=0.32\linewidth]{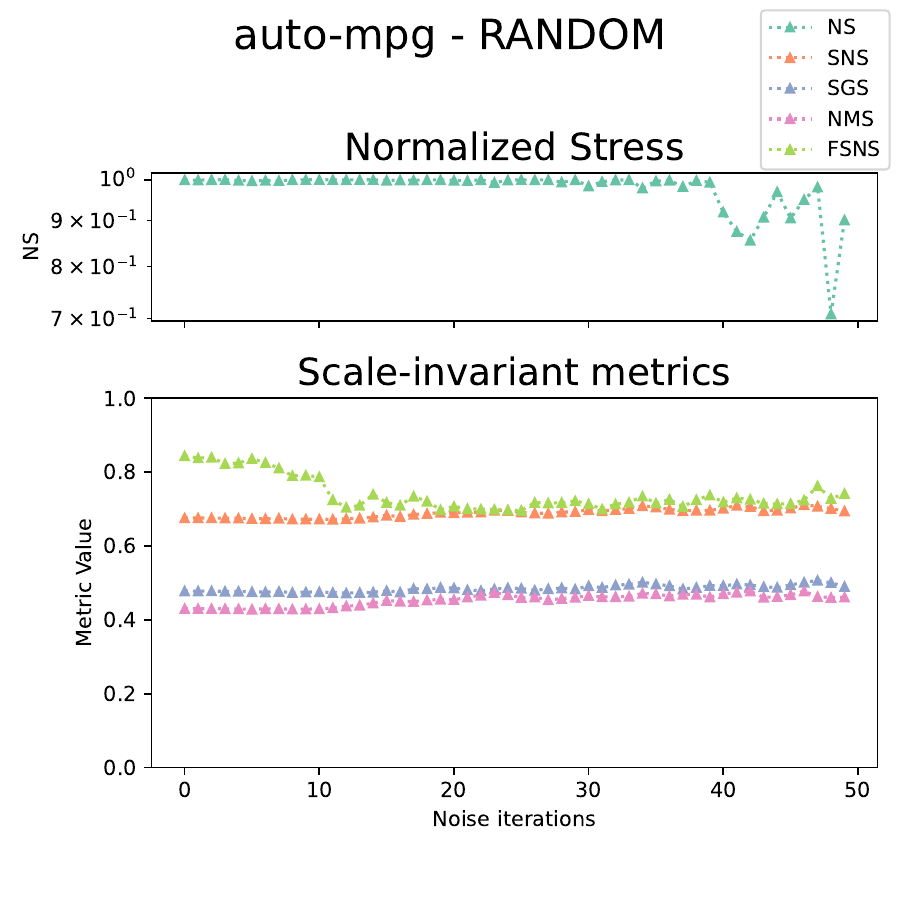}
    \caption{{Each small multiple shows the results of the sensitivity experiment for the Auto-MPG dataset across different algorithms. }}
    \label{fig:ladder-exp-auto}
\end{figure*}

\begin{figure*}
    \centering
    \includegraphics[width=0.32\linewidth]{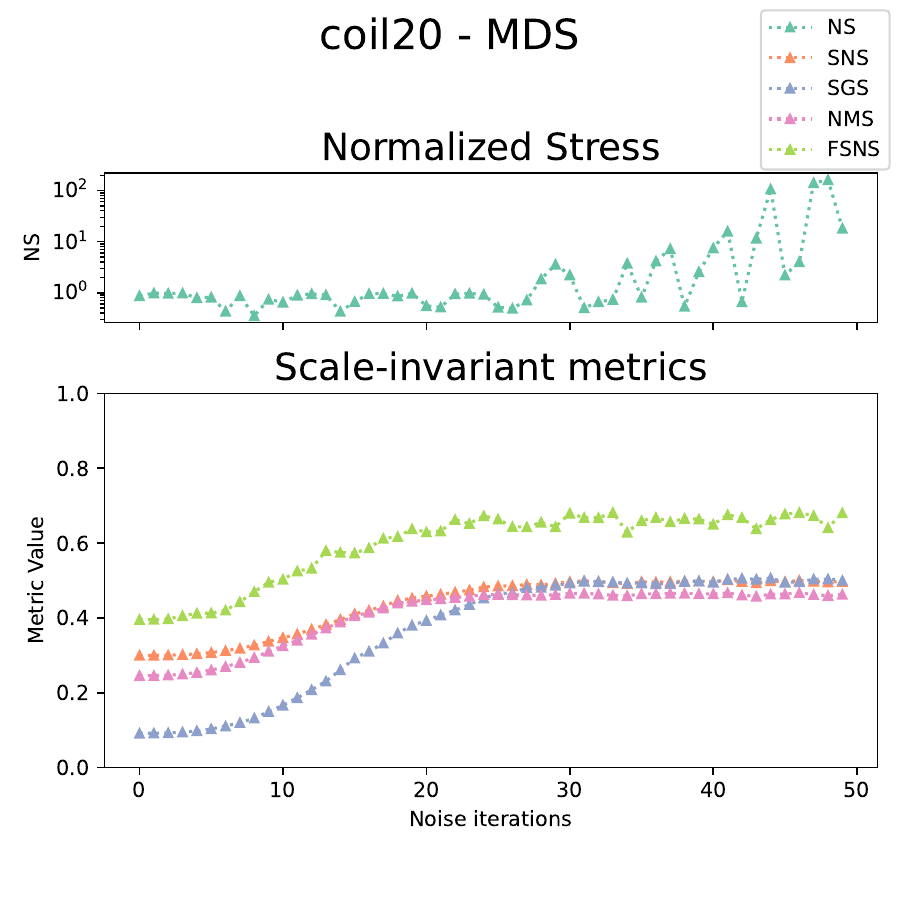}
    \includegraphics[width=0.32\linewidth]{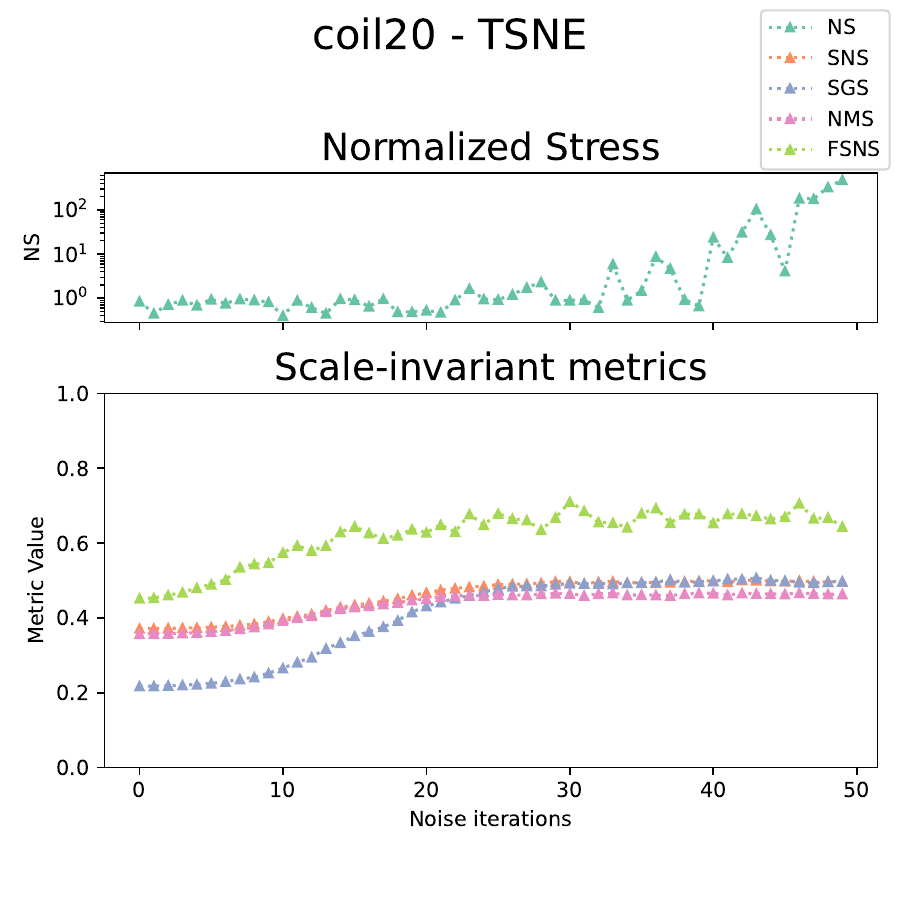}
    \includegraphics[width=0.32\linewidth]{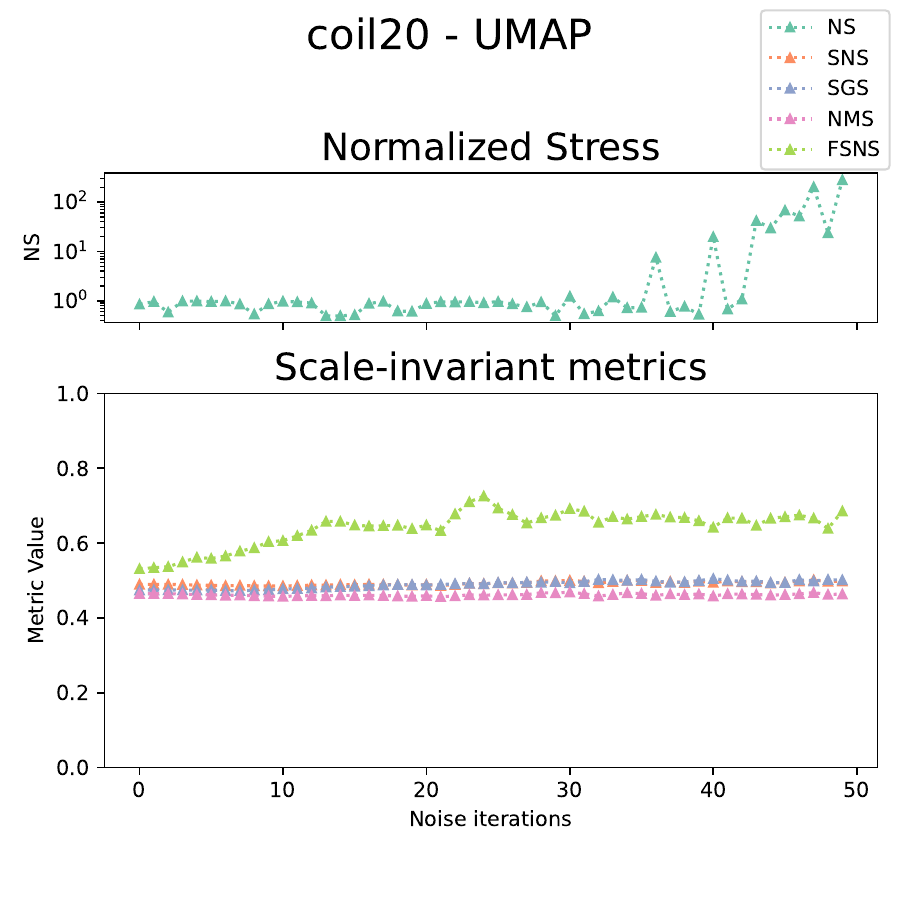}

    \includegraphics[width=0.32\linewidth]{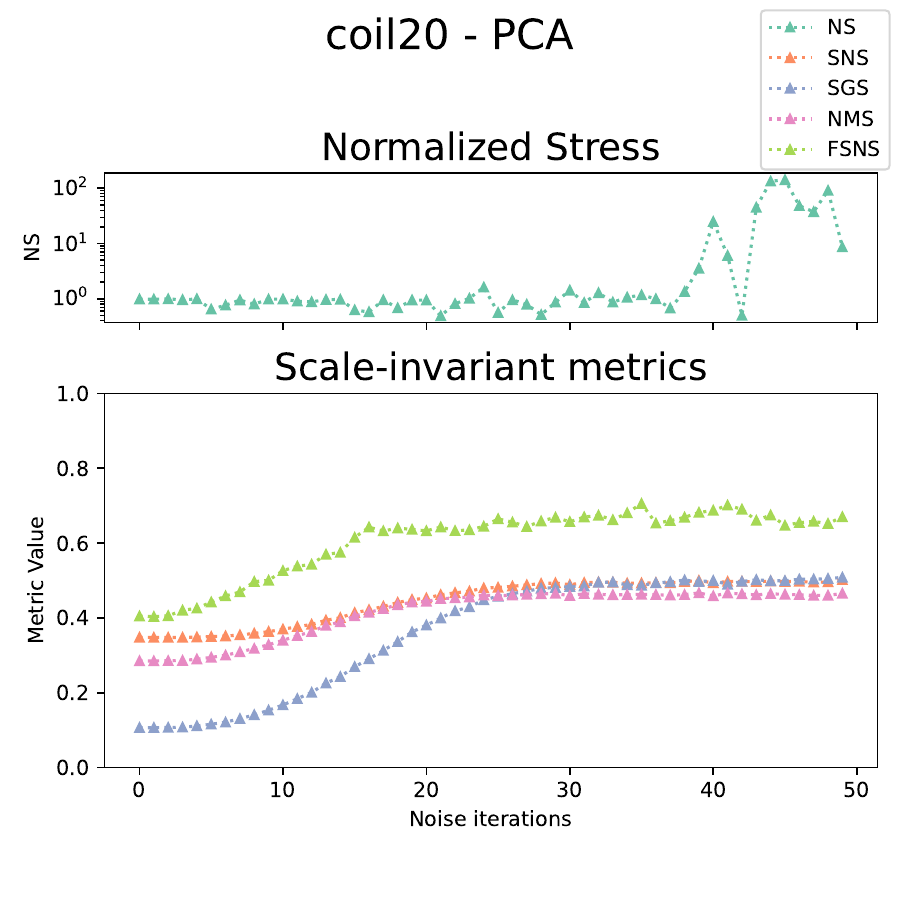}    
    \includegraphics[width=0.32\linewidth]{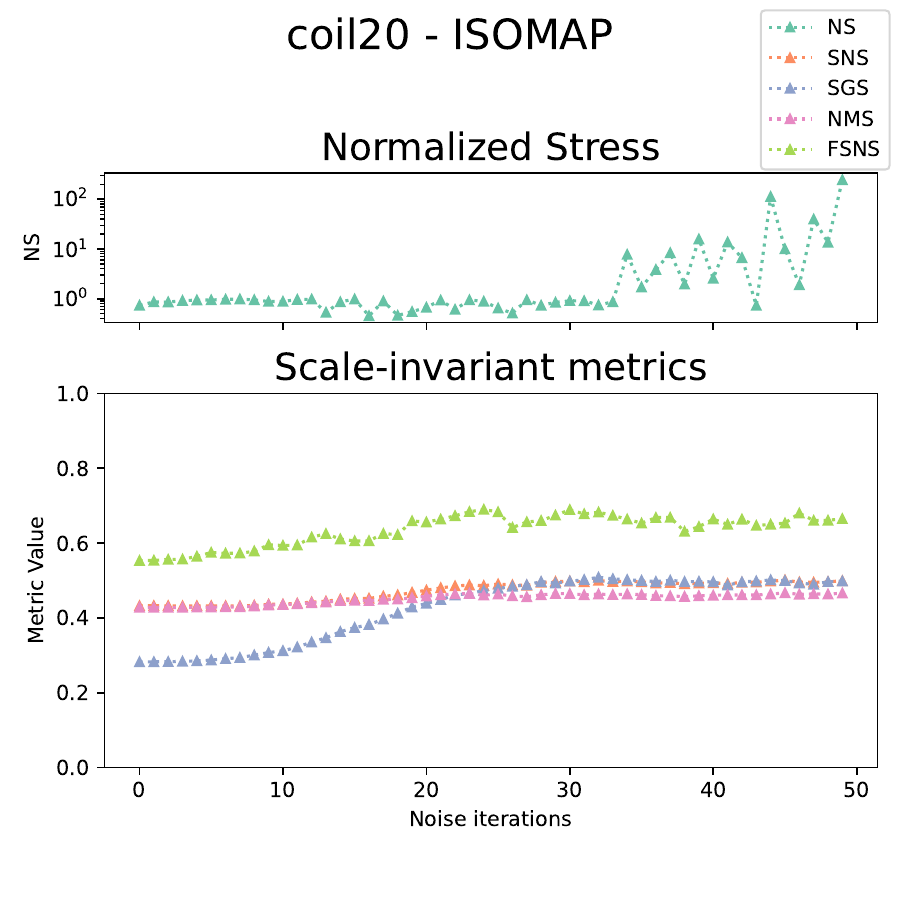}
    \includegraphics[width=0.32\linewidth]{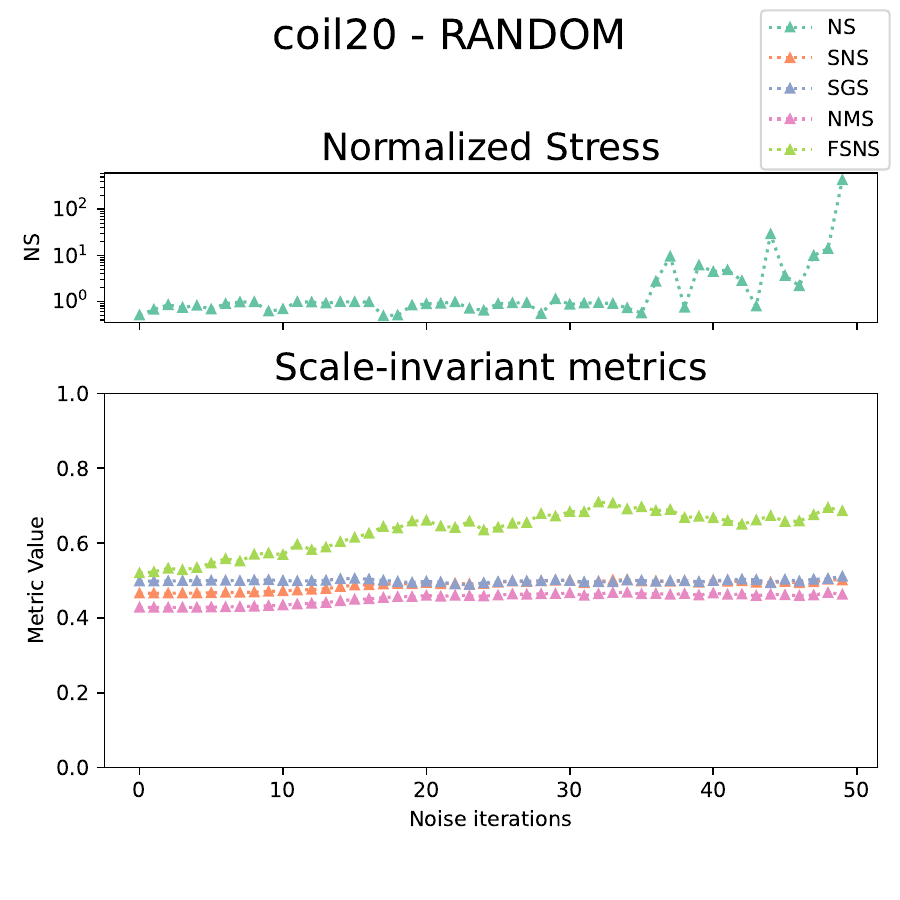}
    \caption{{Each small multiple shows the results of the sensitivity experiment for the Coil20 dataset across different algorithms. }}
    \label{fig:ladder-exp-coil20}
\end{figure*}

\begin{figure*}
    \centering
    \includegraphics[width=0.32\linewidth]{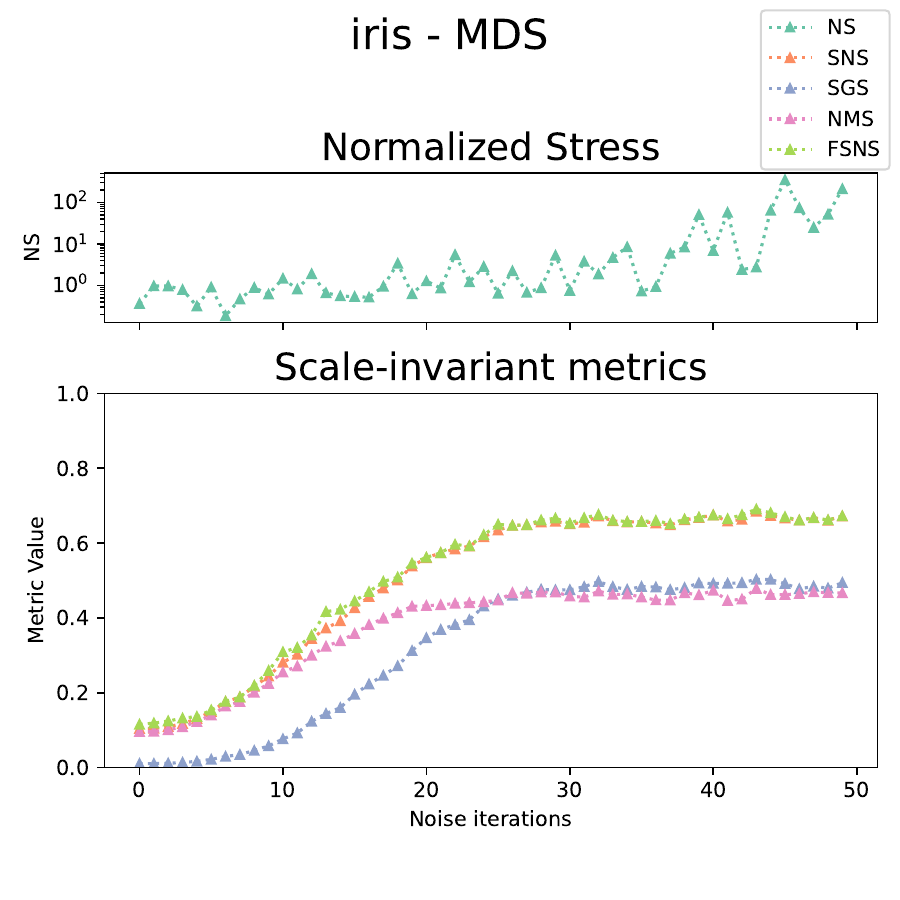}
    \includegraphics[width=0.32\linewidth]{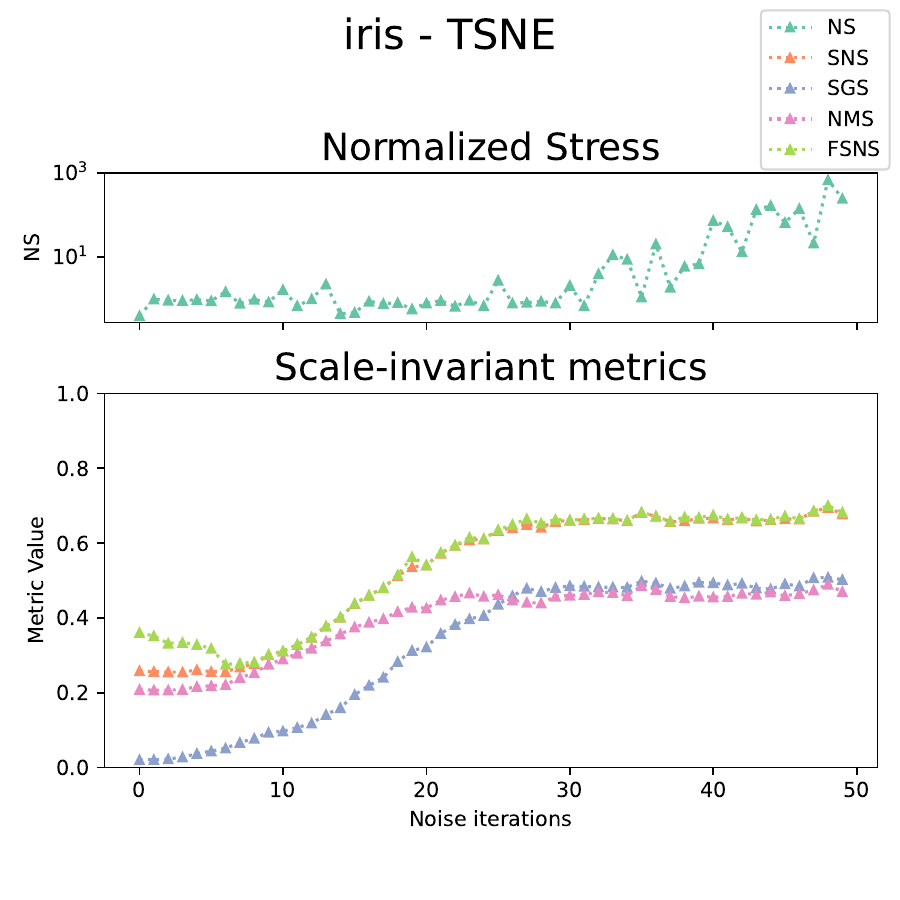}
    \includegraphics[width=0.32\linewidth]{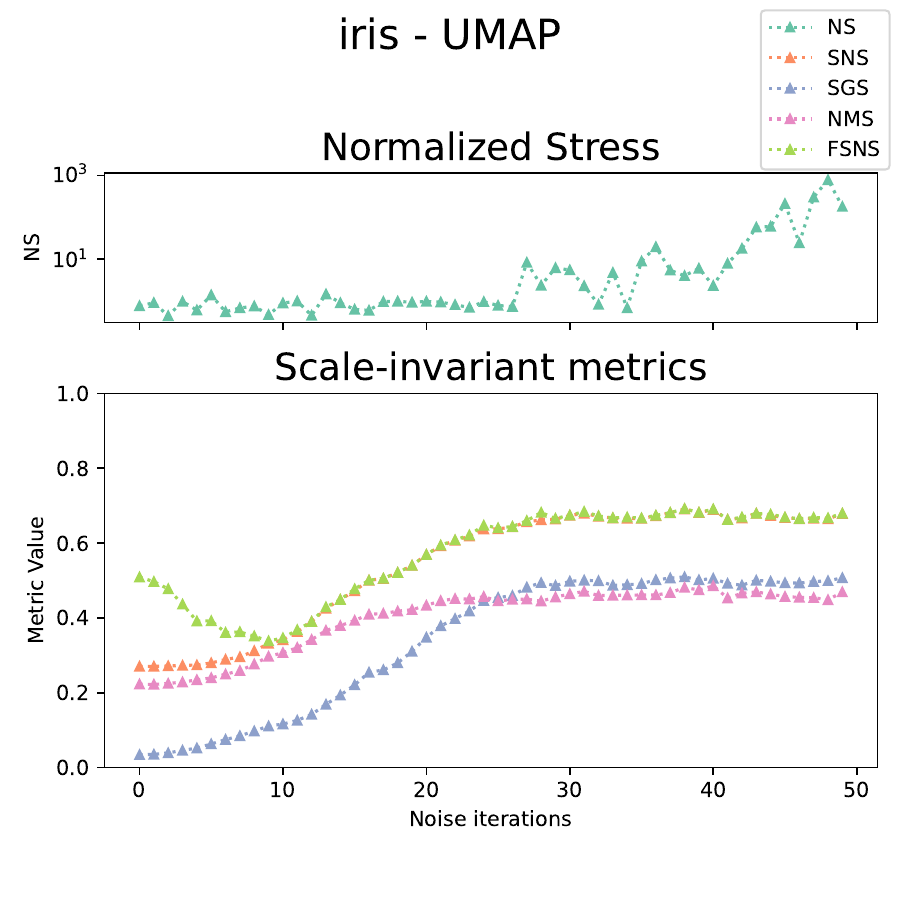}

    \includegraphics[width=0.32\linewidth]{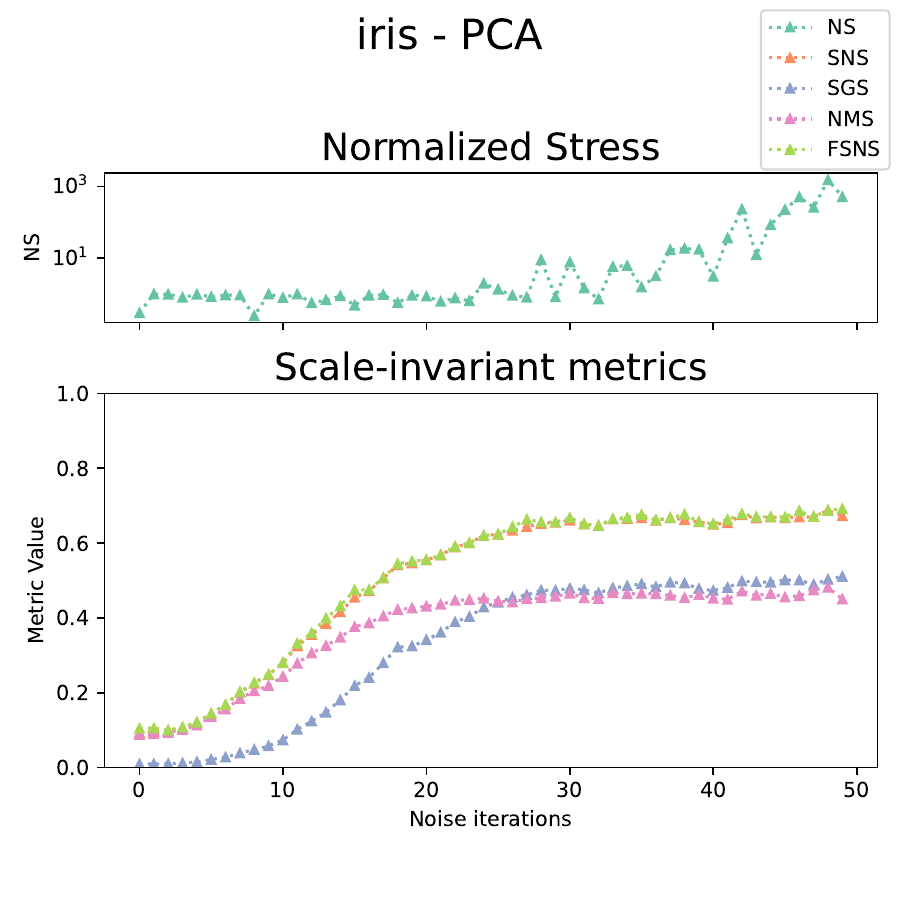}    
    \includegraphics[width=0.32\linewidth]{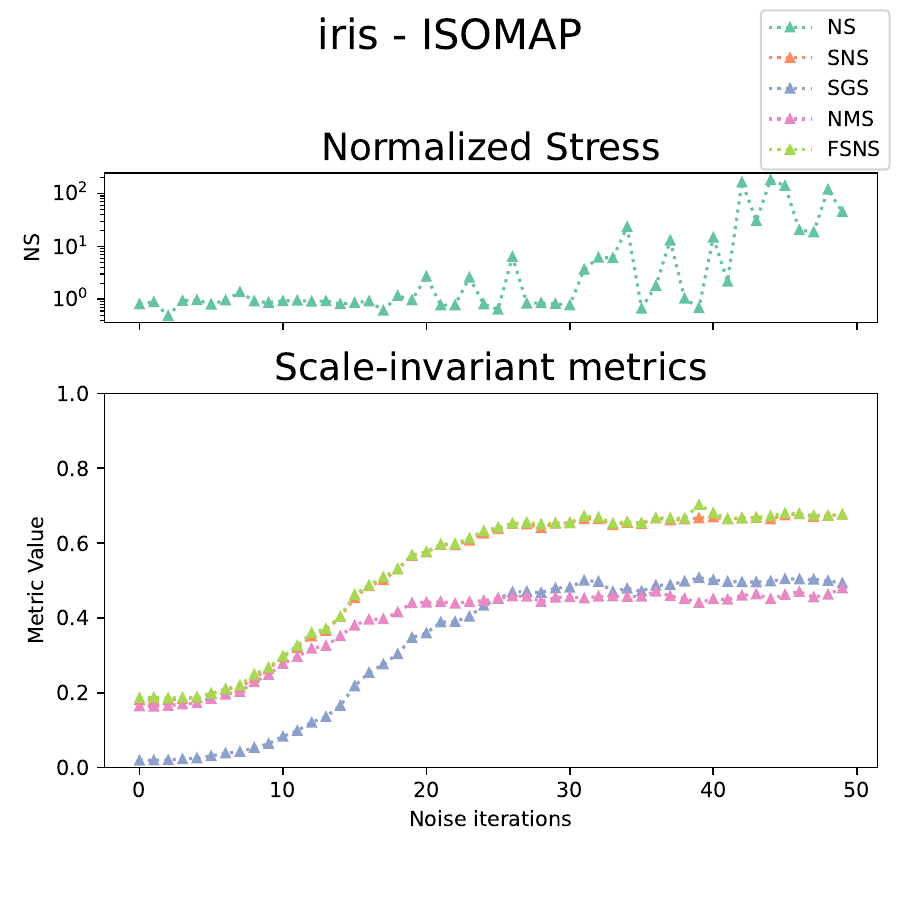}
    \includegraphics[width=0.32\linewidth]{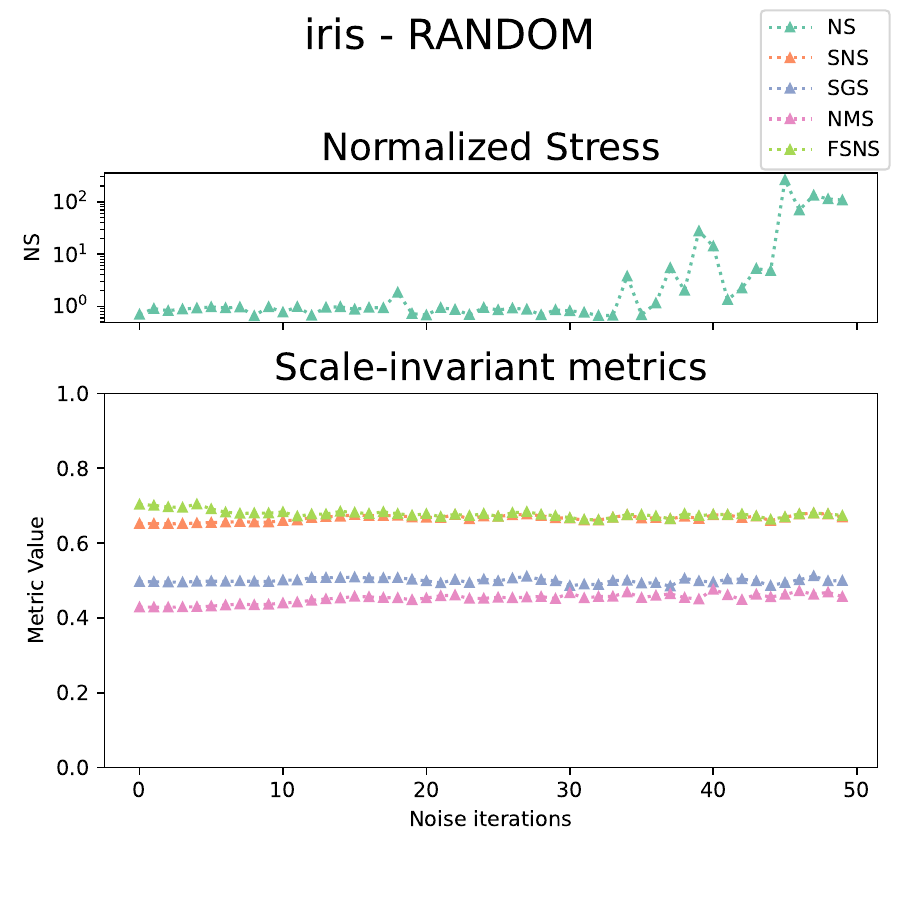}
    \caption{{Each small multiple shows the results of the sensitivity experiment for the Iris dataset across different algorithms. }}
    \label{fig:ladder-exp-iris}
\end{figure*}

\begin{figure*}
    \centering
    \includegraphics[width=0.32\linewidth]{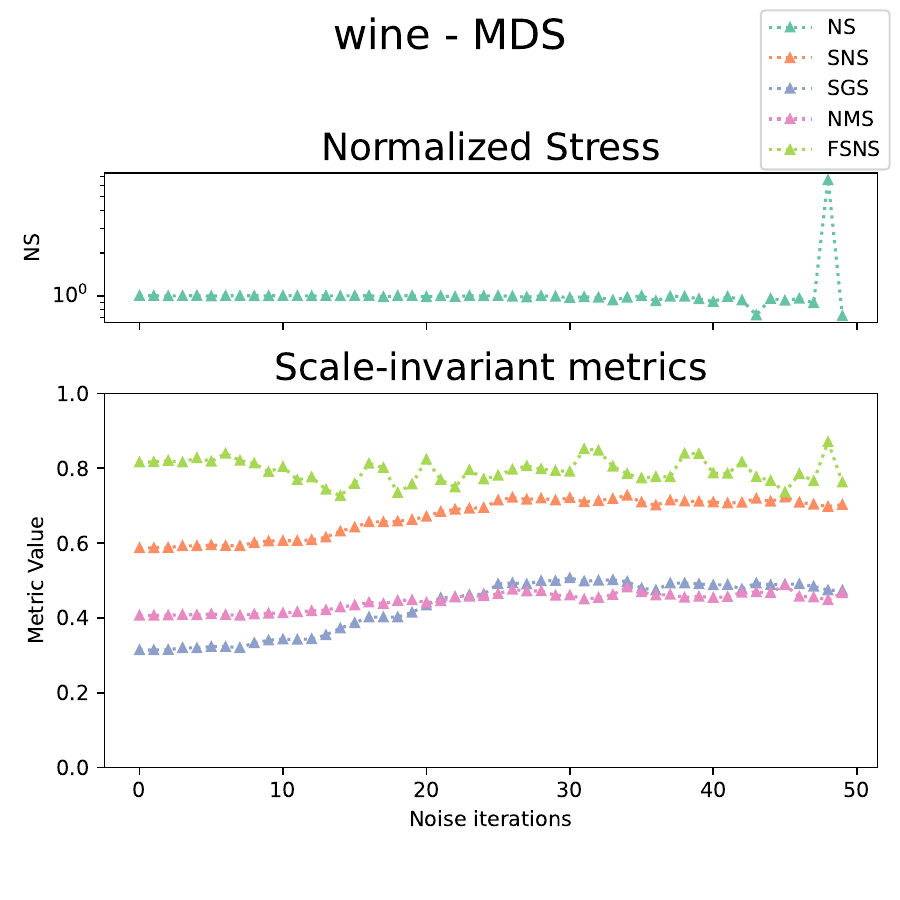}
    \includegraphics[width=0.32\linewidth]{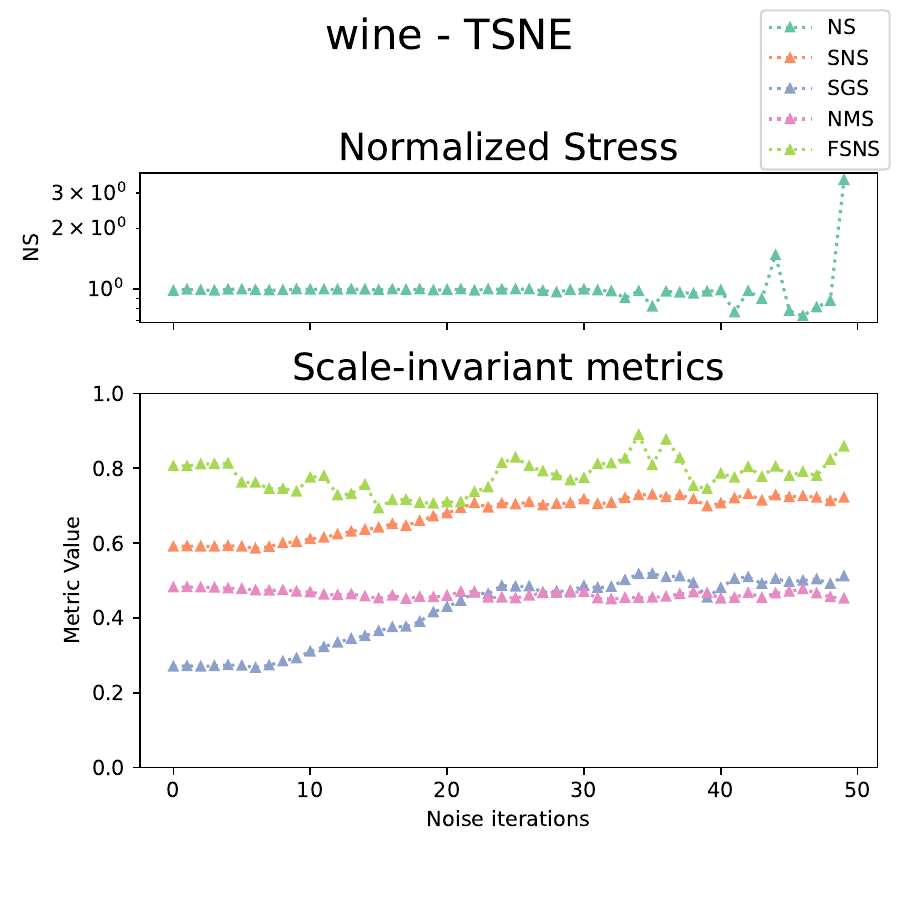}
    \includegraphics[width=0.32\linewidth]{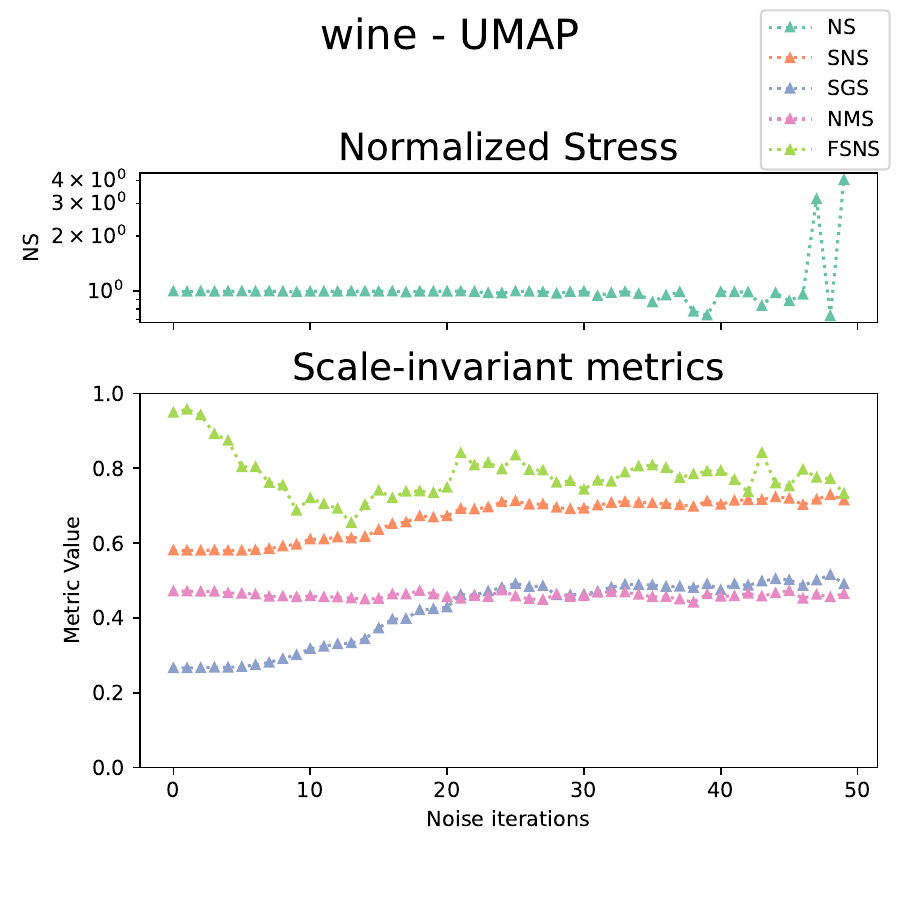}

    \includegraphics[width=0.32\linewidth]{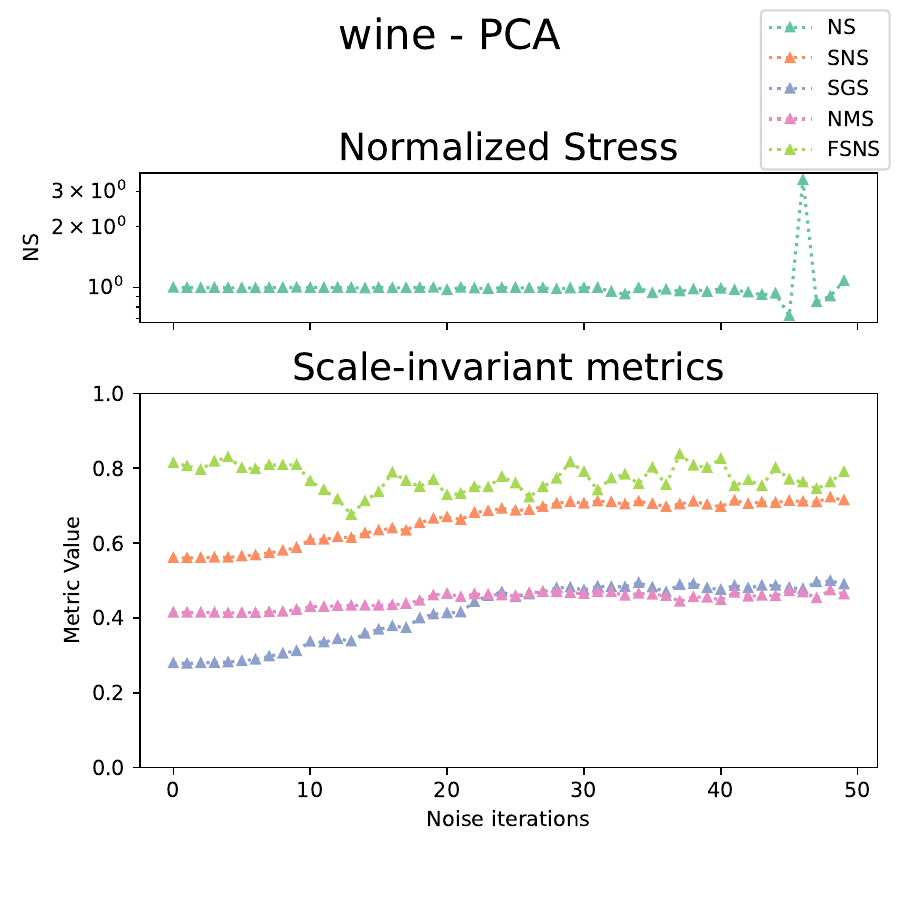}    
    \includegraphics[width=0.32\linewidth]{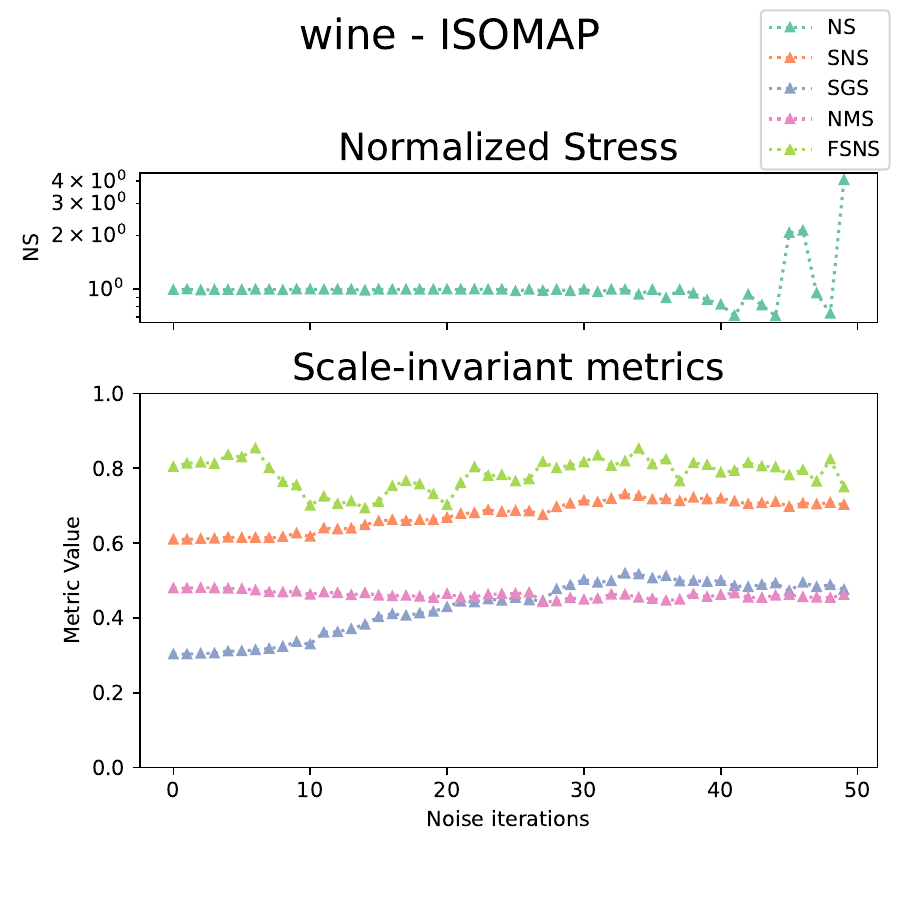}
    \includegraphics[width=0.32\linewidth]{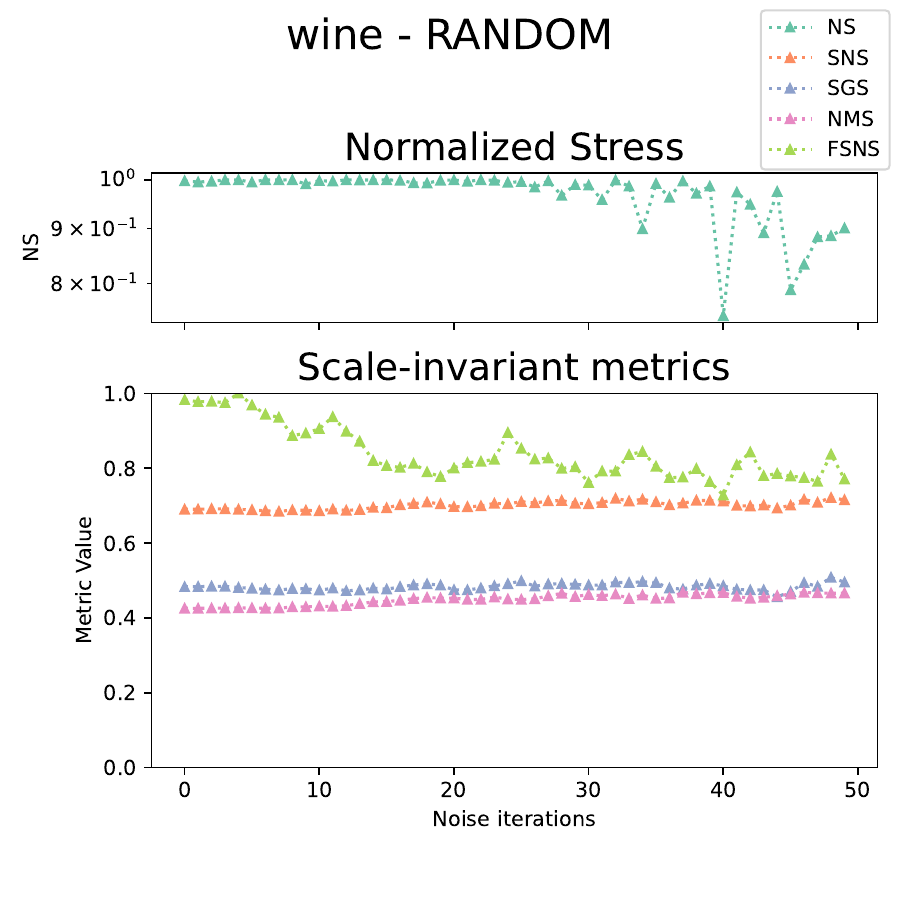}
    \caption{{Each small multiple shows the results of the sensitivity experiment for the Wine dataset across different algorithms. }}
    \label{fig:ladder-exp-wine}
\end{figure*}
-----------------------------------------------------------------------

\end{document}